\documentclass{article} %
\usepackage{iclr2027_conference,times}

\usepackage{amsmath,amsfonts,bm}

\def\eqref#1{equation~\ref{#1}}

\def\1{\bm{1}}

\DeclareMathAlphabet{\mathsfit}{\encodingdefault}{\sfdefault}{m}{sl}
\SetMathAlphabet{\mathsfit}{bold}{\encodingdefault}{\sfdefault}{bx}{n}

\newcommand{\R}{\mathbb{R}}

\renewcommand{\eqref}[1]{\textup{(\ref{#1})}}

\usepackage{graphicx}
\usepackage{booktabs}
\usepackage[ruled,vlined,linesnumbered]{algorithm2e}
\SetAlgoNlRelativeSize{-1}
\usepackage{amsthm}
\usepackage{placeins}
\usepackage{float}
\usepackage{wrapfig}
\usepackage{hyperref}
\makeatother
\usepackage{url}
\usepackage{enumitem}
\usepackage[most]{tcolorbox}
\newtcolorbox{hlbox}{
  enhanced, breakable,
  colback=gray!15, boxrule=0pt, arc=3pt, boxsep=2pt,
  left=6pt, right=6pt, top=3pt, bottom=3pt,
}

\makeatletter
\newif\ifunbreakablecites
\AtBeginDocument{%
  \let\BCITS@natlinkstart\hyper@natlinkstart
  \let\BCITS@natlinkend\hyper@natlinkend
  \def\hyper@natlinkstart#1{\ifunbreakablecites\leavevmode\hbox\bgroup\fi\BCITS@natlinkstart{#1}}%
  \def\hyper@natlinkend{\BCITS@natlinkend\ifunbreakablecites\egroup\fi}%
}
\makeatother

\usepackage{xcolor}
\usepackage{comment}

\newcommand{\bx}{\mathbf{x}}
\newcommand{\by}{\mathbf{y}}
\newcommand{\bz}{\mathbf{z}}
\newcommand{\br}{\mathbf{r}}

\newcommand{\bb}{\mathbf{b}}
\newcommand{\bg}{\mathbf{g}}
\newcommand{\bw}{\mathbf{w}}

\newcommand{\rmd}{\mathrm{d}}

\newcommand{\gradPhit}{\nabla\Phi_t(\bx)}

\newcommand{\LAmap}[1]{\mathcal{D}_t\left(#1\right)}
\newcommand{\Estar}[1]{E^\star\left(#1\right)}
\newcommand{\Eqt}[1]{\mathbb{E}_{q_t}[#1]}
\newcommand{\Varqt}[1]{\operatorname{Var}_{\bx\sim q_t}\left[#1\right]}
\newcommand{\Fhat}{\widehat{F}}
\newcommand{\fhat}{\widehat{f}}

\newcommand{\Lhat}{\widehat{L}}
\newcommand{\pstar}{p^{\star}}

\title{Error-Corrected Inference-Time Scaling for Imperfect Diffusion Models}

\author{
Zuokai Wen$^{1}$ \quad
Louis Grenioux$^{2}$ \quad
Weinan E$^{3}$ \quad
Jiequn Han$^{2}$\thanks{Correspondence to jhan@flatironinstitute.org} \\
$^{1}$School of Artificial Intelligence, Shanghai Jiao Tong University \\
$^{2}$Center for Computational Mathematics, Flatiron Institute \\
$^{3}$School of Mathematical Sciences, Peking University
}

\newtheorem{theorem}{Theorem}
\newtheorem{definition}[theorem]{Definition}

\iclrfinalcopy %
\begin{document}
\maketitle

\fancyhead{}
\lhead{Preprint}

\begin{abstract}
Inference-time scaling adapts pretrained diffusion models to new sampling tasks without additional training.
Existing methods rely primarily on Monte Carlo sampling with more particles, yet are premised on the pretrained model being exact.
In practice, data and training limitations make the model imperfect, and these methods inherit its error.
More particles reduce Monte Carlo error but cannot remove the mismatch between the endpoint and the desired target or the error in tracking the prescribed probability path.
We introduce the \emph{Energy-based Feynman--Kac Corrector (EBFKC)}, a framework for energy-based diffusion models that corrects these errors on the fly given a reference energy.
We first derive Feynman--Kac dynamics that track a prescribed path exactly in the continuous-time population limit even when the model is imperfect, and approximate these dynamics using sequential Monte Carlo with variance-controlling guidance.
To remove the endpoint mismatch, we use the pretrained energy as a surrogate along the diffusion path and progressively incorporate the discrepancy between the learned and target terminal energies.
Experiments on Gaussian mixture models, particle systems, alanine dipeptide, and alanine tetrapeptide show that our method closely matches target distributions and molecular free-energy profiles under annealing and reward tilting, whereas standard inference-time scaling baselines retain substantial sampling errors.
\end{abstract}

\section{Introduction}

Diffusion models \citep{sohl2015deep,ho2020denoising,song2020score}
have become a leading paradigm in generative modeling, achieving
remarkable performance across high-dimensional applications ranging
from visual synthesis
\citep{peebles2023scalable,rombach2022high,karras2022elucidating}
to scientific discovery
\citep{watson2023novo,duan2023accurate,xu2022geodiff}. By learning a
time-dependent score function, a diffusion model defines a transport
from a tractable prior distribution to a complex target distribution.
However, training such models is expensive, particularly as modern datasets and
architectures continue to grow. It is therefore often
impractical to train a separate diffusion model for every new target distribution of interest.

One natural strategy is to perform additional training after
pretraining, commonly referred to as fine-tuning
\citep{domingo2024adjoint,han2024stochastic,uehara2024fine,liu2026flow}.
However, fine-tuning remains computationally expensive. This limitation
has motivated inference-time scaling, which adapts a pretrained
diffusion model to a new sampling task by allocating additional
computation at inference time. One class of existing approaches consists
of guidance-based methods, which modify the reverse dynamics using
classifier guidance \citep{dhariwal2021diffusion}, classifier-free
guidance \citep{ho2022classifier}, or many variants~\citep{he2024manifold,guo2024gradient,li2026robust}.
\unbreakablecitestrue
More recently, probabilistic inference
methods have gained attention as a principled approach to inference-time
scaling by casting the problem within Bayesian inference and Monte Carlo
sampling frameworks \citep{chung2022diffusion,bruna2024provable}, with
particular emphasis on Sequential Monte Carlo (SMC) methods~\citep{wu2023practical,ren2025driftlite, wang2026simple}.
In parallel, energy-based diffusion models have been developed~\citep{yu2026learning,ouyang2026diffusive,guth2026learning}, whose energy representations also support compositional generation and control with Monte Carlo methods~\citep{du2023reduce,thornton2025composition}.
\unbreakablecitesfalse

Despite their theoretical grounding, these methods commonly assume the pretrained diffusion model is perfect~\citep{ren2025driftlite,he2025rne,wang2026simple}.
In practice, finite data, limited capacity, and imperfect optimization make the learned score only approximate, so the model's marginals deviate from the true ones.
Because existing methods treat these approximate marginals as perfect, they inherit this error: adding particles reduces Monte Carlo error but cannot remove the pretrained model error itself.
The error is especially pronounced in out-of-distribution regions, where the learned score is less reliable yet the model is often deployed~\citep{lewis2025scalable}.
Tellingly, \citet{soiffer2026catastrophic} show that in this regime it is score estimation error, not the vanilla inference-time approximation error, that dominates. Inference-time reweighting derived under an exact-model assumption can therefore leave substantial pretrained model error uncorrected, motivating our goal of correcting this error.

In this work, we introduce error-corrected inference-time scaling, realized by the \emph{Energy-based Feynman--Kac Corrector (EBFKC)}, a framework that corrects pretrained model error on the fly given a reference energy.
The reference energy specifies which distribution to sample but not how to explore it: on high-dimensional, multimodal targets, Monte Carlo sampling from the energy alone mixes slowly across modes, whereas a pretrained diffusion model already encodes a learned transport that reaches them efficiently.
We first derive the full Feynman--Kac reweighting potential for any prescribed measure transport path, without assuming that the underlying score satisfies the base diffusion equation.
We approximate the resulting dynamics with SMC and use variance-controlling guidance to mitigate particle-weight degeneracy.
Then we construct an energy-aligned path that uses a pretrained energy-based diffusion model as a surrogate along the diffusion path while progressively correcting the discrepancy between the learned and target energies to remove the endpoint mismatch.
To our knowledge, EBFKC is the \emph{first} inference-time scaling method that corrects the path-tracking error of an imperfect pretrained model and, treating the pretrained energy as an approximate surrogate, provably realigns the terminal distribution to the target energy, rather than composing or reweighting several assumed-exact models.

\paragraph{Our Contributions.} Building on this insight,
our work makes the following contributions:
\begin{itemize}[label=\textbullet, wide, itemsep=0.5em]
\item \textbf{Energy-based Feynman--Kac Corrector (EBFKC).} We derive the Feynman--Kac--type Fokker--Planck equation for a general prescribed probability path, freeing it from the model's own marginals, and adapt the Variance-Controlling SMC~\citep{ren2025driftlite} to track it while controlling particle-weight variance. Unlike the Feynman--Kac Corrector~\citep{skreta2025feynman}, EBFKC tracks the path exactly at the population level even when the pretrained model is imperfect.

\item \textbf{Progressive target-energy correction.} Instantiating this freedom, we keep the imperfect pretrained energy as a surrogate proposal while using the reference energy to pin the terminal distribution to the desired energy-aligned target (EAT), removing the endpoint mismatch that increasing the number of particles alone cannot eliminate.
We call the resulting method \emph{EBFKC on EAT}.

\item \textbf{Error correction across benchmarks.} Across Gaussian mixture models, particle systems, alanine dipeptide, and alanine tetrapeptide, we show that a moderately imperfect pretrained model can severely bias existing inference-time scaling methods, and that EBFKC on EAT corrects this bias at inference time, recovering target distributions and molecular free-energy profiles where these baselines fail.
\end{itemize}

\section{Background and Problem Formulation}

We review score- and energy-based diffusion models, then formulate
inference-time scaling and explain how pretrained model error affects
both the terminal target and path tracking.

\subsection{Score- and Energy-Based Diffusion Models}
\label{sec:score_energy_diffusion}
Diffusion models~\citep{sohl2015deep,ho2020denoising,song2020score} connect the data distribution $\pstar$ to a tractable reference $\rho$ through a forward noising diffusion
\begin{align}
    \rmd\by_s = u_s(\by_s)\,\rmd s + \beta_s\,\rmd\bw_s,
    \qquad \by_0\sim \pstar,
    \qquad s\in[0,T].
    \label{eq:forward_noising_process}
\end{align}
The coefficients $u_s,\beta_s$ are chosen so that the process carries $\by_0\sim\pstar$ to $\by_T\sim\rho$ (typically very close to a standard Gaussian); we take the drift linear, $u_s(\by)=-F_s\by$, so the forward transition kernels are Gaussian in closed form.
Generation runs a reverse clock $t=T-s$; let $p^{\star}_t$ denote the law of $\by_{T-t}$, with $\widetilde{u}_t:=u_{T-t}$ and $\sigma_t:=\beta_{T-t}$ (so $p^\star_0=\rho$ and $p^\star_T=\pstar$). The exact reverse-time drift generating marginals $(p^{\star}_t)_{t\in[0,T]}$ is
\begin{align}
    v_t^\star(\bx) = -\widetilde{u}_t(\bx) + \sigma_t^2 s_t^\star(\bx),
    \qquad
    s_t^\star(\bx) := \nabla_{\bx}\log p^{\star}_t(\bx).
    \label{eq:exact_reverse_drift}
\end{align}
Since $\pstar_t$ is unavailable, diffusion models learn a score model $s_\theta(\bx,t)\approx s_t^\star(\bx)$ from samples of \eqref{eq:forward_noising_process}~\citep{ho2020denoising,song2020score}; we assume such a pretrained model is available. Substituting $s_\theta$ for $s_t^\star$ gives the learned drift $v_t^\theta:=-\widetilde{u}_t+\sigma_t^2 s_\theta$ and the generative diffusion
\begin{align}
    \rmd\bx_t = v_t^\theta(\bx_t)\,\rmd t + \sigma_t\,\rmd\bw_t,
    \qquad \bx_0\sim\rho,
    \qquad t\in[0,T],
    \label{eq:learned_reverse_eq}
\end{align}
whose law we denote $\mu_t$. If $s_\theta=s_t^\star$ for all $(\bx,t)$, the generative and exact reverse diffusions share the same drift and initial law $\rho$, so $\mu_t=\pstar_t$ for all $t$ and in particular $\mu_T=\pstar$. When $s_\theta$ approximates $s_t^\star$, $\mu_T$ is only an approximation of $\pstar$.

The score model $s_\theta$ is often parameterized directly, as in standard score-based models. We instead adopt an \emph{energy-based} parameterization~\citep{Salimans2021should}: a scalar energy $E_\theta(\bx,t)$ defines $p_t^\theta(\bx)=\exp(-E_\theta(\bx,t))/Z_t^\theta$ (with $Z_t^\theta:=\int\exp(-E_\theta(\bx,t))\,\rmd\bx$), and the score follows as $s_\theta=-\nabla_{\bx}E_\theta=\nabla\log p_t^\theta$. Unlike a directly parameterized score, which is a general vector field with no potential and hence no associated density, this yields explicit intermediate densities $(p_t^\theta)_{t\in[0,T]}$; they need not coincide with the reverse-SDE law $\mu_t$, since fixing the spatial score does not enforce the correct evolution in time, nor with the truth $p_t^\star$. They further underlie energy-based diffusion's strength in compositional generation~\citep{du2023reduce,thornton2025composition}, unnormalized density evaluation~\citep{yu2025density,guth2026learning}, and downstream reweighting~\citep{ouyang2026diffusive}.

\begin{hlbox}
\textbf{Why we start from energy-based diffusion.} The same explicit densities $p_t^\theta$, which a bare score cannot provide, are what our method later uses to align the target and to reweight along the prescribed path.
\end{hlbox}

\subsection{Inference-Time Scaling with Imperfect Models}
\label{sec:inference_time_formulation}

Inference-time scaling adapts a pretrained diffusion model to a new sampling target without retraining. In full generality, the goal is to sample from
\begin{align}
    q_T^{\mathrm{ref}}(\bx)
    \propto \left(p^{\mathrm{ref}}(\bx)\right)^\gamma\exp\left(r(\bx)\right),
    \qquad
    p^{\mathrm{ref}}(\bx)\propto\exp\left(-E^{\mathrm{ref}}(\bx)\right),
    \label{eq:general_target}
\end{align}
for an annealing exponent $\gamma>0$, a reward $r$, and a time-independent \emph{reference energy} $E^{\mathrm{ref}}$ ($r=0$ gives the \emph{annealing} task, $\gamma=1$ the \emph{reward-tilting} task). The pretrained model enters because its densities $(p_t^\theta)_{t\in[0,T]}$ are expected to be close to the reference, making it a far better starting point than sampling from $E^{\mathrm{ref}}$ directly. The canonical choice is $p^{\mathrm{ref}}=p^\star$, recovering
\begin{align}
    q_T^\star(\bx)\propto\left(p^\star(\bx)\right)^\gamma\exp\left(r(\bx)\right).
    \label{eq:true_target}
\end{align}
Since each $p_t^\theta$ approximates $p_t^\star$, a natural construction interpolates through the model's own densities \citep{skreta2025feynman,ren2025driftlite,wang2026simple,he2025rne}.

\begin{definition}[Model-Induced Target]
\label{def:model_induced_path}
The \emph{model-induced target (MIT)} is
\begin{align*}
    q_T^{\mathrm{MIT}}(\bx)\propto(p_T^\theta(\bx))^\gamma\exp(r(\bx)),
\end{align*}
which is \eqref{eq:general_target} with $E^{\mathrm{ref}}=E_\theta(\cdot,T)$. It is the terminal distribution of the model-induced path
\begin{align*}
    q_t^{\mathrm{MIT}}(\bx)\propto(p_t^\theta(\bx))^\gamma\exp(r_t(\bx)),
\end{align*}
where the intermediate reward $r_t$ interpolates between $r_0=0$ and $r_T=r$.
\end{definition}

This construction, however, is undermined by an imperfect pretrained model in two distinct ways.

\begin{hlbox}
\textbf{Two failure modes under an imperfect model.} \emph{Endpoint mismatch:} the MIT can differ from the desired $q_T^{\mathrm{ref}}$, even if its path is tracked exactly. \emph{Path-tracking error:} existing reweighting presumes a consistent density $p_t^\theta$ that need not even exist, so it fails to track even the model-induced path.
\end{hlbox}

\textbf{Endpoint mismatch.} Whether this gap matters depends on the reference energy. In data-driven settings such as image generation, the goal is to steer the model's own distribution, so $E^{\mathrm{ref}}=E_\theta(\cdot,T)$ and the MIT is exactly the desired target. In most scientific computing tasks, a physical model supplies an external $E^\star$ and the quantities of interest are defined with respect to $p^\star\propto\exp(-E^\star)$, so the target is $E^{\mathrm{ref}}=E^\star$; here the mismatch must be corrected, yet existing steering methods for pretrained diffusion models are not designed to use a known $E^\star$ for this purpose. More broadly, the same holds whenever the target is defined with the corresponding reference energy $E^{\mathrm{ref}}\neq E_\theta(\cdot,T)$, for which the MIT differs from the desired $q_T^{\mathrm{ref}}$.

\textbf{Path-tracking error.} Concretely, to obtain efficient reweighting formulas that use only the learned score, most existing methods implicitly assume that such a density $p_t^\theta$ exists and satisfies the Fokker--Planck equation associated with \eqref{eq:forward_noising_process}~\citep{skreta2025feynman}. This assumption generally fails for a learned, imperfect score, so the resulting reweighting formulas are biased and no longer track $q_t^{\mathrm{MIT}}$, even at the population level.

\section{Method}

\begin{figure}
    \centering
    \includegraphics[width=0.92\linewidth]{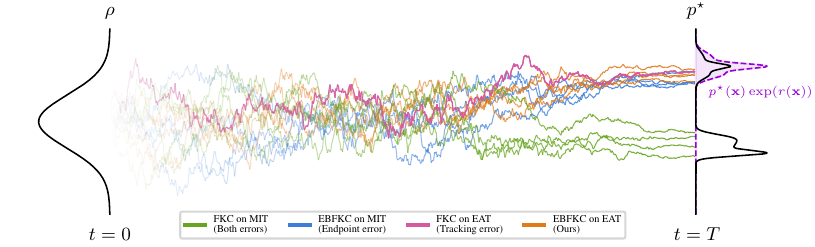}
    \caption{SMC trajectory lineages for the four combinations of reweighting scheme (FKC or EBFKC) and prescribed path (MIT or EAT). The target $p^\star$ is a 64-dimensional mixture of six isotropic Gaussians in two groups (right edge). The figure displays one-dimensional trajectories for four lineages from a population of 2048 particles. The reward $r$ favors the weak mode, with the desired terminal distribution shown in dashed purple. FKC~\citep{skreta2025feynman} on MIT suffers from both errors and ends in the wrong modes. EBFKC on MIT tracks its path but inherits the endpoint mismatch, while FKC on EAT has the correct endpoint but its path-tracking error makes the weights degenerate, so that only one lineage survives. Only EBFKC on EAT, which removes both errors, reaches the desired target.}
    \label{fig:fig1}
\end{figure}

Both issues above point to the same remedy: decouple the sampling path from the model's own marginals.
We therefore track a prescribed path with a general time-dependent potential $\Phi_t$,
\begin{align}
    q_t(\bx)
    =\exp\left(-\Phi_t(\bx)\right)/Z_t,
    \qquad t\in[0,T],
    \label{eq:general_probability_path}
\end{align}
requiring only that the path be normalizable, that $q_0$ admit easy sampling, and that $\Phi_t$ and its required derivatives be evaluable; $Z_t$ need not be known.
The model-induced path is recovered with $\Phi_t^{\mathrm{MIT}}(\bx)=\gamma E_\theta(\bx,t)-r_t(\bx)$.

We then resolve both failure modes with two ingredients, whose effect Fig.~\ref{fig:fig1} illustrates:
\begin{hlbox}
\emph{Energy-based Feynman--Kac Corrector (EBFKC)} (Sec.~\ref{sec:prescribed_path_dynamics}) tracks any prescribed path exactly at the population level, removing the path-tracking error. \emph{Energy-aligned target (EAT)} (Sec.~\ref{sec:energy_augmented_ta_path}) pins the terminal distribution to $q_T^{\mathrm{ref}}$ for any $E^{\mathrm{ref}}$, removing the endpoint mismatch.
\end{hlbox}
Sec.~\ref{sec:connections_smc_ebm} further relates our method to annealed SMC and energy-based modeling.

\subsection{Fokker--Planck Equation for a Prescribed Path}
\label{sec:prescribed_path_dynamics}

For the prescribed path in Eq.~\eqref{eq:general_probability_path}, we derive tracking dynamics directly from its potential $\Phi_t$, without requiring diffusion consistency.
The result applies to any sufficiently smooth, strictly positive, normalizable path, including paths to the learned terminal density or to an externally specified target.

\begin{hlbox}
\begin{theorem}[Fokker--Planck Equation for a Prescribed Path]
\label{thm:general_path_fk}
Assume that $\Phi_t$ is $C^1$ in $t$ and $C^2$ in $\bx$, that $Z_t<\infty$, and that $\widetilde u_t$ is $C^1$ in $\bx$.
For the prescribed path $q_t$ in Eq.~\eqref{eq:general_probability_path},
we define the proposal drift
$
v_t(\bx)
=
-\widetilde{u}_t(\bx)
-
\sigma_t^2
\gradPhit,
$
then $q_t$ satisfies the Feynman--Kac-type Fokker--Planck equation
\begin{align}
\label{eq:fk fpe}
\partial_t q_t
=
-\nabla\cdot(q_t v_t)
+
\frac{\sigma_t^2}{2}\Delta q_t
+
q_t g_t,
\end{align}
where the centered reweighting potential is
$
g_t(\bx)
=
G_t(\bx)
-
\Eqt{G_t}
$,
and the corresponding uncentered potential is
\begin{align}
G_t(\bx)
=
-\partial_t\Phi_t(\bx)
-\nabla\cdot \widetilde{u}_t(\bx)
+
\widetilde{u}_t(\bx)^\top
\gradPhit
-
\frac{\sigma_t^2}{2}
\Delta\Phi_t(\bx)
+
\frac{\sigma_t^2}{2}
\left\|\gradPhit\right\|^2.
\label{eq:general_path_fk_potential}
\end{align}
\end{theorem}
\end{hlbox}

The derivation is provided in
App.~\ref{app:general_path_fk_derivation}.
The state-dependent potential $G_t$ corrects the discrepancy between the proposal diffusion and the prescribed evolution of $q_t$, while centering $G_t$ by its expectation under $q_t$ preserves normalization and eliminates the unknown derivative $\partial_t\log Z_t$.

We simulate Eq.~\eqref{eq:fk fpe} with Sequential Monte Carlo
(SMC). Let $\{(\bx_t^{(i)},w_t^{(i)})\}_{i=1}^N$ denote a system of normalized weighted particles, initialized independently as
$\bx_0^{(i)}\sim q_0\propto\exp(-\Phi_0)$ with $w_0^{(i)}=1/N$. The particles then evolve according to the continuous-time dynamics

\begin{equation}
    \label{eq:weighted_particle_smc}
    \left\{
    \begin{aligned}
        \rmd\bx_t^{(i)}
        &=
        v_t(\bx_t^{(i)})\rmd t
        +
        \sigma_t \rmd\bw_t^{(i)},
        && i\in[N], \\
        \rmd\log w_t^{(i)}
        &=
        \hat{g}_t(\bx_t^{(i)})\rmd t,
        && i\in[N],\\
        \hat{g}_t(\bx_t^{(i)})
        &:=
        G_t(\bx_t^{(i)})
        -
        \textstyle{\sum_{j=1}^{N}
        w_t^{(j)}
        G_t(\bx_t^{(j)})},
        && i\in[N],
    \end{aligned}
    \right.
\end{equation}
In this way, SMC approximates this normalized Feynman--Kac flow using a weighted particle system, with optional resampling to control weight degeneracy.
Simulating Eq.~\eqref{eq:weighted_particle_smc} from $t=0$ to $t=T$ therefore yields a weighted particle approximation of the target distribution $q_T$.
To further mitigate particle weight degeneracy during SMC simulation, we also incorporate Variance-Controlling Guidance (VCG) \citep{ren2025driftlite}, which modifies the particle drift in a principled manner; details are in App.~\ref{sec:vcg_prescribed_path}.
We call this framework the \textbf{Energy-based Feynman--Kac Corrector (EBFKC)}.
We remark that FKC~\citep{skreta2025feynman} follows the same weighted-particle scheme as Eq.~\eqref{eq:weighted_particle_smc} using only the learned score, but its reweighting potential is valid only when $p_t^\theta$ is consistent with the base diffusion (satisfies its Fokker--Planck equation).
Our EBFKC instead evaluates the full potential $G_t$, which requires the energy $E_\theta$ rather than the score alone; in exchange it makes no such assumption and tracks the path $q_t$ exactly in the continuous-time population limit even when the pretrained model is imperfect. 
We also extend FKC to general time-dependent rewards in App.~\ref{app:fkc_time_dependent}.

\subsection{From Model-Induced to Energy-Aligned Target}
\label{sec:energy_augmented_ta_path}

The prescribed-path dynamics provided by EBFKC separate the choice of target from the assumption that the model follows an exact diffusion.
Applying EBFKC to the model-induced path in Definition~\ref{def:model_induced_path} gives \textbf{EBFKC on MIT}.
We now construct a path ending at a general target $q_T^{\mathrm{ref}}$, with the model-induced path recovered with the special choice $E^{\mathrm{ref}}=E_\theta(\cdot,T)$.

\textbf{Energy-Aligned Target.} To reach the reference target rather than the model-induced target, we constrain the terminal distribution of the path.
\begin{definition}[Energy-Aligned Target]
\label{def:target_alignment}
The \emph{energy-aligned target (EAT)} is
$
q_T^{\mathrm{EAT}}(\bx)
\propto
\exp\left(-\Phi_T(\bx)\right)
$ with
\[
\Phi_T(\bx)
= \gamma E^{\mathrm{ref}}(\bx)-r(\bx)+c_T,
\]
where $c_T$ is independent of $\bx$.
\end{definition}

Energy-aligned target constrains only the terminal distribution: the intermediate distributions can be constructed flexibly, provided that the resulting path terminates at $q_T^\mathrm{ref}$.
We retain the model-induced path and progressively incorporate the terminal energy correction
$
\Delta E^{\mathrm{ref}}(\bx)
=E^{\mathrm{ref}}(\bx)-E_\theta(\bx,T).
$
For a smooth correction schedule $\kappa_t$ with $\kappa_0=0$ and $\kappa_T=1$, we instantiate the energy-aligned path through the potential
\begin{align*}
\Phi_t^{\mathrm{EAT}}(\bx)
={}&
\Phi_t^{\mathrm{MIT}}(\bx)
+\gamma\kappa_t\Delta E^{\mathrm{ref}}(\bx)
=
\gamma E_\theta(\bx, t)
-r_t(\bx)
+\gamma\kappa_t\Delta E^{\mathrm{ref}}(\bx).
\end{align*}
When $E^{\mathrm{ref}}=E_\theta(\cdot,T)$, the correction $\Delta E^{\mathrm{ref}}(\bx)$ vanishes and EAT reduces to the MIT; and when $E^{\mathrm{ref}}=E^\star$, the correction closes the terminal energy mismatch and the path terminates at $q_T^\star$. We also introduce a look-ahead map to stabilize target-energy evaluation, with details in App.~\ref{app:lookahead_map}.
Building on the energy-aligned path, we denote the resulting method \textbf{EBFKC on EAT} and summarize it in Alg.~\ref{alg:energy-augmented-vcg-smc}.

\subsection{Connections to Annealed SMC and Energy-Based Modeling}
\label{sec:connections_smc_ebm}

Our method can be viewed as annealed SMC along a diffusion-based probability path. Traditional annealed SMC typically uses a geometric tempering path between the initial and target densities. Its intermediate densities can be evaluated up to normalization, but transporting particles between separated modes can remain difficult~\citep{mate2023learning}. We describe the geometric tempering SMC baseline in App.~\ref{app:tempering_smc} and compare it with EBFKC on EAT in Lennard-Jones systems (Sec.~\ref{sec:particle_systems}).

Existing work replaces the tempering path with a diffusion path, whose progressive Gaussian smoothing eases transport between modes, and uses it to build Boltzmann Generators, i.e., samplers for $p^\star(\bx)\propto\exp(-E^\star(\bx))$ with $E^\star$ given~\citep{phillips2024pdds, grenioux2026diffusion}. These methods use a reweighting strategy similar to that in Thm.~\ref{thm:general_path_fk}, specialized to $\Phi_t=E_\theta(\cdot,t)$, where the architecture of $E_\theta$ enforces $E_\theta(\cdot,T)=E^\star$ by construction. We generalize this to an arbitrary reference energy, sampling from $q_T^{\mathrm{ref}}$ rather than from $p^\star$ alone, and without requiring the terminal constraint to be baked into the architecture.

This perspective also motivates our choice of building energy-based rather than score-based diffusion models.
Reweighting requires evaluating the density along the path, which an energy model provides directly.
With only a score, the density must be recovered by integrating the score for every particle at every step, which is costly and adds numerical error~\citep{sjoberg2026mcmccorr, lam2026metropolis}, making efficient and reliable reweighting impractical.

\section{Experiments}
We evaluate whether EBFKC on EAT recovers the reference target despite pretrained model error, on annealing and reward-tilting tasks for a high-dimensional Gaussian mixture model, particle systems, alanine dipeptide, and alanine tetrapeptide.
We take $E^{\mathrm{ref}}=E^\star$, the scientific setting that motivates our method, for which reference samples are available independently of the pretrained model.
We compare against two baselines that each remove only one failure mode: (1) EBFKC on MIT tracks its path correctly but terminates at the MIT, isolating the endpoint mismatch; (2) FKC on EAT runs FKC with the reward redefined as $\hat r=r-\gamma\Delta E^{\mathrm{ref}}$, which moves its endpoint to $q_T^\star$ (App.~\ref{app:fkc_time_dependent}), isolating the path-tracking error.
Across all systems, only EBFKC on EAT recovers the reference, showing that both failure modes arise in practice and must be addressed. 
To quantify the computational cost of these accuracy gains, we compare runtimes at the reported sampling budgets.
EBFKC on EAT takes approximately $1.1$--$3.6\times$ the runtime of EBFKC on MIT in these comparisons, owing to additional energy and derivative evaluations (App.~\ref{app:gmm_additional_results}--\ref{app:fkc_molecular_free_energy_results}).
Unless marked ``w/o VCG'', all methods use VCG~\citep{ren2025driftlite}, which we found essential for stable weight dynamics in both FKC and EBFKC.
Energy models for particle and molecular systems are trained with Diffusive Classification~\citep{ouyang2026diffusive} (App.~\ref{app:ebm_training}).

\subsection{Gaussian Mixture Model}
We use a 30-dimensional Gaussian mixture model (GMM) to isolate the effect of pretrained model error in a fully controlled setting.
The marginals $\pstar_t(\bx)$ and their energies $E_t^\star(\bx)=-\log \pstar_t(\bx)$ are available in closed form,
which allows us to emulate an imperfect pretrained diffusion model analytically by perturbing the component means and variances.
Varying the perturbation magnitudes allows us to systematically evaluate how sampling quality depends on pretrained model error.
Complete definitions are given in App.~\ref{app:gmm_problem_settings}.

\begin{figure}[t]
    \centering
    \makebox[\textwidth][c]{%
    \begin{minipage}{1.0\textwidth}
    \edef\gmmPanelHeight{\the\dimexpr0.15\textwidth\relax}
    \begin{minipage}[t]{0.69\textwidth}
        \centering
        \includegraphics[height=\gmmPanelHeight]{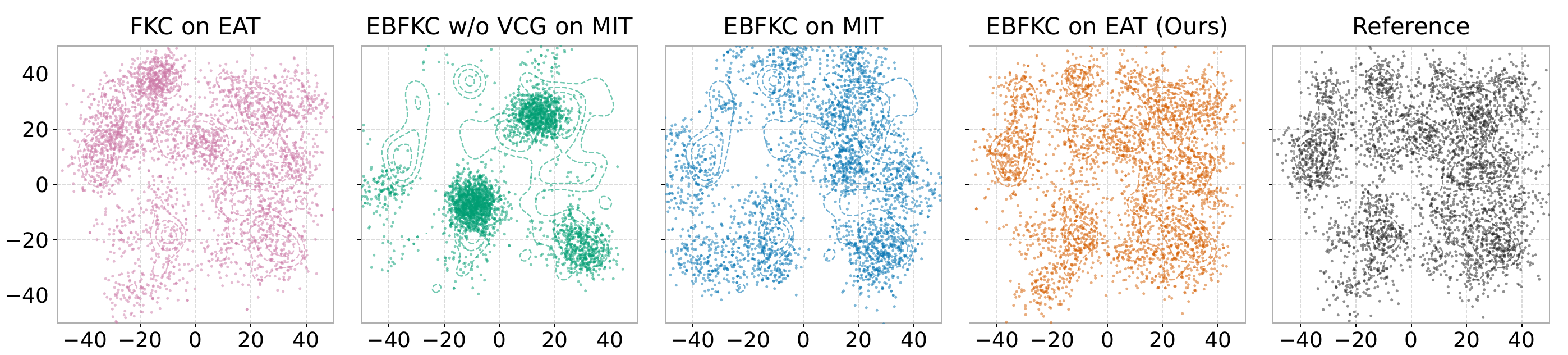}
        \par\small\textbf{(a) Annealing ($\gamma=2$)}
    \end{minipage}
    \hfill
    \begin{minipage}[t]{0.30\textwidth}
        \centering
        \includegraphics[height=\gmmPanelHeight]{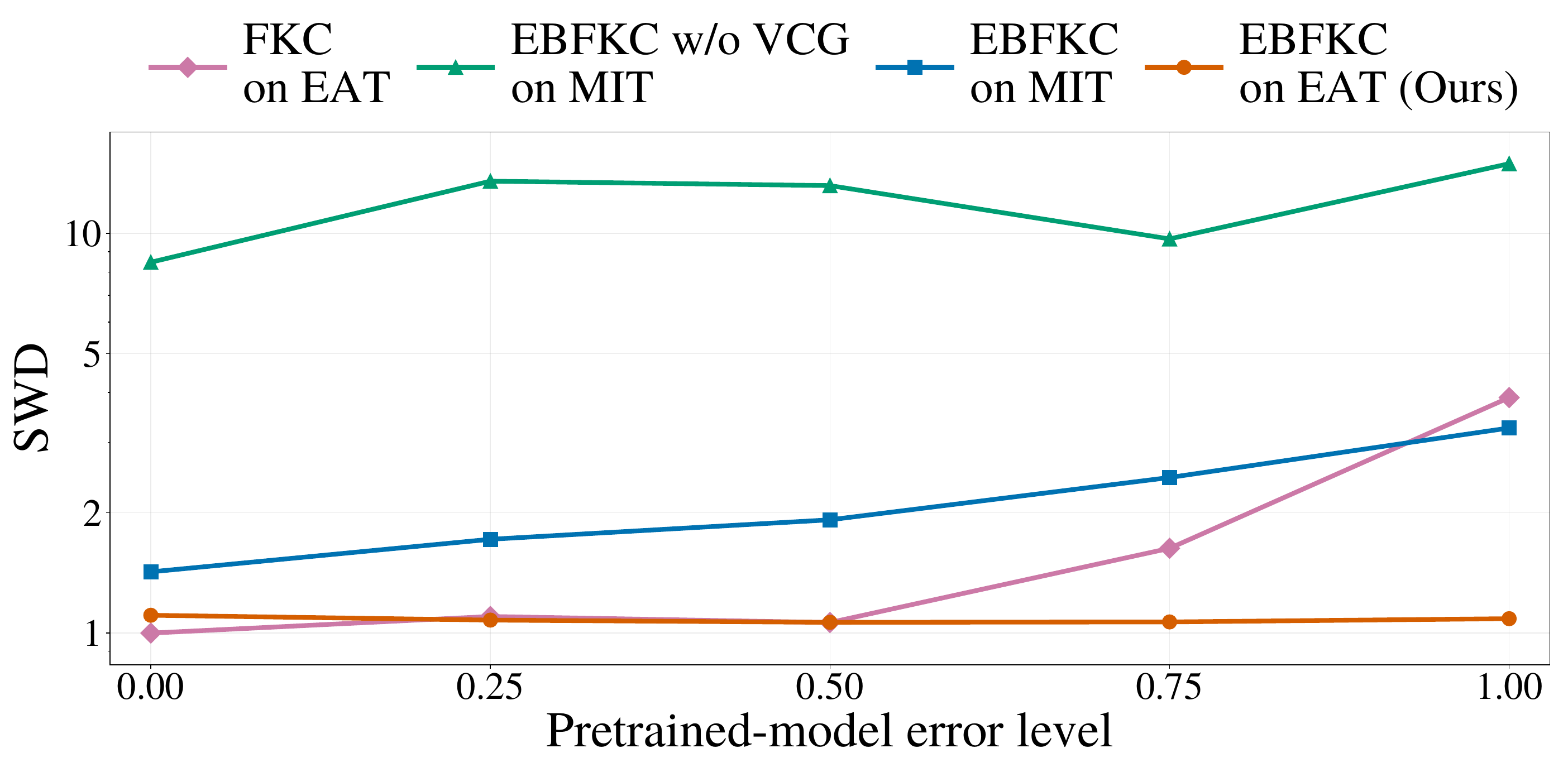}
        \par\small\textbf{(c) Annealing SWD sweep}
    \end{minipage}

    \vspace{0.5em}

    \begin{minipage}[t]{0.69\textwidth}
        \centering
        \includegraphics[height=\gmmPanelHeight]{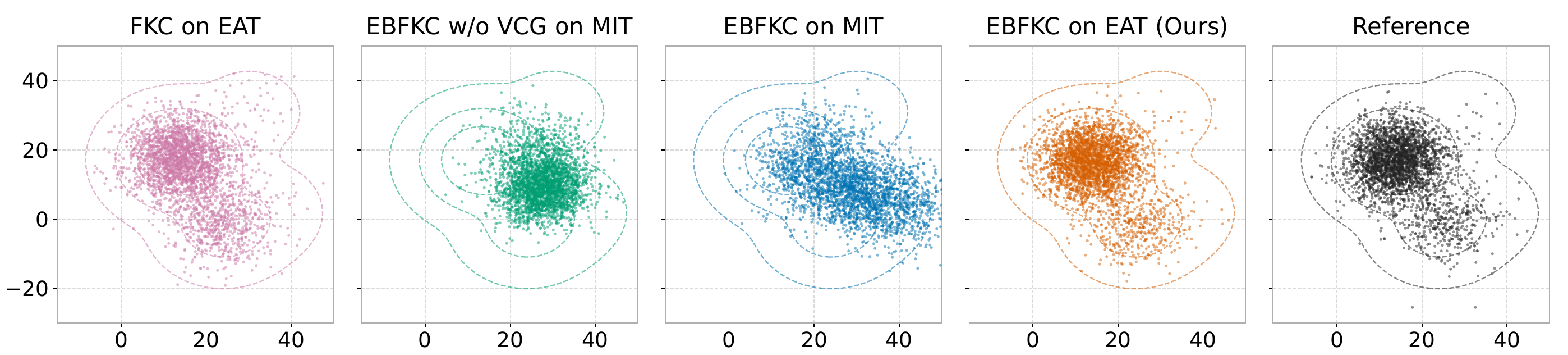}
        \par\small\textbf{(b) Reward tilting ($\sigma=100$)}
    \end{minipage}
    \hfill
    \begin{minipage}[t]{0.30\textwidth}
        \centering
        \includegraphics[height=\gmmPanelHeight]{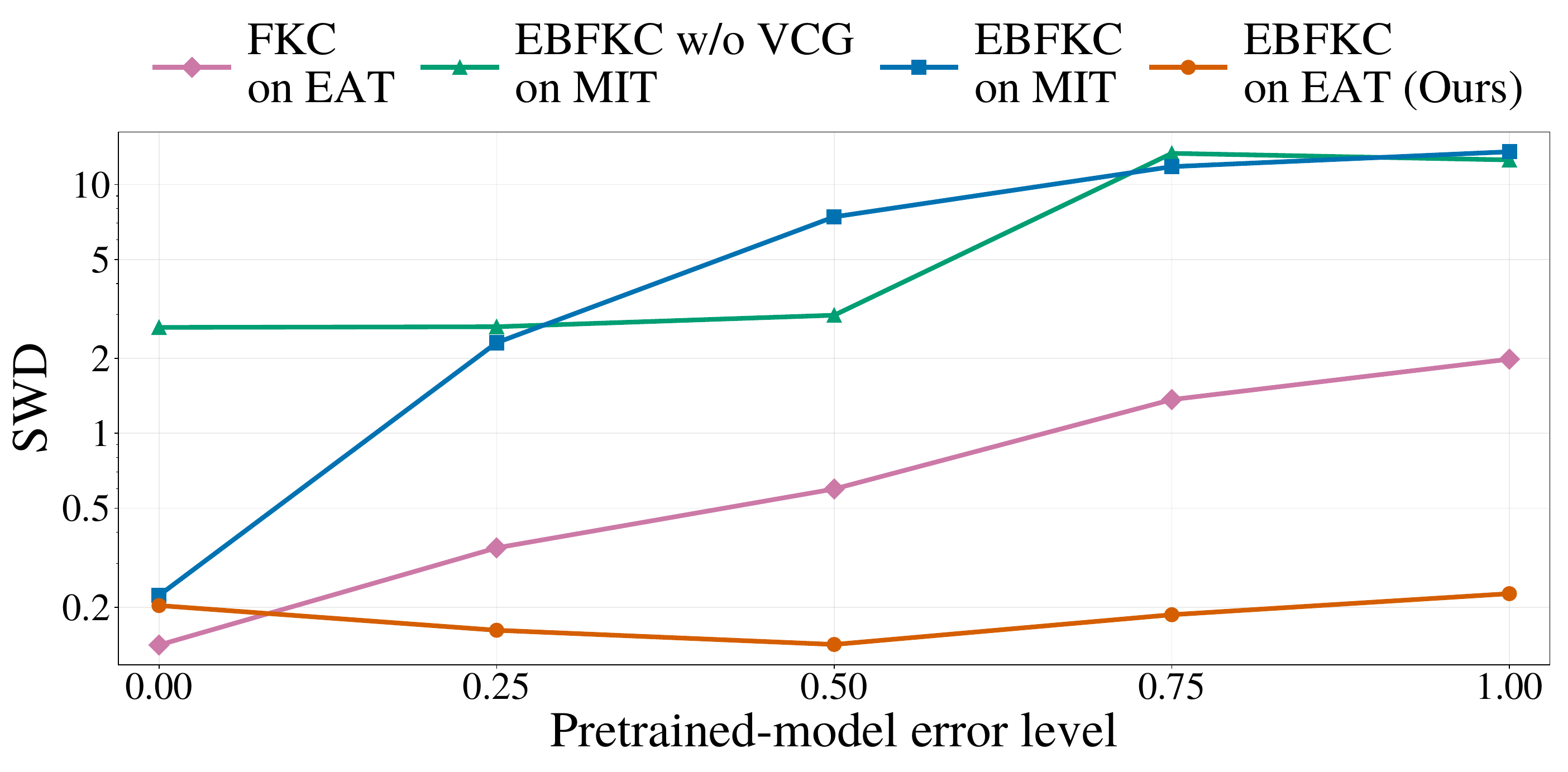}
        \par\small\textbf{(d) Reward-tilting SWD sweep}
    \end{minipage}
    \end{minipage}
    }
    \caption{
    GMM results under controlled pretrained model error.
    (a--b) Sample distributions and (c--d) SWD under increasing pretrained model error
    for annealing and reward tilting.
    }
    \label{fig:gmm_results}
\end{figure}

Both failure modes caused by the imperfect pretrained model appear in the annealing and reward-tilting tasks (Fig.~\ref{fig:gmm_results}(a,b)).
Although the EAT path terminates at the desired target, FKC on EAT remains affected by path-tracking error and fails to sample accurately from that target.
EBFKC on MIT corrects path tracking, but its samples still deviate from the reference because of endpoint mismatch, and without VCG it also suffers from severe weight degeneracy.
Switching from MIT to EAT further removes the endpoint mismatch, allowing EBFKC on EAT to sample accurately from the desired target.

We further evaluate sampling accuracy as the pretrained model error increases.
Fig.~\ref{fig:gmm_results}(c,d) reports the sliced-Wasserstein distance (SWD) for annealing and reward tilting, respectively.
EBFKC on EAT maintains a low SWD across all error levels in both tasks, whereas EBFKC on MIT and FKC on EAT deviate increasingly from the reference as the model error grows.
As shown in App.~\ref{app:gmm_additional_results}, FKC w/o VCG on EAT suffers from severe weight degeneracy, indicating the importance of VCG in stabilizing FKC.
Table~\ref{tab:gmm_results} reports the quantitative comparison at a fixed error level, where EBFKC on EAT has the lowest mean error on all five reported metrics for both tasks.

\begin{table*}[t]
\centering
\caption{
Quantitative comparison on the 30-dimensional GMM under the controlled-error
setting $\sigma_\mu=10$ and $\rho=0.1$. Results are
mean $\pm$ standard deviation over 5 runs.
}
\label{tab:gmm_results}
\scriptsize
\setlength{\tabcolsep}{2.4pt}
\resizebox{\textwidth}{!}{%
\begin{tabular}{llrrrrr}
\toprule
Task & Method & $|\Delta\mathrm{NLL}|$ $\downarrow$ & MMD $\downarrow$ & SWD $\downarrow$
& Mean $L_2$ $\downarrow$ & Cov $F$ $\downarrow$ \\
\midrule
Annealing
& FKC on EAT
& $1.043 \pm 0.196$ & $0.076 \pm 0.007$ & $4.105 \pm 0.375$
& $21.218 \pm 1.462$ & $2437.100 \pm 358.582$ \\
& EBFKC w/o VCG on MIT
& $59.355 \pm 9.014$ & $0.239 \pm 0.035$ & $11.913 \pm 1.374$
& $63.396 \pm 13.044$ & $6770.604 \pm 1682.698$ \\
& EBFKC on MIT
& $59.204 \pm 1.041$ & $0.092 \pm 0.002$ & $3.359 \pm 0.694$
& $9.890 \pm 0.385$ & $1879.349 \pm 129.679$ \\
& EBFKC on EAT (Ours)
& $\boldsymbol{0.133 \pm 0.095}$ & $\boldsymbol{0.022 \pm 0.001}$
& $\boldsymbol{1.110 \pm 0.331}$ & $\boldsymbol{5.493 \pm 0.774}$
& $\boldsymbol{639.193 \pm 27.982}$ \\
\midrule
Reward tilting
& FKC on EAT
& $0.928 \pm 0.374$ & $0.061 \pm 0.019$ & $1.313 \pm 1.191$
& $8.661 \pm 7.779$ & $507.355 \pm 680.777$ \\
& EBFKC w/o VCG on MIT
& $21.575 \pm 10.209$ & $0.435 \pm 0.040$ & $10.136 \pm 6.044$
& $64.321 \pm 29.173$ & $604.058 \pm 878.509$ \\
& EBFKC on MIT
& $22.875 \pm 6.175$ & $0.383 \pm 0.048$ & $10.287 \pm 5.792$
& $63.261 \pm 25.670$ & $1027.872 \pm 1435.608$ \\
& EBFKC on EAT (Ours)
& $\boldsymbol{0.113 \pm 0.016}$ & $\boldsymbol{0.018 \pm 0.008}$
& $\boldsymbol{0.194 \pm 0.102}$ & $\boldsymbol{1.000 \pm 0.684}$
& $\boldsymbol{50.068 \pm 32.847}$ \\
\bottomrule
\end{tabular}
}
\end{table*}
\FloatBarrier

\subsection{Particle Systems}
\label{sec:particle_systems}

We next test whether the correction observed with analytically perturbed
GMMs extends to energy models trained on finite data. We consider the
Double-Well (DW) and Lennard-Jones (LJ) systems, whose time-dependent
energies are no longer available in closed form. We train energy-based
diffusion models with limited training steps to obtain imperfect models
and evaluate their inference-time adaptation to annealing and reward
tilting. Detailed experimental settings are provided in
App.~\ref{app:particle_system_settings}.

\begin{figure}[h]
    \centering
    \begin{minipage}[t]{0.50\textwidth}
        \centering
        \includegraphics[width=\linewidth]
        {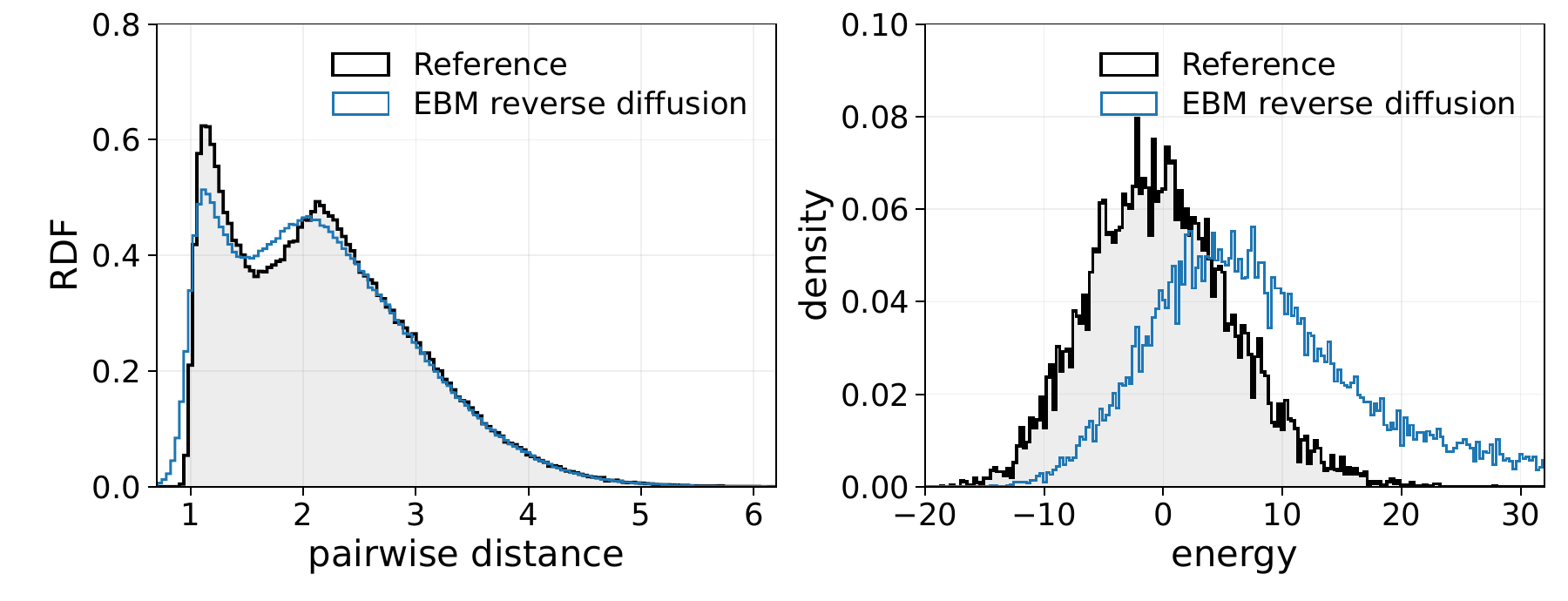}
        \par\textbf{(a) Pretrained model reverse diffusion results}
    \end{minipage}%
    \begin{minipage}[t]{0.50\textwidth}
        \centering
        \includegraphics[width=\linewidth]
        {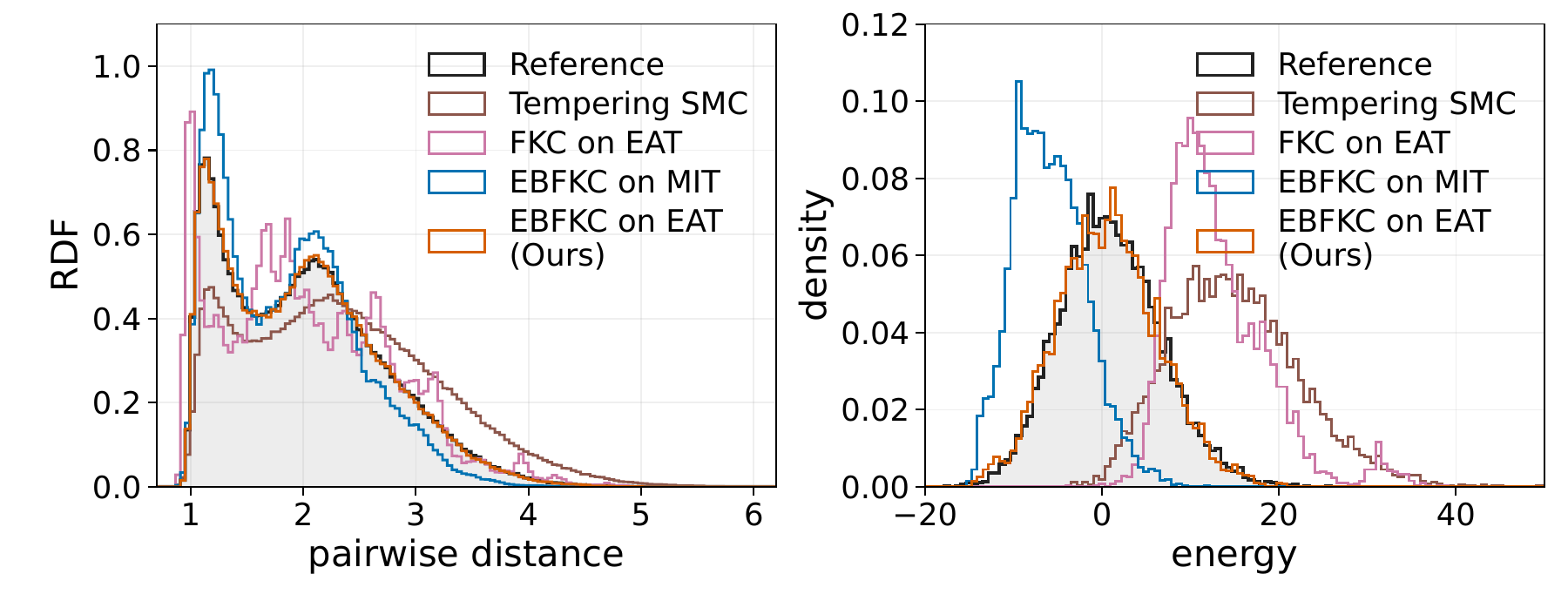}
        \par\textbf{(b) Reward tilting task results}
    \end{minipage}
    \caption{
    LJ results comparing pairwise-distance and energy distributions with the reference. (a) Samples from the pretrained model. (b) Reward-tilting samples from the inference-time methods.
    }
    \label{fig:lj_results}
\end{figure}

\textbf{Double-Well system.}
We first study a 2D 4-particle Double-Well system.
Figures~\ref{fig:dw_annealing} and \ref{fig:dw_reward_tilting} in App.~\ref{app:dw_additional_results} compare the radial distribution functions (RDFs) and energy distributions for annealing and reward tilting.
In both tasks, FKC on EAT and EBFKC on MIT produce visibly shifted RDFs and energy distributions, indicating that these methods retain pretrained model error. In contrast, EBFKC on EAT closely matches the reference pairwise-distance and energy distributions under both sampling objectives.

\textbf{Lennard-Jones system.}
We next consider the more challenging 3D 13-particle LJ system, whose potential has a rugged energy landscape and exhibits singular short-range behavior.
As shown in Fig.~\ref{fig:lj_results}(a), the pretrained model visibly deviates from the reference.
For reward tilting (Fig.~\ref{fig:lj_results}(b)), FKC on EAT and EBFKC on MIT inherit this error, whereas EBFKC on EAT closely matches the reference distributions.
Geometric tempering SMC (App.~\ref{app:tempering_smc}), a standard SMC method working without the pretrained model, also fails to sample the target accurately with the same number of particles, illustrating the benefit of the diffusion path (Sec.~\ref{sec:connections_smc_ebm}).
App.~\ref{app:lj_annealing_results} reports the annealing results.
Table~\ref{tab:particle_results} further confirms that EBFKC on EAT corrects the pretrained model error. 

\subsection{Alanine Dipeptide Annealing}

We next evaluate our method on alanine dipeptide (ALDP) annealing tasks, a standard benchmark for molecular conformational sampling.
Its slow conformational degrees of freedom are characterized by the backbone dihedral angles $\phi$ and $\psi$.
We train an energy-based diffusion model at $800\,\mathrm{K}$ (Fig.~\ref{fig:aldp_800_ebm_phi_psi} in App.~\ref{app:aldp_annealing_results}) and anneal it to $300\,\mathrm{K}$ via inference-time scaling.

Fig.~\ref{fig:aldp_ramachandran} compares the Ramachandran plots of the annealed samples.
EBFKC on EAT recovers both the conformational basins and their relative populations, whereas EBFKC on MIT, with or without VCG, deviates substantially from the $300\,\mathrm{K}$ reference because it anneals toward the MIT rather than the $300\,\mathrm{K}$ Boltzmann distribution.
The marginal distributions of $\phi$ and $\psi$ in Fig.~\ref{fig:aldp_marginals} support this observation.
Table~\ref{tab:aldp_annealing} quantifies the comparison: despite the error of the model pretrained at $800\,\mathrm{K}$, EBFKC on EAT closely matches the reference across all metrics.
FKC was numerically unstable on both paths in our ALDP runs, so we omit it.
The look-ahead ablation study in App.~\ref{app:aldp_annealing_results} shows that removing this map leads to numerical instability, demonstrating its importance in difficult tasks.

\begingroup
\setlength{\intextsep}{6pt}
\begin{figure}[t]
    \centering
    \includegraphics[width=\textwidth]
    {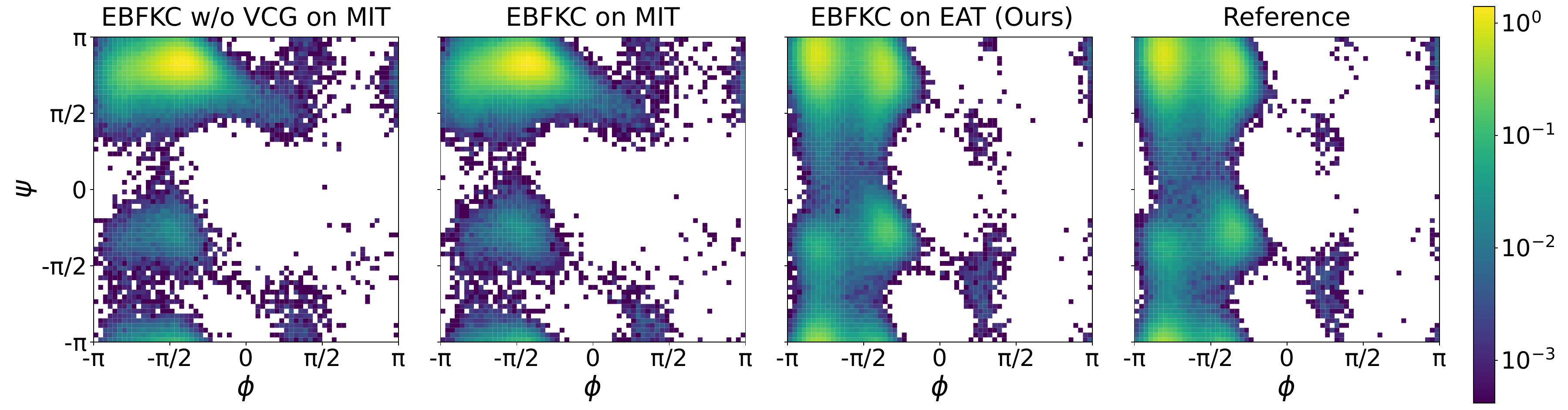}
    \caption{
    Ramachandran plots of the backbone dihedral angles $(\phi,\psi)$ for annealing
    ALDP from $800\,\mathrm{K}$ to $300\,\mathrm{K}$.
    Reference samples at $300\,\mathrm{K}$ are compared with samples generated by EBFKC on MIT with/without VCG and EBFKC on EAT using the same pretrained diffusion model at $800\,\mathrm{K}$.
    }
    \label{fig:aldp_ramachandran}
\end{figure}

\begin{table}[h]
\centering
\caption{
Quantitative ALDP annealing results from $800\,\mathrm{K}$ to $300\,\mathrm{K}$; the first row evaluates the pretrained model at $800\,\mathrm{K}$ against the $800\,\mathrm{K}$ reference.
Results are mean $\pm$ std over 5 runs.
}
\label{tab:aldp_annealing}
\small
\setlength{\tabcolsep}{6pt}
\resizebox{\linewidth}{!}{%
\begin{tabular}{lcrrrr}
\toprule
Methods & Temperature (K) & $(\phi,\psi)$ TVD $\downarrow$ & Basin-pop. error $\downarrow$
& Energy TVD $\downarrow$ & Distance TVD $\downarrow$ \\
\midrule
Pretrained reverse diffusion & $800$ & $0.3005 \pm 0.0022$ & $0.1699 \pm 0.0016$ & $0.6730 \pm 0.0013$ & $0.0253 \pm 0.0001$ \\
\midrule
EBFKC w/o VCG on MIT & $300$ & $0.5089 \pm 0.0163$ & $0.1643 \pm 0.0037$ & $0.4460 \pm 0.0130$ & $0.0366 \pm 0.0008$ \\
EBFKC on MIT & $300$ & $0.5063 \pm 0.0035$ & $0.1657 \pm 0.0026$ & $0.4659 \pm 0.0095$ & $0.0381 \pm 0.0006$ \\
EBFKC on EAT (Ours) & $300$ & $\mathbf{0.0855 \pm 0.0031}$ & $\mathbf{0.0047 \pm 0.0041}$ & $\mathbf{0.0240 \pm 0.0028}$ & $\mathbf{0.0030 \pm 0.0002}$ \\
\bottomrule
\end{tabular}
}
\end{table}
\endgroup

\subsection{Free-Energy Estimation}

Free-energy profiles characterize the relative thermodynamic stability of molecular conformations along selected reaction coordinates.
Their accurate estimation remains a rare-event sampling problem because low-probability conformations are infrequently observed even when a pretrained equilibrium sampler can already generate independent configurations.

\begin{figure}[!htbp]
    \centering
    \begin{minipage}[t]{0.4\textwidth}
        \centering
        \includegraphics[width=0.94\linewidth]
        {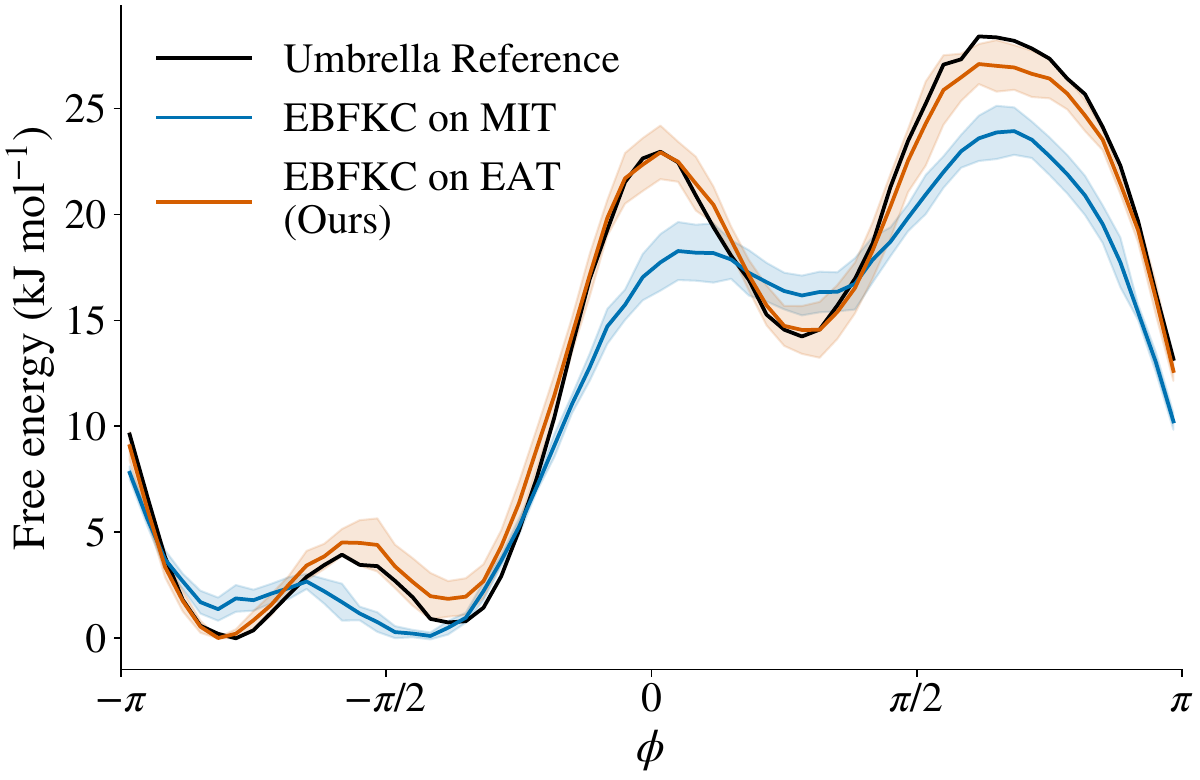}
        \par\small\textbf{(a) ALDP free-energy profile}
    \end{minipage}
    \hspace{0.02\textwidth}
    \begin{minipage}[t]{0.4\textwidth}
        \centering
        \includegraphics[width=0.94\linewidth]
        {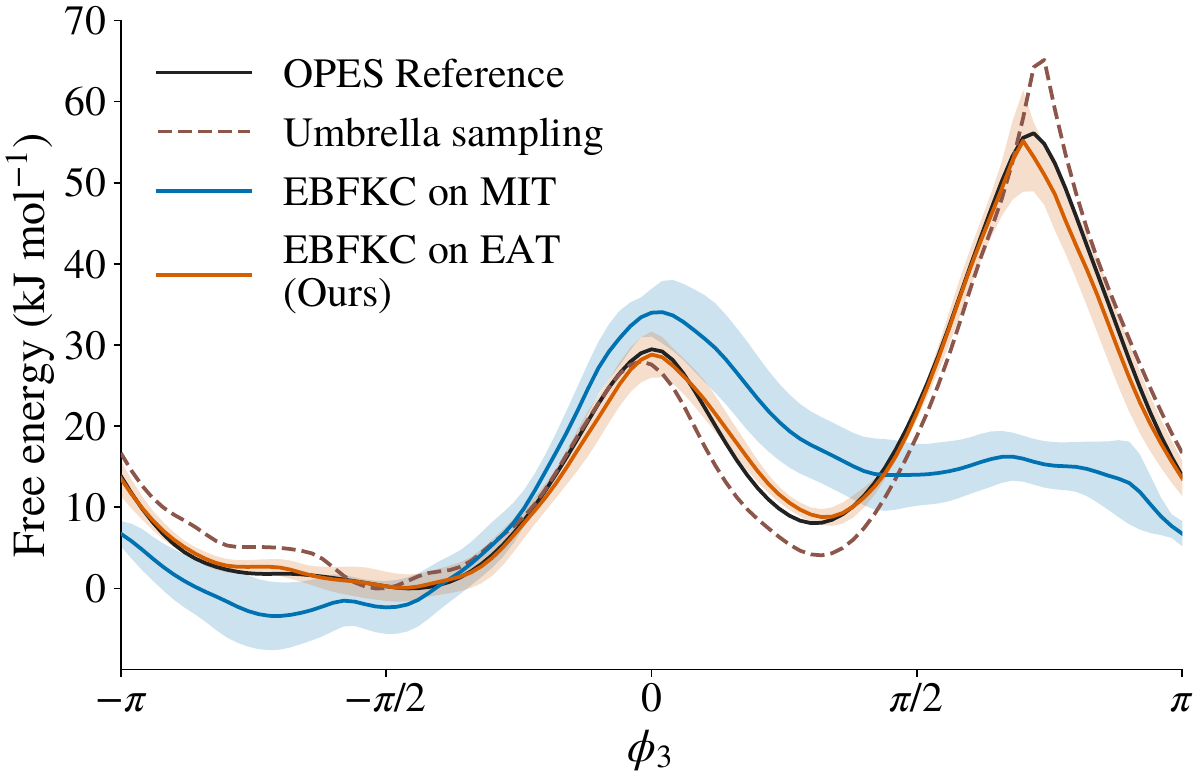}
        \par\small\textbf{(b) Ala-4 free-energy profile}
    \end{minipage}
    \caption{
    One-dimensional molecular free-energy profiles at
    $300\,\mathrm{K}$. (a) ALDP along $\phi$. (b) Ala-4 along $\phi_3$.
    The references are umbrella sampling for ALDP and OPES for Ala-4.  Diffusion-based results show the mean and standard deviation (shaded) over five independent runs.
    }
    \label{fig:molecular_free_energy}
\end{figure}

Motivated by this challenge, we follow UmbrellaDiff~\citep{xie2026enhanced} to estimate one-dimensional free-energy profiles at $300\,\mathrm{K}$ for ALDP along $\phi$ and for the 42-atom alanine tetrapeptide (Ala-4) along its third backbone dihedral $\phi_3$, starting from pretrained diffusion equilibrium samplers (App.~\ref{app:fkc_molecular_free_energy_results}).
Each umbrella window defines a reward-tilting target with $E^{\mathrm{ref}}=E^\star$, which we sample with EBFKC on EAT or EBFKC on MIT, and the weighted particles from all windows are combined with the weighted multistate Bennett acceptance ratio (MBAR)~\citep{shirts2008statistically,xie2026enhanced}.

\textbf{Alanine dipeptide.}
As shown in Fig.~\ref{fig:molecular_free_energy}(a), EBFKC on MIT underestimates the principal barriers because its window-wise samples retain pretrained model error.
In contrast, EBFKC on EAT closely follows the umbrella sampling reference~\citep{torrie1977nonphysical} and reproduces its principal barriers and basin structure, indicating that it samples the reward-tilted target in each umbrella window accurately.

\textbf{Alanine tetrapeptide.}
Ala-4 poses a high-barrier sampling problem: its three slow backbone dihedrals partition the conformational space into eight metastable basins (settings in App.~\ref{app:alanine_tetrapeptide_settings}).
Fig.~\ref{fig:molecular_free_energy}(b) uses as reference OPES~\citep{invernizzi2020rethinking}, which biases all three slow dihedrals $\phi_1$, $\phi_2$, and $\phi_3$.
Traditional umbrella sampling, which biases only $\phi_3$, becomes trapped along the hidden slow variables and yields an inaccurate profile.
EBFKC on EAT, although it also biases only $\phi_3$, closely reconstructs the OPES basin-and-barrier structure, whereas EBFKC on MIT retains the pretrained model error and severely distorts the barrier heights.
Without VCG, the profile of EBFKC on MIT is distorted even further (App.~\ref{app:fkc_molecular_free_energy_results}).

\FloatBarrier
\section{Conclusion}

We introduced the \emph{Energy-based Feynman--Kac Corrector (EBFKC)}, a framework for inference-time scaling with imperfect pretrained energy-based diffusion models given a reference energy.
EBFKC derives Feynman--Kac dynamics that track any prescribed path exactly at the population level and simulates them with SMC, using VCG to mitigate weight degeneracy, thereby removing the path-tracking error.
The energy-aligned target (EAT) uses the pretrained energy as a surrogate along the path and progressively incorporates the reference energy, removing the endpoint mismatch.
Across Gaussian mixture models, particle systems, and molecular systems, both ingredients prove necessary, and EBFKC on EAT recovers the target distributions under annealing and reward tilting and yields accurate molecular free-energy profiles.
Future directions include extending the EBFKC framework to discrete diffusion models~\citep{hasan2026discrete,ren2026fluxlite}, adapting the correction schedules automatically, which we currently design by hand for the high-barrier alanine tetrapeptide system, reducing derivative-evaluation costs, and correcting large pretrained equilibrium samplers for more reliable inference-time scaling in challenging real-world applications~\citep{lewis2025scalable}.

\subsection*{AI use statement}

In this work, we have not used generative AI tools for any tasks with required disclosure. We used generative AI tools to polish the writing and to assist with coding. We have reviewed all AI-assisted work: text edits were checked by the authors, and AI-assisted code was reviewed and tested for correctness. We take responsibility for the final content of this work, including text, claims or artifacts produced with the aid of generative AI.

\subsection*{Reproducibility Statement}

We provide the necessary details to facilitate reproduction of our results. 
The proposed Energy-based Feynman-Kac Corrector is described in detail in the main text, with complete algorithmic specifications and implementation details provided in the appendix. 
The assumptions and derivations underlying our theoretical results, together with complete proofs, are also included in the appendix. 
For all experiments, we report the experimental setups, model architectures, training procedures, sampling configurations, evaluation metrics, and hyperparameters, as well as details of the datasets and target distributions used.
The code will be released later.

\FloatBarrier
{\providecommand{\urlprefix}{}\renewcommand{\url}[1]{}
\bibliography{references}
}
\bibliographystyle{iclr2027_conference}

\appendix
\section{Energy-Based Diffusion Model Training}
\label{app:ebm_training}

Following Diffusive Classification~\citep{ouyang2026diffusive}, we train the
time-dependent energy model $E_\theta$ in up to two stages: Stage~1 relies on
denoising score matching, and an optional Stage~2 further calibrates the
model through noise-level classification.

The forward SDE~\eqref{eq:forward_noising_process} admits the closed-form
solution
\begin{align*}
    \by_s
    =
    \underbrace{\exp\left(-\int_0^s F_u \rmd u\right)}_{=: S(s)} \by_0
    +
    \underbrace{\exp\left(-\int_0^s F_u \rmd u\right) \sqrt{\int_0^s \beta^2_u \exp\left(2\int_0^u F_r \rmd r\right) \rmd u}}_{=: \nu(s)} \bz,
\end{align*}
where $\by_0 \sim \pstar$ and $\bz \sim \mathcal{N}(0, \mathrm{I}_d)$. We write
$\pstar_{t|T}(\cdot \mid \bx_T) = \mathcal{N}\!\left(S(T-t)\bx_T,\, \nu(T-t)^2 \mathrm{I}_d\right)$
for the resulting law of $\bx_t$ conditioned on $\bx_T$. 

Tweedie's formula then relates the score of the marginal to the score of this
conditional,
\begin{align}\label{eq:tweedie}
    \nabla \log \pstar_t(\bx) = \mathbb{E}\left[\nabla_{\bx} \log \pstar_{t|T}(\bx \mid \bx_T) \,\middle|\, \bx_t = \bx\right],
\end{align}
which motivates the Stage~1 denoising score matching objective
\begin{align*}
    \mathcal{L}_{\mathrm{DSM}}(\theta)
    =
    \mathbb{E}\left[
    \lambda(t)
    \left\|
    -\nabla_{\bx}E_\theta(\bx_t,t)
    -
    \nabla_{\bx_t}
    \log \pstar_{t|T}(\bx_t\mid\bx_T)
    \right\|^2
    \right],
\end{align*}
training the model to match the spatial gradient of the energy at each noise
level. Tweedie's formula~\eqref{eq:tweedie} also gives a direct link between
the denoiser and the score,
\begin{align}
    &D_t^\star(\bx)
    = \mathbb{E}[\bx_T \mid \bx_t = \bx]
    = \frac{\bx+\nu(T-t)^2 \nabla_{\bx} \log \pstar_t(\bx)}{S(T-t)}, \nonumber \\
    & D_t(\bx)
    = \frac{\bx-\nu(T-t)^2 \nabla_{\bx} E_\theta(\bx,t)}{S(T-t)}, \label{eq:denoiser_score}
\end{align}
where $D_t$ is the model denoiser obtained by replacing the true score with $-\nabla_{\bx}E_\theta$.

Stage~1 alone, however, only matches local properties of the energy
landscape: it misses the relative proportions between disjoint regions of
the energy (\citealp{Wenliang2021blindness}; \citealp[Appendix~C]{ouyang2026diffusive}).
Stage~2 corrects for this through noise-level classification: we select
noise levels $t_{1:K}$ and train the model to identify the level from which
a perturbed sample was drawn. The resulting classifier is
\begin{align*}
    p_\theta(c=i\mid\bx)
    =
    \frac{\exp\left(-E_\theta(\bx,t_i)\right)}
    {\sum_{j=1}^{K}\exp\left(-E_\theta(\bx,t_j)\right)},
\end{align*}
trained with the classification loss
\begin{align*}
    \mathcal{L}_{\mathrm{DiffCLF}}(\theta)
    =
    -\frac{1}{K}
    \sum_{i=1}^{K}
    \mathbb{E}_{\bx_{t_i}\sim p_{t_i}}
    \left[
    \log p_\theta(c=i\mid\bx_{t_i})
    \right].
\end{align*}
The complete Stage~2 training objective is then
\begin{align*}
    \mathcal{L}_{\mathrm{EBM}}(\theta)
    =
    \mathcal{L}_{\mathrm{DSM}}(\theta)
    +
    \lambda_{\mathrm{clf}}\,\mathcal{L}_{\mathrm{DiffCLF}}(\theta).
\end{align*}

\paragraph{Preconditioned parameterization.}
We follow the preconditioning scheme of~\citet{ouyang2026diffusive, thornton2025composition}.
Let $\Gamma(t)=\nu(T-t)/S(T-t)$ denote the noise-to-signal ratio, and let
$\mu_{\mathrm{data}}$ and $\sigma_{\mathrm{data}}^2$ denote the empirical
mean and scalar variance of the training data. We define the input scale
\begin{align*}
\alpha_{\mathrm{in}}(t)=\sqrt{S(T-t)^2\sigma_{\mathrm{data}}^2+\nu(T-t)^2}
\end{align*}
and input center $\beta_{\mathrm{in}}(t)=S(T-t)\mu_{\mathrm{data}}$, and
parameterize the energy as
\begin{align*}
E_\theta(\by,t)
={}&
\frac{\lVert \by\rVert^2}
{2\alpha_{\mathrm{in}}(t)^2}
-
\frac{\sigma_{\mathrm{data}}}{\Gamma(t)}
U_t^\theta\!\left(
\frac{\by-\beta_{\mathrm{in}}(t)}
{\alpha_{\mathrm{in}}(t)}
\right)
\nonumber\\
&-
\frac{
S(T-t)\mu_{\mathrm{data}}^\top \by
}{
\alpha_{\mathrm{in}}(t)^2
}
+
\frac{1}{2}
\frac{S(T-t)^2}{\alpha_{\mathrm{in}}(t)^2}
\lVert\mu_{\mathrm{data}}\rVert^2
\nonumber\\
&+
\frac{d}{2}
\log\!\left(
2\pi\alpha_{\mathrm{in}}(t)^2
\right).
\end{align*}
With this parameterization, the network only needs to learn a non-Gaussian
residual in standardized coordinates, while the dominant Gaussian scale and
normalization constant are represented analytically. Writing the
standardized input as $\bz=(\by-\beta_{\mathrm{in}}(t))/\alpha_{\mathrm{in}}(t)$,
we construct the scalar residual $U_t^\theta$ from a vector-valued network
$\mathrm{NN}_t^\theta$ via
\begin{align*}
    U_t^\theta(\bz)
    =
    \bz^\top\mathrm{NN}_t^\theta(\bz).
\end{align*}
We parameterize $\mathrm{NN}_t^\theta$ with an $E(n)$-equivariant graph
neural network (EGNN), using sinusoidal time embeddings together with
learnable bond- and atom-type embeddings.

\section{Algorithm Details}

\subsection{Fokker--Planck Equation for a Prescribed Path}
\label{app:general_path_fk_derivation}

We prove Thm.~\ref{thm:general_path_fk} here. Suppose that $q_t$
satisfies the Feynman--Kac-type Fokker--Planck equation
\begin{align}
\partial_t q_t(\bx)
={}&
-\nabla\cdot
\left[
q_t(\bx)v_t(\bx)
\right]
+
\frac{\sigma_t^2}{2}\Delta q_t(\bx)
+
q_t(\bx)g_t(\bx).
\label{eq:app_fk_fpe}
\end{align}
Dividing Equation~\eqref{eq:app_fk_fpe} by $q_t(\bx)$ and
using $\nabla\cdot(q_t v_t)/q_t=\nabla\cdot v_t+
v_t^\top\nabla\log q_t$ and $\Delta q_t/q_t=
\Delta\log q_t+\|\nabla\log q_t\|^2$ gives the reweighting potential
\begin{align}
g_t(\bx)
={}&
\partial_t\log q_t(\bx)
+
\nabla\cdot v_t(\bx)
+
v_t(\bx)^\top\nabla\log q_t(\bx)
\nonumber\\
&-
\frac{\sigma_t^2}{2}
\left[
\Delta\log q_t(\bx)
+
\left\|\nabla\log q_t(\bx)\right\|^2
\right].
\label{eq:app_general_g}
\end{align}

For the prescribed path
$q_t(\bx)=Z_t^{-1}\exp\left(-\Phi_t(\bx)\right)$,
its log-density derivatives satisfy $\partial_t\log q_t=
-\partial_t\Phi_t-\partial_t\log Z_t$, $\nabla\log q_t=
-\nabla\Phi_t$, and $\Delta\log q_t=-\Delta\Phi_t$. Because
$\partial_t\log Z_t$ is state independent, define the uncentered
potential as
$$G_t=g_t+\partial_t\log Z_t.$$
We then substitute the
proposal drift $v_t=-\widetilde{u}_t(\bx)-\sigma_t^2\gradPhit$ into
Equation~\eqref{eq:app_general_g}. Expanding the divergence and
inner-product terms yields
\begin{align*}
G_t(\bx)
={}&
-\partial_t\Phi_t(\bx)
-
\nabla\cdot \widetilde{u}_t(\bx)
+
\widetilde{u}_t(\bx)^\top
\gradPhit
\nonumber\\
&-
\frac{\sigma_t^2}{2}
\Delta\Phi_t(\bx)
+
\frac{\sigma_t^2}{2}
\left\|\gradPhit\right\|^2.
\end{align*}
This is the uncentered potential in
Equation~\eqref{eq:general_path_fk_potential}.

It remains to determine the state-independent normalization term.
Assuming the boundary flux vanishes, integrating
Equation~\eqref{eq:app_fk_fpe} over $\bx$ and using
$\int q_t(\bx)\,\rmd\bx=1$ gives
$\Eqt{g_t}=0$. Since $g_t=G_t-\partial_t\log Z_t$, the
normalization term is $\partial_t\log Z_t=\Eqt{G_t}$.
Consequently, the normalized reweighting potential is
\begin{align*}
g_t(\bx)
=
G_t(\bx)
-
\Eqt{G_t},
\end{align*}
which proves Thm.~\ref{thm:general_path_fk}.

\subsection{Variance-Controlling Guidance}
\label{sec:vcg_prescribed_path}

The Feynman--Kac dynamics in Eq.~\eqref{eq:fk fpe} preserve the prescribed path at the population level, but a finite-particle SMC approximation is prone to weight degeneracy when the reweighting potential has high variance.
Variance-Controlling Guidance (VCG) \citep{ren2025driftlite} exploits the degree of freedom in the Fokker--Planck equation to actively reduce this variance. Adding a control drift $\bb_t$ gives an equivalent Fokker--Planck equation:
\begin{align*}
\partial_t q_t(\bx)
={}&
-\nabla\cdot
\left[
\left(v_t(\bx)+\bb_t(\bx)\right)
q_t(\bx)
\right]
+
\frac{\sigma_t^2}{2}\Delta q_t(\bx)
+
q_t(\bx)\phi_t(\bx),
\end{align*}
where the residual potential is
$
\phi_t(\bx)
=
g_t(\bx)
+
h_t(\bx;\bb_t),
$
and the control potential is
$
h_t(\bx;\bb_t)
=
-\gradPhit\cdot\bb_t(\bx)
+
\nabla\cdot\bb_t(\bx).
$
VCG selects the control drift by directly minimizing the variance of the residual potential:
\begin{align}
\min_{\bb_t}\;
\Varqt{\phi_t(\bx)}
=
\Varqt{g_t(\bx)+h_t(\bx;\bb_t)}.
\label{VCG}
\end{align}
To obtain a computationally lightweight control, we restrict the drift to the linear span of $n$ problem-informed basis:
$\bb_t(\bx)= \sum_{i=1}^n\lambda_t^i\boldsymbol{\psi}_i(\bx,t)$.
Let $h_t^i(\bx)=h_t(\bx;\boldsymbol{\psi}_i)$. Under this linear ansatz, Eq.~\eqref{VCG} reduces to the $n\times n$ system $\mathbf{A}_t\boldsymbol{\lambda}_t=\mathbf{c}_t$, where $
A_{ij}
=
\Eqt{h_t^i h_t^j}$ and $
c_i
=
-\Eqt{g_t h_t^i}. $
Solving this system at each time step yields the control drift that minimizes the residual potential variance within the chosen linear span.
In practice, $\mathbf{A}_t$ may be ill-conditioned, so we solve the regularized system
\begin{equation}
(\mathbf{A}_t+\lambda_{\mathrm{VCG}}\mathbf{I}_n)\boldsymbol{\lambda}_t=\mathbf{c}_t.
\label{eq:vcg-linear-system}
\end{equation}

For the EAT path, we use the following problem-informed basis:
\begin{align*}
\boldsymbol{\psi}_1
\!&=\bx,\enspace
\boldsymbol{\psi}_2
\!=\nabla_{\bx} E_\theta(\bx, t),\enspace
\boldsymbol{\psi}_3
\!=\nabla_{\bx}[\Estar{\LAmap{\bx}}-E_\theta(\LAmap{\bx},T)],\enspace
\boldsymbol{\psi}_4
\!=\nabla_{\bx}r\left(\LAmap{\bx}\right).
\end{align*}
The identity basis $\boldsymbol{\psi}_1$ and learned-energy basis
$\boldsymbol{\psi}_2$ are always active. We activate the target-energy correction basis
$\boldsymbol{\psi}_3$ when $\kappa_t\neq0$, and use the reward basis
$\boldsymbol{\psi}_4$ for reward-tilting tasks. Compared with
DriftLite~\citep{ren2025driftlite}, we add the target-energy correction basis $\boldsymbol{\psi}_3$,
which is significantly important in control drift estimation. At each time, we retain
the active basis, evaluate
$h_t^i=h_t(\cdot;\boldsymbol{\psi}_i)$ on the current weighted particles,
form weighted empirical estimates of $\mathbf{A}_t$ and
$\mathbf{c}_t$, and solve the empirical version of the regularized system in Eq.~\eqref{eq:vcg-linear-system}.

\subsection{Look-ahead Map}
\label{app:lookahead_map}
The EAT path evaluates the reference and learned terminal energies, while both MIT and EAT paths evaluate the reward, at the current state $\bx_t$.
At high noise levels, these states can lie far from the data manifold, where evaluating $E^\star$ and its gradient can be numerically unstable.
Therefore, in the challenging LJ-13 and molecular sampling tasks we evaluate these quantities at a look-ahead point defined by the map
\begin{align}
\LAmap{\bx}=(1-\omega_t)\,\bx+\omega_t\,D_t(\bx),
\label{eq:lookahead_map}
\end{align}
which interpolates between $\bx$ and the denoiser prediction $D_t(\bx)$ of Eq.~\eqref{eq:denoiser_score} with strength $\omega_t\in[0,1]$.
Flow maps \citep{sabour2025test} could play the same role, but the denoiser is directly available from the learned energy.
Two schedules govern the correction: the look-ahead strength $\omega_t$ and the correction schedule $\kappa_t$.
Since the denoiser is weakly informative early in the reverse process, we activate the correction only later; over its active window $\omega_t$ decreases from the denoiser toward the identity, and $\omega_T=0$ gives $\mathcal{D}_T(\bx)=\bx$.
Replacing the terminal-time evaluations by their look-ahead counterparts gives the practical potentials
\begin{align*}
\Phi_t^{\mathrm{MIT,LA}}(\bx)
&=\gamma E_\theta(\bx, t)-r_t(\LAmap{\bx}),\\
\Phi_t^{\mathrm{EAT,LA}}(\bx)
&=\gamma E_\theta(\bx, t)-r_t(\LAmap{\bx})+\gamma\kappa_t\,\Delta E^{\mathrm{ref}}(\LAmap{\bx}).
\end{align*}
Because $\omega_T=0$ and $\kappa_T=1$, the terminal potential is still $\gamma E^{\mathrm{ref}}(\bx)-r(\bx)$, so the look-ahead map alters only the intermediate marginals, not the endpoint.
An imperfect denoiser can still send a few particles to non-physical configurations, so we apply the alive filter (App.~\ref{app:alive_filter}) for numerical stability.

\subsection{Derivative Computation and Stochastic Laplacian Estimation}
\label{app:stochastic_laplacian}

\paragraph{Hutchinson trace estimation.}
We use Hutchinson trace estimator~\citep{hutchinson1989stochastic}
to estimate the Laplacian term. Let
$H_t(\bx)=\nabla_{\bx}^2\Phi_t(\bx)$,
and let $\boldsymbol{\epsilon}\in\R^d$ be a random probe satisfying
$\mathbb{E}[\boldsymbol{\epsilon}\boldsymbol{\epsilon}^{\top}]
=\mathbf{I}_d$. Since $\Delta\Phi_t=\operatorname{tr}(H_t)$, the
Hutchinson identity gives
\begin{align*}
\Delta\Phi_t(\bx)
=
\mathbb{E}_{\boldsymbol{\epsilon}}
\left[
\boldsymbol{\epsilon}^{\top}
H_t(\bx)
\boldsymbol{\epsilon}
\right].
\end{align*}
Given $R$ independent probes, we evaluate
$\ell_t^{(r)}(\bx)=(\boldsymbol{\epsilon}^{(r)})^{\top}
H_t(\bx)\boldsymbol{\epsilon}^{(r)}$ and average them as
$\Lhat_t(\bx)=R^{-1}\sum_{r=1}^{R}
\ell_t^{(r)}(\bx)$.
Both Rademacher and standard Gaussian probes satisfy the required
second-moment condition. For the symmetric Hessian
$H_t(\bx)$, the Gaussian estimator additionally contains the
contribution from the Hessian diagonal:
\begin{align*}
\operatorname{Var}_{\mathrm{Gauss}}
\left[\Lhat_t(\bx)\right]
-
\operatorname{Var}_{\mathrm{Rad}}
\left[\Lhat_t(\bx)\right]
=
\frac{2}{R}
\sum_i
\left[H_t(\bx)\right]_{ii}^{2}.
\end{align*}
Thus, Rademacher probes have no greater variance than Gaussian probes,
and the inequality is strict whenever the Hessian has a nonzero
diagonal entry. We therefore use Rademacher probes by default for the
stochastic estimation of all Laplacian terms.

\paragraph{Derivatives and matrix-free Hessian--vector products.}
We first detail how to compute derivatives of functions composed with
the look-ahead map in Eq.~\eqref{eq:lookahead_map}.
For any differentiable function $f:\R^d\to\R$, define the
look-ahead-composed function
\begin{align*}
F_t(\bx)=f\left(\LAmap{\bx}\right).
\end{align*}
Let $J_{\mathcal{D}_t}(\bx)$ denote the Jacobian of
$\mathcal{D}_t$ with respect to $\bx$. The spatial gradient and time
derivative of $F_t$ follow from the chain rule:
\begin{align*}
\nabla_{\bx}F_t(\bx)
&=
J_{\mathcal{D}_t}(\bx)^{\top}
\nabla f\left(\LAmap{\bx}\right),
\\
\partial_tF_t(\bx)
&=
\nabla f\left(\LAmap{\bx}\right)^{\top}
\partial_t\mathcal{D}_t(\bx).
\end{align*}
For the look-ahead map in Eq.~\eqref{eq:lookahead_map},
these two derivatives of the map are
\begin{align*}
J_{\mathcal{D}_t}(\bx)
&=
\left(1-\omega_t\right)\mathbf{I}_d
+\omega_tJ_{D_t}(\bx),
\\
\partial_t\mathcal{D}_t(\bx)
&=
\dot{\omega}_t\left(D_t(\bx)-\bx\right)
+\omega_t\partial_tD_t(\bx).
\end{align*}
The Hessian of the composite function contains contributions from
both $f$ and the look-ahead map:
\begin{align*}
\nabla_{\bx}^2F_t(\bx)
={}&
J_{\mathcal{D}_t}(\bx)^{\top}
\nabla^2f\left(\LAmap{\bx}\right)
J_{\mathcal{D}_t}(\bx)
\\
&+
\sum_{j=1}^d
\left[\nabla f\left(\LAmap{\bx}\right)\right]_j
\nabla_{\bx}^2\left[\mathcal{D}_t(\bx)\right]_j.
\end{align*}
We evaluate the required Hessian-vector product (HVP) without
explicitly forming the Hessian.
For the learned intermediate energy, second-order automatic
differentiation directly computes
\begin{align*}
\nabla_{\bx}^2 E_\theta(\bx,t)\boldsymbol{\epsilon}
=
\nabla_{\bx}
\left[
\left(\nabla_{\bx}E_\theta(\bx,t)\right)^\top
\boldsymbol{\epsilon}
\right].
\end{align*}
For a look-ahead-composed term $F_t$, we instead approximate the HVP
by applying a centered finite difference to the complete spatial
gradient:
\begin{align}
\nabla_{\bx}^2F_t(\bx)\boldsymbol{\epsilon}
\approx{}&
\frac{
\nabla_{\bx}F_t(\bx+\delta\boldsymbol{\epsilon})
-\nabla_{\bx}F_t(\bx-\delta\boldsymbol{\epsilon})
}{2\delta}
\nonumber\\
={}&
\frac{1}{2\delta}
\bigg[
J_{\mathcal{D}_t}(\bx+\delta\boldsymbol{\epsilon})^{\top}
\nabla f\left(\mathcal{D}_t(\bx+\delta\boldsymbol{\epsilon})\right)
\nonumber\\
&\qquad-
J_{\mathcal{D}_t}(\bx-\delta\boldsymbol{\epsilon})^{\top}
\nabla f\left(\mathcal{D}_t(\bx-\delta\boldsymbol{\epsilon})\right)
\bigg],
\label{eq:app_target_fd_hvp}
\end{align}
where $\delta>0$ is the finite-difference step size. By applying the
finite difference to the complete composite gradient, this
approximation captures the second-order contributions from both $f$
and $\mathcal{D}_t$, without explicitly forming $\nabla^2 f$ or the
componentwise Hessians of $\mathcal{D}_t$. In particular, we use this
approximation with $f=E^*$ because OpenMM provides only first-order
derivatives of the target energy.

In implementation, we evaluate the composite gradient without
materializing either $J_{\mathcal{D}_t}$ or $J_{D_t}$. For an
arbitrary evaluation point $\bz$, let
\begin{align*}
\left\{
\begin{aligned}
\bx_t^{\mathrm{LA}}(\bz) &= \mathcal{D}_t(\bz), \\
\bg_{\bz} &= \nabla f\!\left(\bx_t^{\mathrm{LA}}(\bz)\right).
\end{aligned}
\right.
\end{align*}
We obtain $\bg_{\bz}$ by automatic differentiation for differentiable
neural potentials. For $f=E^*$, we instead obtain this energy gradient
from the force returned by OpenMM, with the sign reversed. Using
$\bg_{\bz}$ as the upstream gradient at the output of $\mathcal{D}_t$,
backpropagation computes the vector-Jacobian product
\begin{align*}
\mathbf{G}_t(\bz)
&=
J_{\mathcal{D}_t}(\bz)^\top \bg_{\bz}
=
\nabla_{\bz}F_t(\bz).
\end{align*}
Because $D_t$ is evaluated as part of $\mathcal{D}_t$, backpropagation
also applies $J_{D_t}^\top$ internally without explicitly constructing
either Jacobian. Finally, setting
$\bz_{\pm}=\bx\pm\delta\boldsymbol{\epsilon}$, we approximate the
composite HVP by
\begin{align*}
\nabla_{\bx}^2F_t(\bx)\boldsymbol{\epsilon}
\approx
\frac{\mathbf{G}_t(\bz_+)-\mathbf{G}_t(\bz_-)}{2\delta},
\end{align*}
which is Eq.~\eqref{eq:app_target_fd_hvp}. Thus, the desired
second-order derivative can be computed by HVP,
whereas each composite gradient in
its finite-difference approximation is evaluated by a first-order VJP.
The same calculation applies to other look-ahead-composed terms for
which only first-order derivatives are available. However, the OpenMM
finite-difference approximation is time-consuming and cannot be
parallelized across particles. Using Machine-learned interatomic
potential (MLIP)
such as DPA4~\citep{li2026dpa4} to evaluate target energy
can help with this problem, and we leave
this for future work.

\paragraph{Variance correction for stochastic log-weights.}
An unbiased stochastic Laplacian generally does not yield an
unbiased incremental importance weight, because SMC exponentiates the
estimated log-weight increment. Consider the EDM forward process
\citep{karras2022elucidating} with $u_s=0$ and $\beta_s^2=2s$, so that
$\widetilde{u}_t=0$ and $\sigma_t^2=\beta_{T-t}^2=2(T-t)$; the potential
in Eq.~\eqref{eq:general_path_fk_potential} then reduces to
\begin{align*}
G_t(\bx)
=
-\partial_t\Phi_t(\bx)
-(T-t)\Delta\Phi_t(\bx)
+(T-t)\left\|\nabla\Phi_t(\bx)\right\|^2.
\end{align*}
For a fixed particle $\bx$ at time $t_k$, let $h_k=t_{k+1}-t_k$ and
define
\begin{align*}
D_k
&=
-\partial_t\Phi_{t_k}(\bx)
+(T-t_k)\left\|\nabla\Phi_{t_k}(\bx)\right\|^2,
\\
L_k
&=
\Delta\Phi_{t_k}(\bx),
\qquad
c_k=h_k(T-t_k).
\end{align*}
Up to a particle-independent centering term, which cancels once the
weights are normalized, the corresponding Euler log-weight increment is
\begin{align*}
\Delta\log w_k
=
h_kG_{t_k}(\bx)
=
h_kD_k-c_kL_k.
\end{align*}

Replacing $L_k$ by its stochastic $R$-probe estimate
$\widehat L_k=\Lhat_{t_k}(\bx)$ gives
\begin{align*}
\widehat{\Delta\log w}_k
=
\Delta\log w_k+\eta_k,
\qquad
\eta_k=-c_k(\widehat L_k-L_k).
\end{align*}
By the Hutchinson identity, $\mathbb{E}_{\boldsymbol{\epsilon}}[\eta_k\mid\bx]=0$, so
the estimated log-weight increment is unbiased conditional on the
particle, but the incremental weight itself is not: exponentiating gives
\begin{align*}
\mathbb{E}_{\boldsymbol{\epsilon}}
\left[
\exp\left(\widehat{\Delta\log w}_k\right)
\mid\bx
\right]
=
\exp(\Delta\log w_k)
\mathbb{E}_{\boldsymbol{\epsilon}}[\exp(\eta_k)\mid\bx]
\geq
\exp(\Delta\log w_k),
\end{align*}
where the inequality follows from Jensen's inequality and is strict
whenever the trace noise is non-degenerate. Since the trace-estimation
variance generally depends on $\bx$, this upward bias varies across
particles and thereby distorts their normalized relative weights.
Under the Gaussian approximation $\eta_k\mid\bx\sim\mathcal{N}(0,v_k)$,
the moment-generating function gives
\begin{align*}
\mathbb{E}_{\boldsymbol{\epsilon}}
\left[
\exp\left(\widehat{\Delta\log w}_k\right)
\mid\bx
\right]
=
\exp\left(\Delta\log w_k+\frac{v_k}{2}\right).
\end{align*}
Subtracting $v_k/2$ from the noisy log-weight increment therefore
cancels the multiplicative factor introduced by the additional trace
noise, following the classical cumulant-based bias correction of
\citet{zwanzig1954high} for exponentiated stochastic quantities. For
$R\geq2$, we estimate the conditional variance of the mean trace
estimator by
\begin{align*}
\widehat{\sigma}_{L,k}^2
=
\frac{1}{R(R-1)}
\sum_{r=1}^{R}
\left(
\ell_{t_k}^{(r)}(\bx)-\widehat L_k
\right)^2.
\end{align*}
Since the Laplacian estimate enters the log-weight increment with
coefficient $-c_k$, its additional variance at step $k$ is estimated by
\begin{align*}
\widehat v_k
=c_k^2\widehat{\sigma}_{L,k}^2.
\end{align*}
At every SMC step, we subtract half of this estimated variance from
the noisy log-weight increment:
\begin{align}
\widehat{\Delta\log w}_k^{\mathrm{corr}}
&=
h_kD_k
-c_k\widehat L_k
-\frac{1}{2}\widehat v_k
\nonumber\\
&=
h_kD_k
-c_k\widehat L_k
-\frac{1}{2}c_k^2\widehat{\sigma}_{L,k}^2.
\label{eq:app_corrected_log_weight}
\end{align}

In our implementation, we apply the variance correction in Eq.~\eqref{eq:app_corrected_log_weight} only to the Ala-4 free-energy estimation experiment.
In the other experiments, we did not observe a noticeable effect of this correction on sampling performance.

\paragraph{Hutch++ estimator.}
We can also estimate $\Delta\Phi_t(\bx)$ using advanced Hutch++
\citep{meyer2021hutch++}. Its key idea is to separate the Hessian into
a data-dependent low-rank component and an orthogonal residual. The
trace on the sketched subspace is evaluated directly, while stochastic
probes are applied only to the residual. Thus, when the sketch captures
the dominant spectral components of the Hessian, the randomly estimated
residual has lower variance than a Hutchinson estimate applied to the
full Hessian. Concretely, given an $m$-column Gaussian sketch
$S\in\R^{d\times m}$, let
$Q=\operatorname{orth}(H_t(\bx)S)$ and
$P_Q=\mathbf{I}_d-QQ^{\top}$. We then draw independent residual probes
$\bg^{(1)},\ldots,\bg^{(R)}\sim\mathcal{N}(0,\mathbf{I}_d)$.
For each probe, define the single-probe Hutch++ estimate
\begin{align*}
\ell_t^{++,(r)}(\bx)
=
\operatorname{tr}
\left(Q^{\top}H_t(\bx)Q\right)
+
\left(\bg^{(r)}\right)^{\top}
P_Q
H_t(\bx)
P_Q
\bg^{(r)}.
\end{align*}
Their average is $\Lhat_t^{++}=R^{-1}\sum_{r=1}^{R}
\ell_t^{++,(r)}$. The same low-rank term appears in every single-probe
estimate and therefore cancels when these estimates are centered.
For Hutch++, we can also adopt the variance correction method
and apply Eq.~\eqref{eq:app_corrected_log_weight}
to reduce the corresponding additional variance. Hutch++ is not used in our implementation and we leave it for future work.

\subsection{FKC with a time-dependent reward}
\label{app:fkc_time_dependent}

\citet[Proposition~3.4]{skreta2025feynman} considers tilts of the form $\beta_t r$. We extend it to an arbitrary deterministic, time-dependent reward $R_t$, which covers both the MIT and EAT paths.

\paragraph{Time-dependent reward tilting.}
Let $p_t>0$ satisfy the base diffusion equation
\begin{align*}
    \partial_t p_t
    =-\nabla\cdot\!\left[p_t\left(-\widetilde u_t+\sigma_t^2\nabla\log p_t\right)\right]
    +\frac{\sigma_t^2}{2}\Delta p_t,
\end{align*}
and let $R_t$ be $C^1$ in time and $C^2$ in space, with $Z_t=\int p_t(x)\,e^{R_t(x)}\,dx\in(0,\infty)$. We claim that, under the usual integrability and vanishing-boundary-flux conditions, the tilted path $q_t=Z_t^{-1}p_t\,e^{R_t}$ is represented by the weighted process
\begin{align}
    dX_t
    &=\left[-\widetilde u_t(X_t)+\sigma_t^2\nabla\log p_t(X_t)+\frac{\sigma_t^2}{2}\nabla R_t(X_t)\right]dt+\sigma_t\,dW_t,
    \label{eq:fkc_general_reward_drift}\\
    d\ell_t
    &=G_t^R(X_t)\,dt,
    \qquad
    G_t^R=\partial_t R_t+\left\langle\nabla R_t,\,\frac{\sigma_t^2}{2}\nabla\log p_t-\widetilde u_t\right\rangle,
    \label{eq:fkc_general_reward_weight}
\end{align}
initialized with $X_0\sim q_0$ and $\ell_0=0$, with weights proportional to $e^{\ell_t}$. The representation is exact in the sense that the normalized weighted marginals equal $q_t$.

To verify this claim, define $a_t=-\widetilde u_t+\frac{\sigma_t^2}{2}\nabla\log p_t$. Since $\nabla p_t=p_t\nabla\log p_t$, the diffusion term in the base equation partially cancels the score term in the drift, leaving the continuity equation $\partial_t p_t=-\nabla\cdot(p_ta_t)$. Differentiating $q_t=Z_t^{-1}p_te^{R_t}$ and using $\partial_t\log Z_t=\mathbb{E}_{q_t}[G_t^R]$, which follows by integrating over $x$ under the boundary conditions, gives
\begin{align*}
    \partial_t q_t
    =-\nabla\cdot(q_ta_t)+q_t\left(G_t^R-\mathbb{E}_{q_t}[G_t^R]\right).
\end{align*}
Since $\nabla q_t=q_t\nabla\log q_t$, the terms $-\nabla\cdot\bigl(q_t\,\sigma_t^2\nabla\log q_t /2\bigr)$ and $\sigma_t^2\Delta q_t / 2$ cancel. Adding the former to the drift and the latter as a diffusion term therefore leaves the equation unchanged. With $\nabla\log q_t=\nabla\log p_t+\nabla R_t$, the resulting drift is exactly~\eqref{eq:fkc_general_reward_drift}, and the reaction term is realized by the weights~\eqref{eq:fkc_general_reward_weight}. Taking $R_t=\beta_t r$ recovers Proposition~3.4 of \citet{skreta2025feynman}.

\paragraph{Instantiation on MIT and EAT.}
Both paths follow directly from \eqref{eq:fkc_general_reward_drift}--\eqref{eq:fkc_general_reward_weight} with a suitable choice of reward. For the MIT path, we simply take $R_t=r_t$. For the EAT path, we take
\begin{align*}
    R_t(\bx)=r_t(\bx)-\gamma\kappa_t\,\Delta E^{\mathrm{ref}}(\bx).
\end{align*}

Since $r_T=r$ and $\kappa_T=1$, the terminal reward is $R_T=r-\gamma\Delta E^{\mathrm{ref}}$, and the endpoint of the tilted path is
\begin{align*}
    q_T(\bx) &\propto\left(p_T^\theta(\bx)\right)^\gamma \exp\left({R_T(\bx)}\right) \\
    &\propto \exp\left({-\gamma E_\theta(\bx,T)}\right)\,\exp\left({r(\bx)-\gamma E^{\mathrm{ref}}(\bx)+\gamma E_\theta(\bx,T)}\right) \\
    & = \exp\left({-\gamma E^{\mathrm{ref}}(\bx)+r(\bx)}\right) \\
    & \propto q_T^{\mathrm{ref}}(\bx).
\end{align*}
FKC on EAT is therefore FKC on the model-induced construction with the redefined reward $\hat r=R_T$, under which the MIT coincides with $q_T^{\mathrm{ref}}$ and the endpoint mismatch vanishes by construction.
Any error of FKC on EAT at the population level therefore stems from path tracking.

\paragraph{Scope of the extension.}
These constructions extend FKC to both paths, but they still require $p_t^\theta$ to satisfy the base diffusion equation. For an imperfect model this condition generally fails, so the extended FKC formulas need not track the prescribed path, even at the population level. EBFKC avoids this limitation: it evaluates the full prescribed-path potential of Thm.~\ref{thm:general_path_fk} without imposing diffusion consistency.

\subsection{Tempering model-free SMC}
\label{app:tempering_smc}

\paragraph{Tempering path.}
As a model-free baseline, we use traditional SMC along a geometric
tempering path with a scheduled reward.
Let $\rho=\mathcal{N}(0,\Sigma_0)$ with $\Sigma_0\succ0$ and $\Phi_G(\bx)=\tfrac12\bx^\top\Sigma_0^{-1}\bx$.
For a smooth nondecreasing schedule $\lambda(t)$ satisfying
$\lambda(0)=0$ and $\lambda(T)=1$, we define the tempering path as
\begin{align}
\Phi_t^{\mathrm{temp}}(\bx)
&=(1-\lambda(t))\Phi_G(\bx)
+\lambda(t)\gamma E^*(\bx)-\alpha_tr(\bx),
\label{eq:tempering_path}\\
q_t^{\mathrm{temp}}(\bx)
&\propto\rho(\bx)^{1-\lambda(t)}
\bigl(p^\star(\bx)\bigr)^{\gamma\lambda(t)}
\exp\bigl(\alpha_tr(\bx)\bigr).\nonumber
\end{align}
With $\alpha_0=0$ and $\alpha_T=1$, the path starts at
$q_0^{\mathrm{temp}}=\rho$ and terminates at the desired target
$q_T^{\mathrm{temp}}\propto\exp(-\gamma E^*+r)$.

\paragraph{Prescribed-path Langevin Feynman--Kac equation.}
For a prescribed path $q_t=Z_t^{-1}\exp(-\Phi_t)$, this baseline uses
an annealed Langevin proposal. We choose a space-independent diffusivity
$\varepsilon_t\geq0$, with drift $v_t=-\varepsilon_t\nabla\Phi_t$
and diffusion coefficient $\sigma_t=\sqrt{2\varepsilon_t}$.
Assume $\Phi_t$ is $C^1$ in time
and $C^2$ in space, $0<Z_t<\infty$, differentiation under the
integral is valid, and boundary fluxes vanish. Then
\begin{align*}
\partial_t q_t
&=-\nabla\!\cdot(q_tv_t)+\varepsilon_t\Delta q_t+q_tg_t,
\\
G_t(\bx)&=-\partial_t\Phi_t(\bx),
\qquad g_t(\bx)=G_t(\bx)-\mathbb{E}_{q_t}[G_t].\nonumber
\end{align*}
Indeed, $\nabla q_t=-q_t\nabla\Phi_t$ makes the drift and diffusion
terms cancel, while differentiating $q_t$ gives $\partial_tq_t=q_tg_t$.
For the tempering path in Eq.~\eqref{eq:tempering_path}, the uncentered
reweighting potential is
$G_t(\bx)=\dot\lambda(t)[\Phi_G(\bx)-\gamma E^*(\bx)]
+\dot{\alpha}_tr(\bx)$.

\paragraph{Annealed Langevin dynamics with reweighting.}
With $\Phi_t=\Phi_t^{\mathrm{temp}}$, the corresponding weighted
particle dynamics between resampling events are
\begin{align*}
\rmd\bx_t^{(i)}
&=-\varepsilon_t\nabla\Phi_t(\bx_t^{(i)})\rmd t
+\sqrt{2\varepsilon_t}\,\rmd\bw_t^{(i)},
\\
\rmd\log w_t^{(i)}
&=\left[G_t(\bx_t^{(i)})
-\sum_{j=1}^Nw_t^{(j)}G_t(\bx_t^{(j)})\right]\rmd t.\nonumber
\end{align*}
Starting from $\bx_0^{(i)}\overset{\mathrm{iid}}{\sim}q_0^{\mathrm{temp}}$
with $w_0^{(i)}=1/N$, we discretize the particle SDE using
Euler--Maruyama and apply left-point Euler updates to the log-weights.
We then normalize the weights and resample when the ESS falls below
the prescribed threshold, resetting the weights to $1/N$.

\subsection{Alive Filter for numerical stability}
\label{app:alive_filter}

During the look-ahead step of SMC inference, a small number of noisy particles can be mapped to severely non-physical configurations, producing non-finite or extremely large target energies and gradients, which will destroy the following VCG and SMC reweighting.
Following the numerical stabilization strategy used in~\citep{tan2025scalable}, we apply an energy-based \emph{alive filter} to prevent numerically unstable, extremely high-energy configurations from propagating through SMC dynamics.

Specifically, at time $t_k$, we first use look-ahead map
to predict each particle's terminal position
$
    \bx_{t_k}^{\mathrm{LA},(i)}=\mathcal{D}_{t_k}\!\left(\bx_{t_k}^{(i)}\right).
$
In molecular experiments, we use the dimensionless reduced potential energy $u(\bx)=U_{\mathrm{OpenMM}}(\bx)/(R T_{\mathrm{target}})$ at the target temperature $T_{\mathrm{target}}$, where $U_{\mathrm{OpenMM}}$ is the molar potential energy and $R$ is the molar gas constant.
Given an energy threshold $c_{\mathrm{alive}}(t_k)$, particle $i$ is declared alive if both $u\!\left(\bx_{t_k}^{\mathrm{LA},(i)}\right)$ and its gradient are finite, and $u\!\left(\bx_{t_k}^{\mathrm{LA},(i)}\right)<c_{\mathrm{alive}}(t_k)$.

Let $\mathcal{S}$ be the surviving set, we renormalize the current SMC
weights over the surviving particles,
\begin{align*}
    \bar{w}_i
    =
    \frac{w_{t_k}^{(i)}}{
    \sum_{j\in\mathcal{S}}w_{t_k}^{(j)}},
    \qquad i\in\mathcal{S},
\end{align*}
and use systematic resampling from $\{\bar{w}_i\}_{i\in\mathcal{S}}$
to draw $N-|\mathcal{S}|$ parent particles whose copies replace the filtered
particles. The post-filter weights are
\begin{align*}
    \left\{
    \begin{aligned}
        w_{t_k}^{\prime(i)}
        &=
        \frac{|\mathcal{S}|}{N}\bar{w}_i,
        && i\in\mathcal{S},\\
        w_{t_k}^{\prime(j)}
        &=
        \frac{1}{N},
        && j\notin\mathcal{S}.
    \end{aligned}
    \right.
\end{align*}
Thus, the alive filter preserves both the particle count and the total
weight since $\sum_i w_{t_k}^{\prime(i)}=1$. The replacement particles
are sampled from the surviving set in proportion to the renormalized SMC
weights $\{\bar{w}_i\}_{i\in\mathcal{S}}$.
The alive filter samples from a truncated distribution by removing the extreme high-energy tail. In practice, this truncation removes only a few particles, and we observe no degradation in algorithm performance.

\begin{algorithm}[t]
\small
\DontPrintSemicolon
\caption{Energy-based Feynman--Kac Corrector on Energy-Aligned Target (EBFKC on EAT)}
\label{alg:energy-augmented-vcg-smc}
\KwIn{
Learned energy $E_\theta$, target energy $E^*$, path schedule $(\alpha,\kappa)$;
time grid $0 = t_0 < \cdots < t_K = T$; particle count $N$; ESS threshold $\tau$.
}
Construct the EAT path $q_t\propto\exp(-\Phi_t)$;
Initialize $\bx_0^{(i)}\sim q_0$ and $w_0^{(i)}\gets1/N$ for $i\in[N]$\;
\For{$k\gets0$ \KwTo $K-1$}{
Select the active EAT basis and solve the weighted VCG
system in Eq.~\eqref{eq:vcg-linear-system} for $\boldsymbol\lambda_{t_k}$\;
Form the control drift
$\bb_{t_k}=\sum_{j\in\mathcal I(t_k)}\lambda_{t_k}^j\boldsymbol{\psi}_j$
and residual potential $\phi_{t_k}=\hat g_{t_k}+h_{t_k}(\cdot;\bb_{t_k})$\;
Compute $\delta\ell_k^{(i)}\gets\Delta t_k\,
\phi_{t_k}(\bx_{t_k}^{(i)})$ at the current particle positions\;
Propagate particles to $t_{k+1}$ by Euler--Maruyama with drift
$v_{t_k}+\bb_{t_k}$ and diffusion coefficient $\sigma_{t_k}$\;
Reweight $w_{t_{k+1}}^{(i)}\propto w_{t_k}^{(i)}\exp(\delta\ell_k^{(i)})$
and normalize\;
\If{$\operatorname{ESS}(w_{t_{k+1}}^{(i)})<\tau$}{
Resample particles using their weights and reset
$w_{t_{k+1}}^{(i)}\gets1/N$\;
}
}
\KwOut{Weighted particles $\{(\bx_T^{(i)},w_T^{(i)})\}_{i=1}^N$
approximating $q_T^\star\propto\exp(-\gamma E^*+r)$.}
\end{algorithm}

\FloatBarrier
\section{Experimental Details}

\subsection{GMM settings}
\label{app:gmm_problem_settings}

\paragraph{Target distribution.}
We consider an equally weighted Gaussian mixture model (GMM) with $K=40$ components in $d=30$ dimensions:
\begin{align*}
p^*(\bx)
=
\frac{1}{K}
\sum_{k=1}^{K}
\mathcal{N}
\left(
\bx;
\boldsymbol{\mu}_k,
50\mathbf{I}_d
\right),
\end{align*}
where the component means are sampled independently from
$\operatorname{Unif}([-40,40]^d)$. We apply the variance exploding (VE)
forward process, whose samples at noise level $s$ are given by
$\by_s=\by_0+s\boldsymbol{\epsilon}$ for
$s\in[0,T]$, with $\by_0\sim\pstar$ and $\boldsymbol{\epsilon}\sim
\mathcal{N}(0,\mathbf{I}_d)$ independently. The resulting forward marginals are
\begin{align*}
r_s^\star(\by)
=
\frac{1}{K}
\sum_{k=1}^{K}
\mathcal{N}
\left(
\by;
\boldsymbol{\mu}_k,
(50+s^2)\mathbf{I}_d
\right).
\end{align*}
Under the generative-time convention $t=T-s$, these marginals give $p_t^\star(\bx)=r_{T-t}^\star(\bx)$. 
Consequently, both the intermediate energy $E_t^\star(\bx)=-\log \pstar_t(\bx)$ and its score function $s_t^*(\bx)=-\nabla_\bx E_t^\star(\bx)$ are available in closed form. 
This allows us to emulate pretrained model error through explicit perturbations to the component means and variance, with direct control over the perturbation magnitudes.

\paragraph{Controlled pretrained model error.}
We emulate an imperfect pretrained model by independently perturbing the component means and variance. 
For each component, we add an independent Gaussian perturbation to its mean:
$$\widehat{\boldsymbol{\mu}}_k=
\boldsymbol{\mu}_k+\boldsymbol{\xi}_k,$$
where $\boldsymbol{\xi}_k\sim\mathcal{N}(0,\sigma_\mu^2\mathbf{I}_d)$. 
We also perturb the reference variance $\sigma_s^2=50+s^2$ by rescaling the noise contribution by a factor of $1+\rho$ to obtain the perturbed variance
$$\widehat{\sigma}_s^2=50+(1+\rho)s^2.$$
These perturbations define the marginals
\begin{align*}
r_s^\theta(\by)
=
\frac{1}{K}
\sum_{k=1}^{K}
\mathcal{N}
\left(
\by;
\widehat{\boldsymbol{\mu}}_k,
\widehat{\sigma}_s^2\mathbf{I}_d
\right),
\end{align*}
with corresponding generative-time marginals $p_t^\theta(\bx)=r_{T-t}^\theta(\bx)$ and intermediate energy $\widehat{E}(\bx,t)=-\log p_t^\theta(\bx)$.

For the comparisons in Table~\ref{tab:gmm_results}, we set $\sigma_\mu=10$ and $\rho=0.1$. 
To evaluate robustness across perturbation magnitudes, we sweep each perturbation parameter over five evenly spaced values: $\sigma_\mu \in \{0, 2.5, 5, 7.5, 10\}$ and $\rho \in \{0, 0.025, 0.05, 0.075, 0.1\}$.

\paragraph{Inference-time targets.}
Following Eq.~\eqref{eq:true_target}, both tasks use the unified terminal target
\begin{align*}
q_T^\star(\bx)
\propto
\left(p^*(\bx)\right)^\gamma
\exp\left(r(\bx)\right).
\end{align*}
For annealing, we set $r(\bx)=0$ and $\gamma=2$ to sharpen the reference distribution. 
For reward tilting, we set $\gamma=1$ and use a quadratic reward centered at $\boldsymbol{\mu}\sim\mathcal{N}(\mathbf{0},100\mathbf{I}_d)$, with $\boldsymbol{\Sigma}=\sigma\mathbf{I}_d$ and $\sigma=100$:
\begin{align*}
r(\bx)
=
-\frac{1}{2}
\left(\bx-\boldsymbol{\mu}\right)^\top
\boldsymbol{\Sigma}^{-1}
\left(\bx-\boldsymbol{\mu}\right).
\end{align*}
This reward favors samples near $\boldsymbol{\mu}$, allowing us to test whether inference-time sampling can redirect probability mass toward
this region despite pretrained model error.

\subsection{Particle system settings}
\label{app:particle_system_settings}

Following the setup of DriftLite~\citep{ren2025driftlite}, we consider two particle systems: Double-Well (DW-4) and Lennard-Jones (LJ-13).
For a configuration $\bx=(\br_1,\ldots,\br_n)$ with a system-specific interaction energy $H(\bx)$, the equilibrium distribution at temperature $T_{\mathrm{phys}}$ is
\begin{align}
p^\star(\bx)
\propto
\exp\left\{
-\frac{1}{T_{\mathrm{phys}}}
\left[
H(\bx)
+
\frac{\lambda}{2}
\sum_{i=1}^{n}
\left\|
\br_i-\overline{\br}
\right\|^2
\right]
\right\},
\label{eq:particle_boltzmann_target}
\end{align}
where $\lambda$ controls harmonic confinement around configuration center $\overline{\br}$. 
For both particle systems, we obtain
ground truth reference samples from underdamped Langevin dynamics
simulations using the BAOAB splitting scheme.

For reward tilting, we use the following quadratic reward to favor
spatially compact configurations centered at $\overline{\br}$:
\begin{align}
r(\bx)
=
-\frac{\lambda'}{2}
\sum_{i=1}^{n}
\left\|
\br_i-\overline{\br}
\right\|^2.
\label{eq:particle_harmonic_reward}
\end{align}
Combining Eqs.~\eqref{eq:particle_boltzmann_target} and
\eqref{eq:particle_harmonic_reward} gives
\begin{align*}
p^\star(\bx)\exp\left(r(\bx)\right)
\propto
\exp\left\{
-\frac{1}{T_{\mathrm{phys}}}
\left[
H(\bx)
+
\frac{\lambda+T_{\mathrm{phys}}\lambda'}{2}
\sum_{i=1}^{n}
\left\|
\br_i-\overline{\br}
\right\|^2
\right]
\right\}.
\end{align*}
Thus, the reward preserves the interaction potential and temperature
while increasing the effective confinement strength from $\lambda$ to $\lambda+T_{\mathrm{phys}}\lambda'$. The default reward strength is $\lambda'=0.5$ for both particle systems.

\paragraph{Double-Well system.}
The DW-4 benchmark consists of four particles in two dimensions, so
$\bx\in\R^{4\times 2}$. Let
$d_{ij}=\|\br_i-\br_j\|$ denote the distance between a
pair of particles. The interaction energy is
\begin{align*}
H_{\mathrm{DW}}(\bx)
=
\sum_{i<j}
\left[
a(d_{ij}-d_0)
+b(d_{ij}-d_0)^2
+c(d_{ij}-d_0)^4
\right].
\end{align*}
We set $a=0$, $b=-4$, $c=0.9$, and $d_0=4$. The resulting pair
potential has two preferred separations,
$d_{ij}=4\pm\frac{2}{3}\sqrt{5}$, which induce the characteristic
double-well structure. The harmonic-confinement strength in
Eq.~\eqref{eq:particle_boltzmann_target} is $\lambda=0.05$.

\paragraph{Lennard-Jones system.}
The LJ-13 benchmark contains thirteen particles in three dimensions,
with $\bx\in\R^{13\times 3}$. Its interaction energy is
\begin{align*}
H_{\mathrm{LJ}}(\bx)
=
\frac{\epsilon}{2}
\sum_{i<j}
\left[
\left(\frac{\sigma}{d_{ij}}\right)^{12}
-2\left(\frac{\sigma}{d_{ij}}\right)^6
\right].
\end{align*}
The inverse-power terms produce strong short-range repulsion and
longer-range attraction, yielding a rugged energy landscape for the
thirteen-particle cluster. We use $\sigma=1$, $\epsilon=2$, and harmonic
confinement strength $\lambda=1$.

\subsection{Alanine Dipeptide settings}
\label{app:alanine_dipeptide_settings}

\paragraph{Molecular system and reference data.}
Following the setup of RNE~\citep{he2025rne}, we consider alanine dipeptide (ALDP), a 22-atom molecule represented by $\bx\in\R^{22\times 3}$. 
Its potential energy is specified by the AMBER ff96 force field with a generalized Born implicit-solvent model.

We use the configurations provided by RNE: 210,384 at $300\,\mathrm{K}$ and 150,000 at $800\,\mathrm{K}$. 
The $300\,\mathrm{K}$ reference ensemble was generated by a $5\,\mu\mathrm{s}$ molecular dynamics simulation using a Langevin middle integrator with a friction coefficient of $1\,\mathrm{ps}^{-1}$ and a time step of $2\,\mathrm{fs}$.
We center each configuration before training.

\paragraph{Free-energy estimation.}
We estimate the one-dimensional free-energy profile at $300\,\mathrm{K}$ along the backbone dihedral angle $\phi$. The periodic reaction coordinate is $\xi_{\mathrm{ALDP}}(\bx)=\phi(\bx)$, computed from the atom quadruplet $(4,6,8,14)$. 
To obtain the reference profile, we perform umbrella sampling with 25 windows whose centers are uniformly spaced over the periodic domain:
\[
\phi_i=-\pi+\frac{2\pi i}{25},\qquad i=0,\ldots,24.
\]
In each window, we use a Langevin middle integrator with a friction coefficient of $1\,\mathrm{ps}^{-1}$ and a time step of $2\,\mathrm{fs}$.
Each trajectory consists of $50{,}000$ equilibration steps ($100\,\mathrm{ps}$) followed by $1{,}000{,}000$ production steps ($2\,\mathrm{ns}$). We save one configuration every $1{,}000$ production steps ($2\,\mathrm{ps}$), yielding $1{,}000$ configurations per window.

\subsection{Alanine Tetrapeptide settings}
\label{app:alanine_tetrapeptide_settings}

\paragraph{Molecular system and reference data.}
Following \citet{invernizzi2022skipping}, we consider a 42-atom alanine tetrapeptide (Ala-4) system. 
The original benchmark constrains covalent bonds involving hydrogen atoms and represents configurations using 98 internal coordinates. 
We instead impose no bond-length constraints (\texttt{constraints=None} in OpenMM) and use Cartesian coordinates $\bx\in\R^{42\times3}$, retaining the stretching degrees of freedom of bonds involving hydrogen atoms. 
The backbone has three pairs of Ramachandran angles, $(\phi_i,\psi_i)_{i=1}^{3}$. The three $\phi_i$ angles are slow variables, and their sign combinations define eight metastable basins.

To generate reference data for the implicit-solvent equilibrium ensemble at $300\,\mathrm{K}$, we run molecular dynamics with on-the-fly probability enhanced sampling (OPES)~\citep{invernizzi2020rethinking}, applying the bias along all three slow dihedrals $\phi_1$, $\phi_2$, and $\phi_3$.
We use a friction coefficient of $1\,\mathrm{ps}^{-1}$ and a time step of $1\,\mathrm{fs}$. 
The simulation consists of $500{,}000$ equilibration steps ($0.5\,\mathrm{ns}$) without OPES bias, followed by $100{,}000{,}000$ production steps ($100\,\mathrm{ns}$) with the bias updated every $1{,}000$ steps ($1\,\mathrm{ps}$). 
The resulting dataset contains $100{,}000$ configurations with associated reweighting weights. We use these weights to estimate the equilibrium marginal of $\phi_3$ and obtain the reference free-energy profile.

\paragraph{Free-energy estimation.}
We estimate the one-dimensional free-energy profile at $300\,\mathrm{K}$ along $\xi_{\mathrm{Ala4}}(\bx)=\phi_3(\bx)\in[-\pi,\pi)$, whose range is covered by overlapping umbrella windows. 
For the traditional umbrella sampling baseline, we use the ALDP settings in App.~\ref{app:alanine_dipeptide_settings}, with the bias applied to $\phi_3$.
For umbrella diffusion sampling, we combine the terminal weighted particles from all windows using weighted MBAR, as described in App.~\ref{app:molecular_free_energy_settings}. 
All diffusion-based reward-tilting methods use the same window definitions, per-window sampling budgets, and angular bins. Each resulting free-energy profile is shifted so that its minimum is zero.

\subsection{Molecular free-energy estimation methods}
\label{app:molecular_free_energy_settings}
We estimate the potential of mean force (PMF) along a periodic reaction coordinate $\xi(\bx)\in[-\pi,\pi)$. 
At temperature $T_{\mathrm{phys}}$, the target equilibrium distribution is $p^*(\bx)\propto\exp[-\beta_{\mathrm{th}}U(\bx)]$, where $U$ is the molecular potential energy and $\beta_{\mathrm{th}}=1/(k_{\mathrm B}T_{\mathrm{phys}})$.

\paragraph{Traditional umbrella sampling.}
Traditional umbrella sampling covers the reaction-coordinate range with $K$ overlapping biased ensembles. 
Because a dihedral angle is periodic, we measure displacement from a window center $c$ by $d_{\mathbb T}(\zeta,c)=\operatorname{atan2}(\sin(\zeta-c), \cos(\zeta-c))$. 
Window $k$ uses the restraint potential $B_k(\bx)=\kappa_k^{\mathrm{umb}} d_{\mathbb T}(\xi(\bx),c_k)^2/2$ and its dimensionless counterpart $b_k(\bx)=\beta_{\mathrm{th}}B_k(\bx)$.
The desired biased ensemble in each window is therefore
\begin{align}
q_k^*(\bx)
=
\frac{1}{Z_k}
p^*(\bx)
\exp\left[-b_k(\bx)\right].
\label{eq:app_umbrella_target}
\end{align}
Here, $Z_k$ is the normalization constant of window $k$.
In traditional umbrella sampling, molecular dynamics under $U+B_k$ generates samples for each window. 
MBAR then combines the generally correlated samples across windows to account for the bias potentials and estimate equilibrium statistics under $p^*$.

\paragraph{Umbrella diffusion sampling.}
Following UmbrellaDiff~\citep{xie2026enhanced}, we formulate sampling within each umbrella window as a reward-tilting task with $\gamma=1$ and $r_k(\bx)=-b_k(\bx)$, so that the desired terminal distribution is $q_k^*$ in Eq.~\eqref{eq:app_umbrella_target}. 
For each window, we initialize particles independently from the Gaussian prior and run the SMC sampler based on the pretrained diffusion models, obtaining terminal weighted particles
$
\left\{
\left(\bx_{k,i},w_{k,i}\right)
\right\}_{i=1}^{N_k}.
$

\paragraph{Weighted MBAR.}
We combine weighted particles across umbrella windows using the weighted multistate Bennett acceptance ratio (weighted MBAR) estimator of \citet{xie2026enhanced}. 
For window $k$, we compute the effective sample size as $\operatorname{ESS}_k=(\sum_i w_{k,i})^2/ \sum_i w_{k,i}^2$.
We normalize the relative weights within each window as $\widetilde w_{k,i}=w_{k,i}/\sum_{j=1}^{N_k}w_{k,j}$ and assign each sample the effective mass $\alpha_{k,i}=\operatorname{ESS}_k\widetilde w_{k,i}$.
The total effective mass of window $k$ is then $M_k=\sum_i \alpha_{k,i}=\operatorname{ESS}_k$.
This construction preserves the relative particle weights within each window and sets its total effective mass to $\operatorname{ESS}_k$.

Pooling all windows gives $\{(\bx_n,\alpha_n)\}_{n=1}^N$, where $N=\sum_{k=1}^K N_k$. 
We estimate the reduced free energies $\{\fhat_k\}_{k=1}^K$ of the biased ensembles by solving the weighted MBAR self-consistency equations, for $k=1,\ldots,K$:
\begin{align}
\exp(-\fhat_k)
=
\sum_{n=1}^{N}
\frac{
\alpha_n\exp\left[-b_k(\bx_n)\right]
}{
\sum_{\ell=1}^{K}
M_\ell
\exp\left[\fhat_\ell-b_\ell(\bx_n)\right]
}.
\label{eq:app_weighted_mbar}
\end{align}
These equations determine the free energies up to a common additive
constant, which does not affect normalized observables.

Given these reduced free energies, we reweight the pooled particles to estimate observables in a target state $a\in\{0,1,\ldots,K\}$. 
The unnormalized weight of sample $\bx_n$ in this state is
\begin{align*}
W_n^{(a)}
 =
\frac{
\alpha_n\exp\left[-b_a(\bx_n)\right]
}{
\sum_{\ell=1}^{K}
M_\ell
\exp\left[\fhat_\ell-b_\ell(\bx_n)\right]
}.
\end{align*}
Here, $a=1,\ldots,K$ index the sampled biased states, whereas $a=0$ denotes the unbiased equilibrium state, for which $b_0\equiv0$. 
Because no samples are drawn directly from the unbiased state, it does not appear in the denominator of Eq.~\eqref{eq:app_weighted_mbar}. 
We denote its unnormalized weights by $W_n\equiv W_n^{(0)}$ and normalize them as $\overline W_n=W_n/\sum_m W_m$. 
When the window particles consistently approximate $q_k^*$, the expectation of an observable $o(\bx)$ under the unbiased ensemble $p^*$ is estimated using these normalized equilibrium weights:
\begin{align*}
\mathbb{E}_{p^*}[o(\bx)]
\approx
\sum_{n=1}^{N}\overline W_n o(\bx_n).
\end{align*}

\paragraph{Free-energy profile estimation.}
We use the normalized equilibrium weights $\overline W_n$ to estimate the marginal density along $\xi$ and construct the PMF for ALDP and Ala-4.
The equilibrium marginal density is $p_\xi(\zeta)=\int p^*(\bx) \delta(\xi(\bx)-\zeta)\,\mathrm d\bx$.
For a partition $\{I_j\}$ of $[-\pi,\pi)$, we estimate the probability mass in bin $j$ using the weighted samples:
\begin{align*}
\widehat m_j
 =
\sum_n
\overline W_n
\mathbf{1}\!\left\{\xi(\bx_n)\in I_j\right\}.
\end{align*}
The corresponding density estimate is $\widehat p_j=\widehat m_j/|I_j|$, which gives the discrete PMF
\begin{align*}
\Fhat_j
=
-k_{\mathrm B}T_{\mathrm{phys}}
\log \widehat p_j
+C,
\end{align*}
where $C$ is an arbitrary additive constant. We choose $C$ so that $\min_j\Fhat_j=0$ and plot each estimate at the corresponding bin center.

\subsection{Training and inference details}
\label{app:training_inference_details}

\paragraph{Training configuration.}
We train the energy-based diffusion models using the Adam optimizer
with a learning rate of $5\times10^{-4}$. The batch size is $1024$
for Stage~1 and $512$ for Stage~2 by default.
The loss weighting function is
\begin{equation*}
\lambda(\sigma)
=
\frac{\sigma^2+\sigma_{\mathrm{data}}^2}
{\sigma^2\sigma_{\mathrm{data}}^2},
\end{equation*}
where $\sigma$ is the noise level and $\sigma_{\mathrm{data}}^2$
is the empirical scalar variance of the training data. 
Table~\ref{tab:egnn_architectures} lists the architecture and number of training epochs for each system.

\begin{table}[t]
\centering
\caption{EGNN architectures and EBM training details.}
\label{tab:egnn_architectures}
\small
\setlength{\tabcolsep}{4pt}
\begin{tabular}{lccp{0.49\textwidth}}
\toprule
System & Layers & Hidden dim. & EBM training epochs \\
\midrule
DW-4 & 4 & 32 & Stage~1: 125 epochs \\
LJ-13 & 4 & 32 & Stage~1: 300 epochs \\
ALDP ($800\,\mathrm{K}$) & 4 & 32 & Stage~1: 200 epochs; Stage~2: 200 epochs \\
ALDP ($300\,\mathrm{K}$) & 4 & 32 & Stage~1: 200 epochs; Stage~2: 200 epochs \\
Ala-4 & 6 & 128 & Stage~1: 500 epochs \\
\bottomrule
\end{tabular}
\end{table}

\paragraph{Diffusion schedule.}
During inference, we use the following EDM-style noise schedule:
\begin{equation*}
\sigma_i
=
\left[
\sigma_{\max}^{1/7}
+\frac{i}{K}
\left(\sigma_{\min}^{1/7}-\sigma_{\max}^{1/7}\right)
\right]^7,
\qquad i=0,\ldots,K,
\end{equation*}
where $K$ denotes the number of inference steps and $\sigma_i$ is the noise level at step $i$.
For numerical stability, we truncate integration at $t_K=T-\sigma_{\min}$, corresponding to a small positive terminal noise level $\sigma_{\min}$.
For molecular systems, we use $(K,\sigma_{\min},\sigma_{\max})=(1000,0.002,80)$.
The corresponding settings are $(1000,0.005,50)$ for particle systems and
$(500,0.005,50)$ for GMM, with $T=\sigma_{\max}$ throughout.
For the pretrained reverse-diffusion evaluations, we use the second-order Heun sampler with 512 steps.
For annealing experiments, we initialize particles from the annealed Gaussian prior $q_0(\bx)\propto\rho(\bx)^\gamma$.
For $\rho=\mathcal{N}(0,\sigma_{\max}^2 \mathbf{I})$, this corresponds to sampling from $\mathcal{N}(0,\frac{\sigma_{\max}^2}{\gamma} \mathbf{I})$ with uniform initial weights.

\paragraph{Reward and correction schedules.}
We instantiate the intermediate reward as $r_t(\bx)=\alpha_t r(\bx)$.
The reward schedule $\alpha_t$ and correction schedule $\kappa_t$ satisfy
\begin{align*}
\alpha_0=\kappa_0=0,\qquad
\alpha_T=\kappa_T=1.
\end{align*}
We define the linear normalized progress as
\begin{equation*}
p(t)=\operatorname{clip}\!\left(
\frac{t}{T-\sigma_{\min}},0,1\right)\in[0,1].
\end{equation*}
At the inference grid points $t_i=T-\sigma_i$, this equals
$(\sigma_{\max}-\sigma_i)/(\sigma_{\max}-\sigma_{\min})$.
To control when each schedule activates and how it increases, we construct the delayed-power smootherstep non-linear progress based on the quintic polynomial $S(u)=6u^5-15u^4+10u^3$:
\begin{equation*}
A(p;s,q)
=S\!\left(
\left[\operatorname{clip}\!\left(\frac{p-s}{1-s},0,1\right)\right]^q
\right),
\qquad s\in[0,1),\quad q>0.
\end{equation*}
Then the reward and correction schedules can be configured independently as
\begin{align*}
\alpha_t&=A\!\left(p(t);s_\alpha,q_\alpha\right),\qquad
\kappa_t=A\!\left(p(t);s_\kappa,q_\kappa\right).
\end{align*}
Here, $s$ sets the activation point, while $q$ controls the growth shape. 
Separate parameter pairs specify the reward and correction schedules.
Table~\ref{tab:schedule_parameters} lists the settings for each experiment.

For Ala-4 umbrella diffusion sampling, a single pair of schedules does not serve all windows well, so we choose the activation order per window; since both orders share the same terminal values $\alpha_T=\kappa_T=1$, this does not change the window target.
For windows centered at $\phi_3<0$, where the pretrained model already provides adequate particle support, we use a \emph{reward-first} path with $s_\alpha<s_\kappa$, so that the reward is activated before the target correction.
For windows centered at $\phi_3>0$, a low-probability region that the pretrained model covers poorly, we use a \emph{target-first} path with $s_\alpha>s_\kappa$, which first corrects the pretrained model error to improve particle coverage and then concentrates the corrected particles within the window.

\begin{table}[H]
\centering
\caption{Reward and target-correction schedule parameters.}
\label{tab:schedule_parameters}
\small
\setlength{\tabcolsep}{5pt}
\begin{tabular}{@{}p{0.39\textwidth}cp{0.33\textwidth}@{}}
\toprule
Experiment & $(s_\kappa,q_\kappa)$ & $(s_\alpha,q_\alpha)$ \\
\midrule
DW-4 & $(0.999,3)$ & (0.999,3) \\
LJ-13 & $(0.999,4)$ & $(0.999,2)$ \\
ALDP annealing & $(0.999,3)$ & - \\
ALDP umbrella diffusion sampling & $(0.999,3)$ & $(0.9988,1)$ \\
Ala-4 umbrella diffusion sampling & $(0.999,3)$ &
Reward-first: $(0.9988,1)$\newline
Target-first: $(0.9994,1)$ \\
\bottomrule
\end{tabular}
\end{table}

\paragraph{Look-ahead schedule.}
We parameterize the look-ahead strength by the noise level so that the look-ahead map transitions from the denoiser prediction to the identity as sampling approaches the endpoint. 
We define
\begin{equation*}
\omega_\sigma
=S\!\left(
\operatorname{clip}\!\left(
\frac{\sigma-\sigma_{\min}}
{\sigma_{\mathrm{LA}}-\sigma_{\min}},0,1
\right)
\right),
\end{equation*}
For $\sigma_{\mathrm{LA}}>\sigma_{\min}$, the schedule equals one for $\sigma\geq\sigma_{\mathrm{LA}}$ and decreases to zero at $\sigma_{\min}$.
We set $\sigma_{\mathrm{LA}}=4\sigma_{\min}$ by default.

\paragraph{Inference configuration.}
We use $N=8192$ particles for GMM, DW-4, and LJ-13, and $N=1000$ particles per batch for ALDP annealing. 
For free-energy estimation, we use $N=200$ particles per window for ALDP and $N=12$ for Ala-4.
For the Ramachandran plots, we pool terminal particles from 250 batches of $1{,}000$ particles each, yielding $250{,}000$ samples.

Across all SMC experiments, we use systematic resampling with an ESS threshold of $0.9N$. 
For methods using VCG, the regularization parameter is $\lambda_{\mathrm{VCG}}=10^{-6}$.
Unless otherwise stated, experiments use five random seeds.

For Hutchinson trace estimation, we use $R=2$ probes for DW-4, $R=4$ for LJ-13 and ALDP, and $R=8$ for Ala-4. 
When finite differences are required, we use a default step size of $\delta=10^{-4}$.

For the alive filter, we set the energy cutoff at time $t_k$ to $c_{\mathrm{alive}}(t_k)=\frac{10}{\kappa_{t_k}}$ for ALDP and $c_{\mathrm{alive}}(t_k)=\frac{10^{20}}{\kappa_{t_k}}$ for Ala-4.
For ALDP, the terminal cutoff of $10$ exceeds the maximum reduced potential energy of $-8.35$ observed among 125,000 reference configurations at $300\,\mathrm{K}$. 
Thus, none of these reference configurations exceeds the energy cutoff. The alive filter is used to exclude numerically problematic configurations encountered during SMC inference.

\paragraph{MALA rejuvenation.}
To improve particle mixing in molecular experiments,
we apply Metropolis-adjusted Langevin
algorithm (MALA) rejuvenation during SMC inference. For ALDP annealing,
we enable MALA rejuvenation when $\kappa_t>10^{-4}$ and perform 2 MALA steps
every 25 SMC steps. ALDP umbrella diffusion sampling uses the same
rule, with 5 additional MALA steps at the endpoint.
For Ala-4 umbrella diffusion sampling, we instead enable MALA from the 500th SMC step and perform 2 MALA steps every 10 SMC steps, followed by 5 final MALA steps. 
In all molecular experiments, we additionally perform four MALA steps immediately after each resampling event.

\paragraph{Tempering SMC configuration.}
For the tempering SMC baseline, we use $N=8192$ for LJ-13 and a
uniform time grid with 1000 steps,
\begin{equation*}
t_k=\frac{k}{1000},\quad k=0,\ldots,1000,
\qquad \Delta t=0.001.
\end{equation*}
We use the DABG~\citep{grenioux2026diffusion} exponential tempering schedule
\begin{equation*}
\lambda(t)=\frac{\delta_{\mathrm{temp}}^{\,1-t}-\delta_{\mathrm{temp}}}
{1-\delta_{\mathrm{temp}}},
\qquad \delta_{\mathrm{temp}}=10^{-5},
\end{equation*}
and choose the diffusivity schedule
\begin{equation*}
\varepsilon(t)=\varepsilon_{\mathrm{final}}
+\bigl(\varepsilon_{\mathrm{initial}}-\varepsilon_{\mathrm{final}}\bigr)(1-t)^2,
\qquad t\in[0,1],
\end{equation*}
where $\varepsilon_{\mathrm{initial}}=1500$ and
$\varepsilon_{\mathrm{final}}=1$.

\subsection{Evaluation metrics}
\label{app:evaluation_metrics}
We describe the metrics and evaluation protocols used in our experiments.
Throughout this section, $P$ and $Q$ denote the generated and reference empirical distributions with normalized sample weights.
$E(\bx)=-\gamma\log p^\star(\bx)-r(\bx)$ denotes the target energy for the inference-time scaling task.

\paragraph{GMM evaluation.}
Following DriftLite~\citep{ren2025driftlite}, we evaluate GMM results using the following metrics.
Let the generated samples be weighted samples $\{(w_i,\bx_i)\}_{i=1}^N$ and the reference samples be uniform-weight samples $\{(1/M,\by_j)\}_{j=1}^M$.
\begin{itemize}[leftmargin=*]
    \item \textit{Negative log-likelihood (NLL) difference:}
    $|\Delta\mathrm{NLL}|=|\mathbb{E}_{P}[E]-\mathbb{E}_{Q}[E]| $
    measures the log-likelihood difference between the generated and reference distributions.

    \item \textit{Maximum mean discrepancy (MMD):}
    $\mathrm{MMD}=\|\mathbb{E}_{P}[k(\bx,\cdot)]-
    \mathbb{E}_{Q}[k(\bx,\cdot)]\|_{\mathcal H}$ measures the gap between kernel mean embeddings. 
    We use the RBF kernel $k(\bx,\by)=\exp(-\|\bx-\by\|^2/(2\sigma_k^2))$ with bandwidth $\sigma_k=20$ for GMM and $\sigma_k=5$ for the particle systems.
    In all experiments, we approximate the kernel using 2048 random Fourier features.
    \item \textit{Sliced-Wasserstein distance (SWD):}
    We report the computationally efficient sliced-Wasserstein distance
    $\mathrm{SWD}_2^2(P,Q)\approx L^{-1}\sum_{\ell=1}^{L}
    W_2^2(\pi_{\theta_\ell}P,\pi_{\theta_\ell}Q)$,
    where $\pi_\theta$ projects configurations onto direction $\theta$.
    In all experiments, we use $L=10$ projection directions sampled uniformly from the unit sphere.
    \item \textit{Mean $L_2$:}
    $\mathrm{Mean}\ L_2=\|\hat{\boldsymbol{\mu}}_x-\hat{\boldsymbol{\mu}}_y\|_2$ measures the Euclidean distance between the global sample means,
    where $\hat{\boldsymbol{\mu}}_x=\sum_{i=1}^N w_i\bx_i$ and $\hat{\boldsymbol{\mu}}_y=M^{-1}\sum_{j=1}^M\by_j$.
    \item \textit{Cov $F$:}
    $\mathrm{Cov}\ F=\|\hat{\boldsymbol{\Sigma}}_x-\hat{\boldsymbol{\Sigma}}_y\|_F$ measures the Frobenius distance between the global sample covariance matrices.
    Here, $\hat{\boldsymbol{\Sigma}}_x$ is the weighted sample covariance with degrees-of-freedom correction $1-\sum_{i=1}^N w_i^2$,
    and $\hat{\boldsymbol{\Sigma}}_y$ is the reference sample covariance with denominator $M-1$.
\end{itemize}

\paragraph{Particle system evaluation.}
Following DriftLite~\citep{ren2025driftlite}, we evaluate particle systems using the NLL difference, MMD and SWD defined above, together with two additional metrics for interparticle structure and the target-energy distribution.
\begin{itemize}[leftmargin=*]
    \item \textit{Radial Distribution Function Wasserstein-1 Distance:}
    $W_1^{\mathrm{RDF}}=W_1(g_P,g_Q)$ compares pooled pairwise-distance
    distributions. Here, the ensemble-averaged interatomic distance distribution
    $g$ describes the distribution of interparticle distances and
    characterizes the spatial arrangement of particles.
    \item \textit{Energy Wasserstein-1 distance:}
    $W_1^E=W_1(E_{\#}P,E_{\#}Q)$ compares the full target-energy
    distributions, where $E_{\#}P$ denotes the distribution of $E(\bx)$
    for $\bx\sim P$.
\end{itemize}

\paragraph{ALDP annealing evaluation.}
We evaluate ALDP annealing using four metrics. 
For each run reported in Table~\ref{tab:aldp_annealing}, we compare 25,000 generated samples with all RNE reference configurations at the corresponding temperature. 
Before evaluation, we center each conformation by subtracting its mean atomic position and apply chirality folding to match the reference enantiomer.
To assess pretrained model error at $800\,\mathrm{K}$, we evaluate reverse-diffusion samples generated using the second-order Heun sampler with 512 steps, as described in App.~\ref{app:training_inference_details}.

\begin{itemize}[label=\textbullet,leftmargin=*]
\item \textit{Ramachandran TVD:} We compare the joint distributions of the backbone dihedral angles $(\phi,\psi)$ using normalized $64\times64$ histograms over $[-\pi,\pi)^2$. 
The angles $\phi$ and $\psi$ are computed from the atom quadruplets $(4,6,8,14)$ and $(6,8,14,16)$, respectively.
Let $P_{ij}$ and $Q_{ij}$ denote the generated and reference probabilities
in bin $(i,j)$, respectively. 
The Ramachandran total variation distance
(TVD) is
\begin{align*}
D_{\mathrm{Ram}}
=
\frac{1}{2}
\sum_{i=1}^{64}\sum_{j=1}^{64}
\left|P_{ij}-Q_{ij}\right|.
\end{align*}

\item \textit{Basin-population error:} We compare the relative populations of four regions in the Ramachandran plane, defined by the signs of $\phi$ and $\psi$:
$\alpha_R$ for $\phi<0,\psi<0$; extended/$\beta$ for $\phi<0,\psi\geq0$; $\mathrm{C7}_{\mathrm{ax}}$ for $\phi\geq0,\psi<0$; and $\alpha_L$ for $\phi\geq0,\psi\geq0$. 
Let $p_b$ and $q_b$ denote the generated and reference probabilities in basin $b$, respectively. 
The basin-population error is
\begin{align*}
E_{\mathrm{basin}}
=
\frac{1}{2}
\sum_{b=1}^{4}\left|p_b-q_b\right|.
\end{align*}

\item \textit{Energy TVD:} We compare the reduced-energy distributions using 100 equal-width bins spanning the MD reference energy range. 
We evaluate the potential energy $U(\bx)$ of each conformation using the unconstrained OpenMM ALDP system and convert it to the reduced energy $u(\bx)=\beta U(\bx)$, where $\beta=1/(k_{\mathrm{B}}T)$.
Following the evaluation convention of FKC~\citep{skreta2025feynman} and RNE~\citep{he2025rne}, we exclude samples outside the reference energy range before normalizing the histograms. 
Let $P_i^{(E)}$ and $Q_i^{(E)}$ denote the normalized generated and reference bin probabilities, respectively. 
The energy TVD is
\begin{align*}
D_{\mathrm{energy}}
=
\frac{1}{2}
\sum_{i=1}^{100}
\left|P_i^{(E)}-Q_i^{(E)}\right|.
\end{align*}

\item \textit{Distance TVD:} We compare pooled distributions of all $\binom{22}{2}=231$ pairwise atomic distances using 200 equal-width bins spanning the MD reference distance range. 
We compute these distances for each conformation and pool them across conformations. 
Let $P_i^{(d)}$ and $Q_i^{(d)}$ denote the normalized generated and reference bin probabilities, respectively. 
The distance TVD is
\begin{align*}
D_{\mathrm{distance}}
=
\frac{1}{2}
\sum_{i=1}^{200}
\left|P_i^{(d)}-Q_i^{(d)}\right|.
\end{align*}
\end{itemize}

\section{Additional Experimental Results}

\makeatletter
\setlength{\@fptop}{0pt}
\setlength{\@fpsep}{12pt}
\setlength{\@fpbot}{0pt plus 1fil}
\makeatother

\renewcommand{\bottomfraction}{0.8}
\renewcommand{\floatpagefraction}{0.85}
\setcounter{topnumber}{4}
\setcounter{bottomnumber}{3}
\setcounter{totalnumber}{6}
\setlength{\textfloatsep}{10pt plus 2pt minus 2pt}
\setlength{\floatsep}{10pt plus 2pt minus 2pt}
\setlength{\intextsep}{10pt plus 2pt minus 2pt}

\subsection{GMM results}
\label{app:gmm_additional_results}

\begin{figure}[!htbp]
    \centering
    \begin{minipage}[t]{0.44\textwidth}
        \centering
        \includegraphics[width=\linewidth]{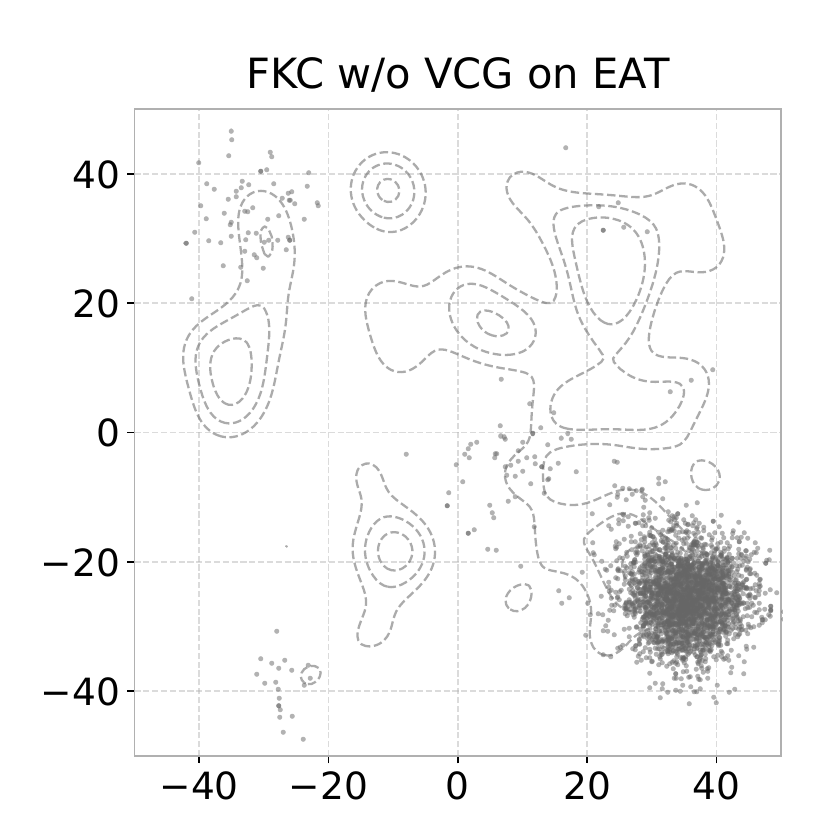}
        \par\small\textbf{(a) Annealing}
    \end{minipage}
    \hspace{0.03\textwidth}
    \begin{minipage}[t]{0.44\textwidth}
        \centering
        \includegraphics[width=\linewidth]{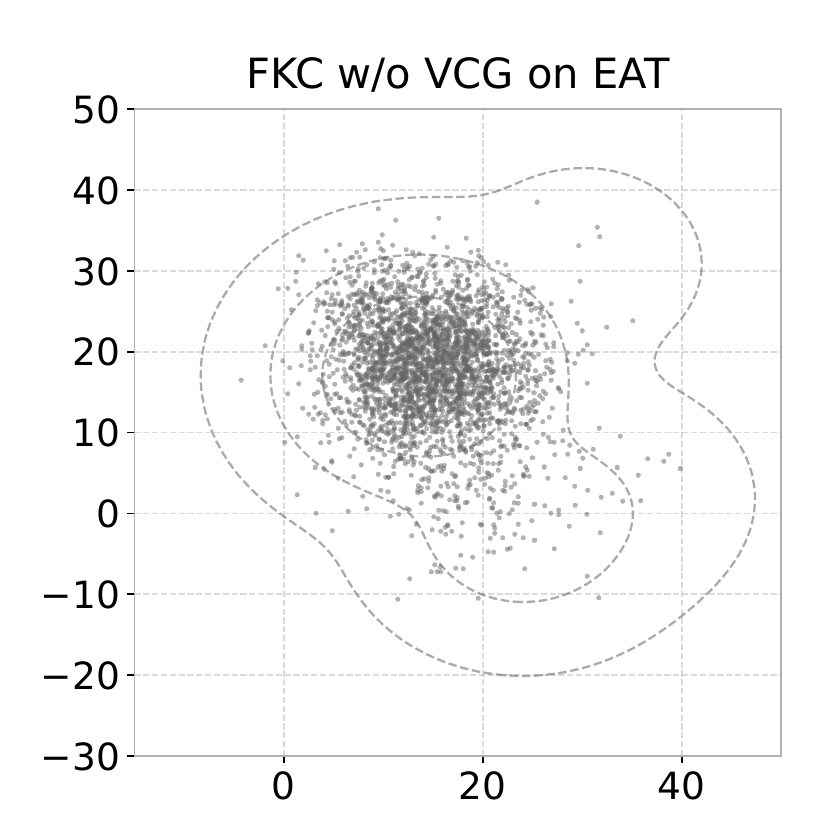}
        \par\small\textbf{(b) Reward tilting}
    \end{minipage}
    \caption{Sample distribution of FKC w/o VCG on EAT in the GMM task under (a) annealing and (b) reward tilting.}
    \label{fig:gmm_fkc_no_vcg_on_eat}
\end{figure}

Without VCG, FKC on EAT concentrates samples in a limited subset of the target modes under both annealing and reward tilting, leaving other modes poorly represented (Fig.~\ref{fig:gmm_fkc_no_vcg_on_eat}).

\begin{figure}[!htbp]
    \centering
    \begin{minipage}[t]{0.48\textwidth}
        \centering
        \includegraphics[width=\linewidth]{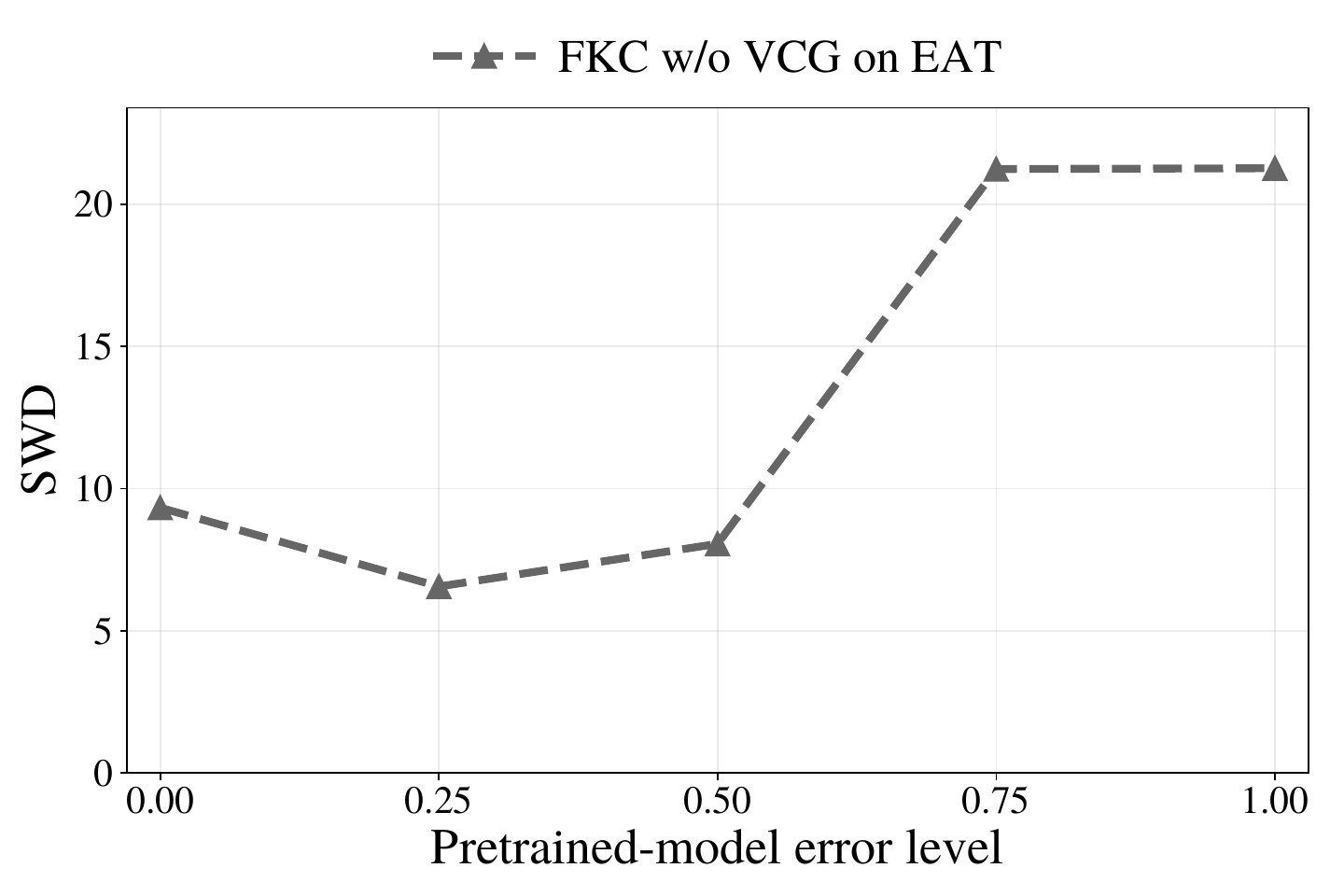}
        \par\small\textbf{(a) Annealing}
    \end{minipage}
    \hfill
    \begin{minipage}[t]{0.48\textwidth}
        \centering
        \includegraphics[width=\linewidth]{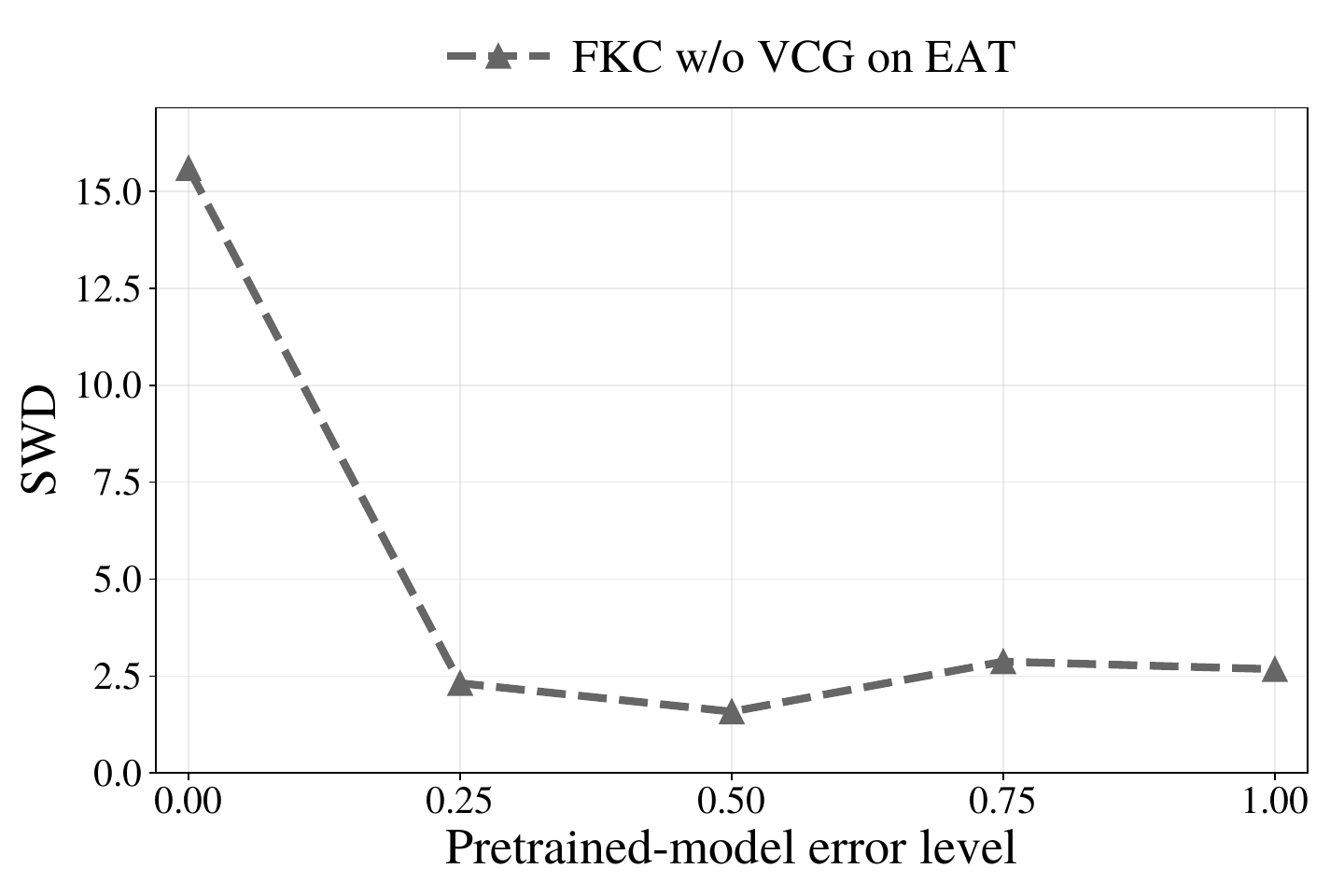}
        \par\small\textbf{(b) Reward tilting}
    \end{minipage}
    \caption{SWD of FKC w/o VCG on EAT across pretrained-model error levels on the GMM under (a) annealing and (b) reward tilting.}
    \label{fig:gmm_no_vcg_swd_sweep}
\end{figure}

The SWD sweep in Fig.~\ref{fig:gmm_no_vcg_swd_sweep} further shows substantial sampling error even at zero pretrained-model error, with no monotonic dependence on the perturbation level in either task.
Together with the FKC on EAT results using VCG in Fig.~\ref{fig:gmm_results}, these observations support the essential role of VCG in stablizing Feynman-Kac dynamics.

Table~\ref{tab:gmm_runtime} reports the runtime of each method.
FKC on EAT and EBFKC on EAT take longer than either MIT variant.
Both EAT methods require additional evaluations for the target-energy
correction, which add to the sampling cost.

\begin{table}[!htbp]
\centering
\caption{
Runtime comparison on the 30-dimensional GMM under $\sigma_\mu=10$ and $\rho=0.1$.
Results are mean $\pm$ std over 5 runs.
}
\label{tab:gmm_runtime}
\small
\setlength{\tabcolsep}{5pt}
\begin{tabular}{lcc}
\toprule
Method & Annealing ($\gamma=2.0$) & Reward tilting ($\sigma = 100$) \\
\midrule
FKC on EAT & $5.357 \pm 0.406$ (s) & $5.996 \pm 0.736$ (s) \\
EBFKC w/o VCG on MIT & $3.098 \pm 0.537$ (s) & $3.283 \pm 0.445$ (s) \\
EBFKC on MIT  & $3.626 \pm 0.376$ (s) & $4.061 \pm 0.963$ (s) \\
EBFKC on EAT (Ours) & $5.267 \pm 0.082$ (s) & $5.503 \pm 0.149$ (s) \\
\bottomrule
\end{tabular}
\end{table}

\subsection{Double-well results}
\label{app:dw_additional_results}

\begin{figure}[!htbp]
    \centering
    \includegraphics[width=\textwidth]
    {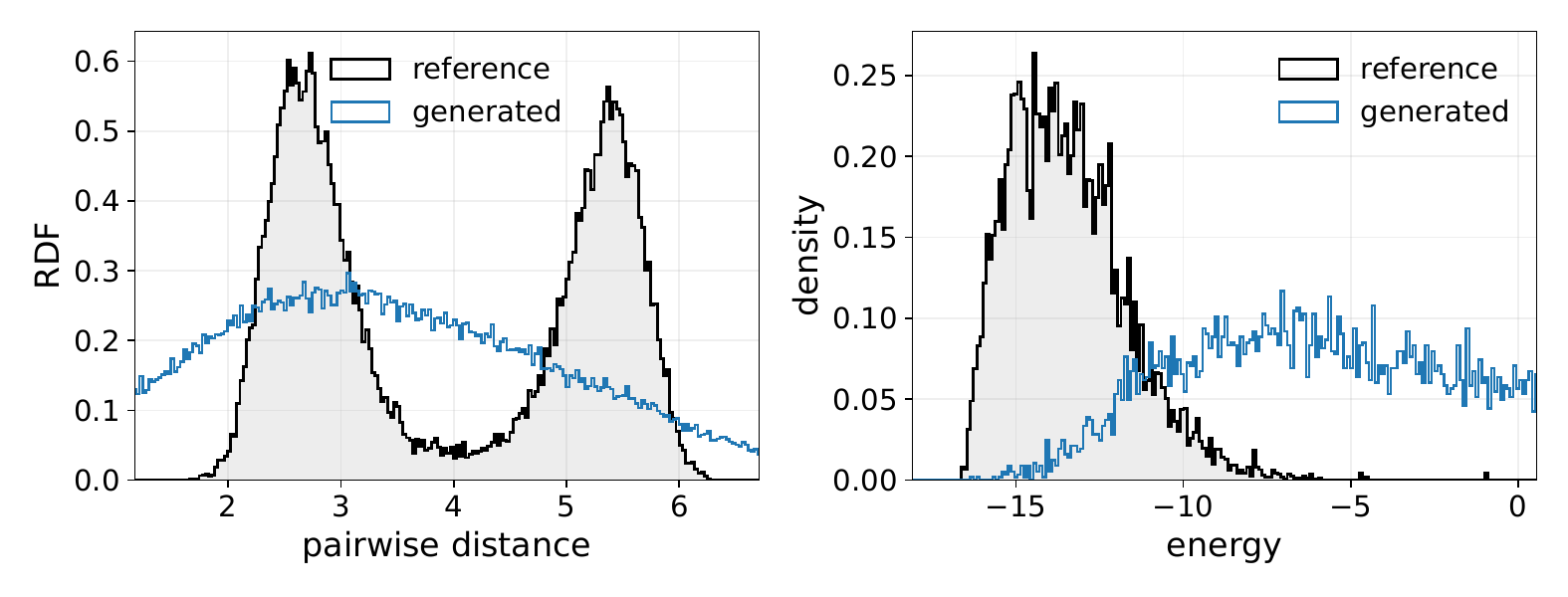}
    \caption{
    Comparison of the reference distribution and reverse diffusion
    samples from the pretrained energy-based diffusion model on the
    double-well system.
    }
    \label{fig:dw_ebm_reverse_diffusion}
\end{figure}

\begin{figure}[!htbp]
    \centering
    \includegraphics[width=0.9\textwidth]{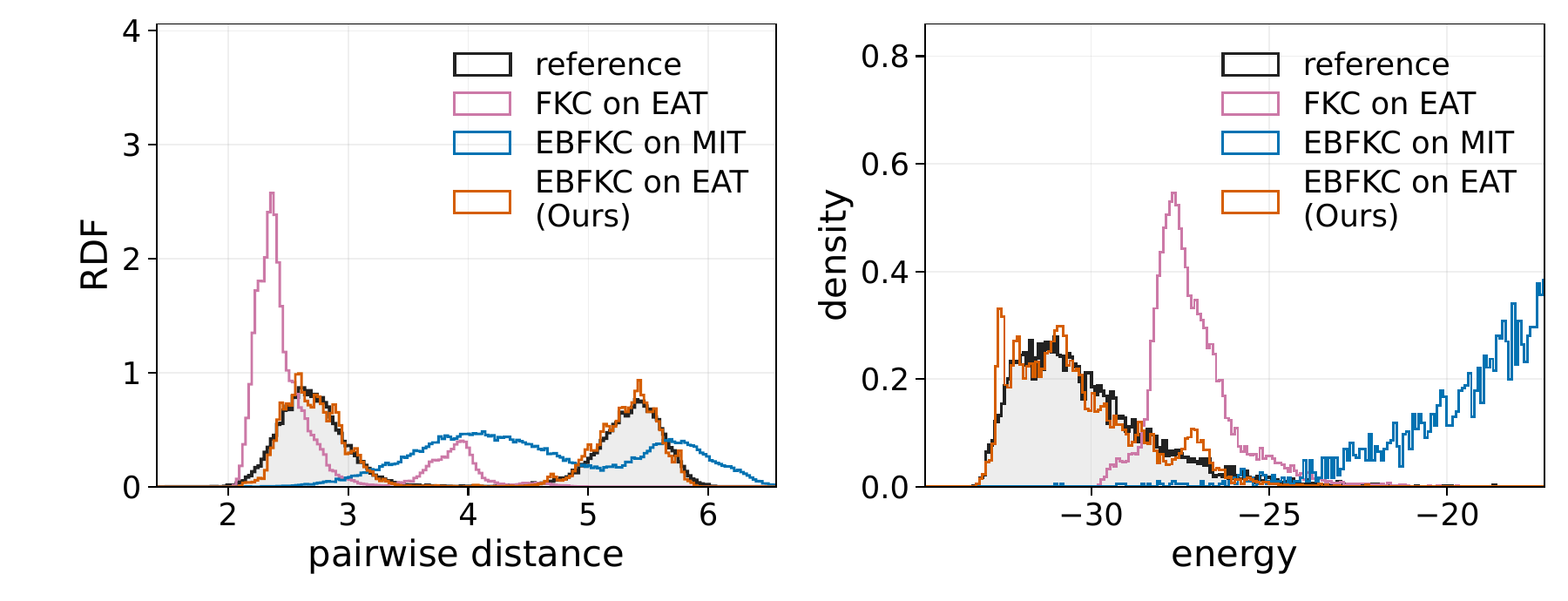}
    \begin{minipage}[t]{0.43\textwidth}
        \centering
        \textbf{(a) Radial Distribution Function}
    \end{minipage}
    \hfill
    \begin{minipage}[t]{0.43\textwidth}
        \centering
        \textbf{(b) Energy Distribution}
    \end{minipage}
    \caption{
    Comparison of the reference distribution, FKC on EAT, EBFKC on MIT, and EBFKC on EAT for the double-well system under annealing.
    }
    \label{fig:dw_annealing}
\end{figure}

\begin{figure}[!htbp]
    \centering
    \includegraphics[width=0.9\textwidth]{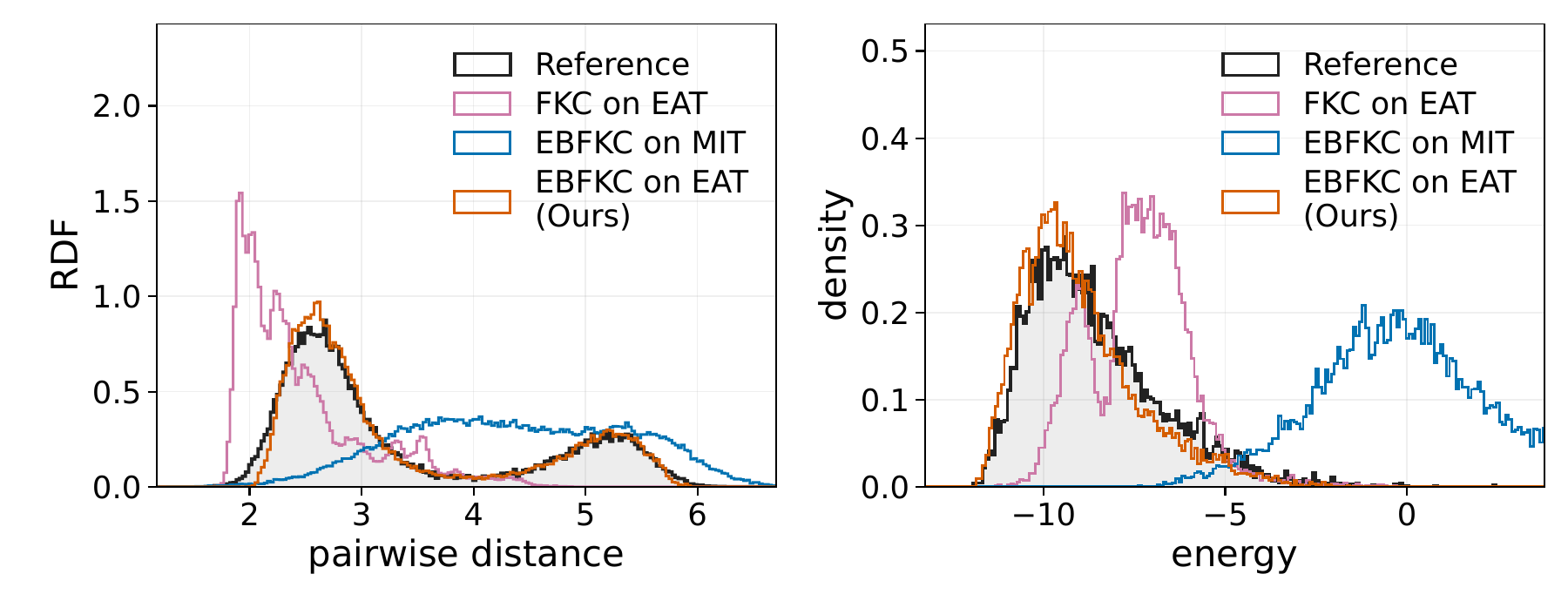}
    \begin{minipage}[t]{0.43\textwidth}
        \centering
        \textbf{(a) Radial Distribution Function}
    \end{minipage}
    \hfill
    \begin{minipage}[t]{0.43\textwidth}
        \centering
        \textbf{(b) Energy Distribution}
    \end{minipage}
    \caption{
    Comparison of the reference distribution, FKC on EAT, EBFKC on MIT, and EBFKC on EAT for the double-well system under reward tilting.
    }
    \label{fig:dw_reward_tilting}
\end{figure}

Fig.~\ref{fig:dw_ebm_reverse_diffusion} demonstrates that the pretrained energy-based diffusion model is imperfect: 
its reverse diffusion samples fail to reproduce the reference RDF peaks and exhibit an energy distribution shifted toward higher energies.
Fig.~\ref{fig:dw_annealing} and Fig.~\ref{fig:dw_reward_tilting} show the
results of annealing and reward tilting, respectively. In both tasks,
EBFKC on EAT closely matches the target RDF and energy distribution, whereas FKC on EAT and EBFKC on MIT retain substantial deviations from the reference.

Table~\ref{tab:dw_runtime} reports the runtime of each method on the DW-4 tasks.
For DW-4, we evaluate the required second-order derivatives of the target energy directly, without finite differences, so EBFKC on EAT takes reasonable 11\% longer than EBFKC on MIT in both tasks.

\begin{table}[!htbp]
\centering
\caption{
Runtime comparison on the DW-4 tasks. Results are
mean $\pm$ std over 5 runs.
}
\label{tab:dw_runtime}
\small
\setlength{\tabcolsep}{5pt}
\begin{tabular}{lcc}
\toprule
Method & Annealing ($\gamma=2.0$)  & Reward tilting ($\lambda'=0.5$)  \\
\midrule
FKC on EAT & $412.594 \pm 16.254$ (s) & $416.153 \pm 17.053$ (s) \\
EBFKC w/o VCG on MIT & $342.789 \pm 16.330$ (s) & $339.989 \pm 13.836$ (s) \\
EBFKC on MIT  & $367.378 \pm 14.568$ (s) & $365.100 \pm 18.214$ (s) \\
EBFKC on EAT (Ours) & $406.058 \pm 16.066$ (s) & $403.458 \pm 14.271$ (s) \\
\bottomrule
\end{tabular}
\end{table}

\subsection{Lennard--Jones results}
\label{app:lj_annealing_results}

\begin{figure}[!htbp]
    \centering
    \includegraphics[width=\textwidth]{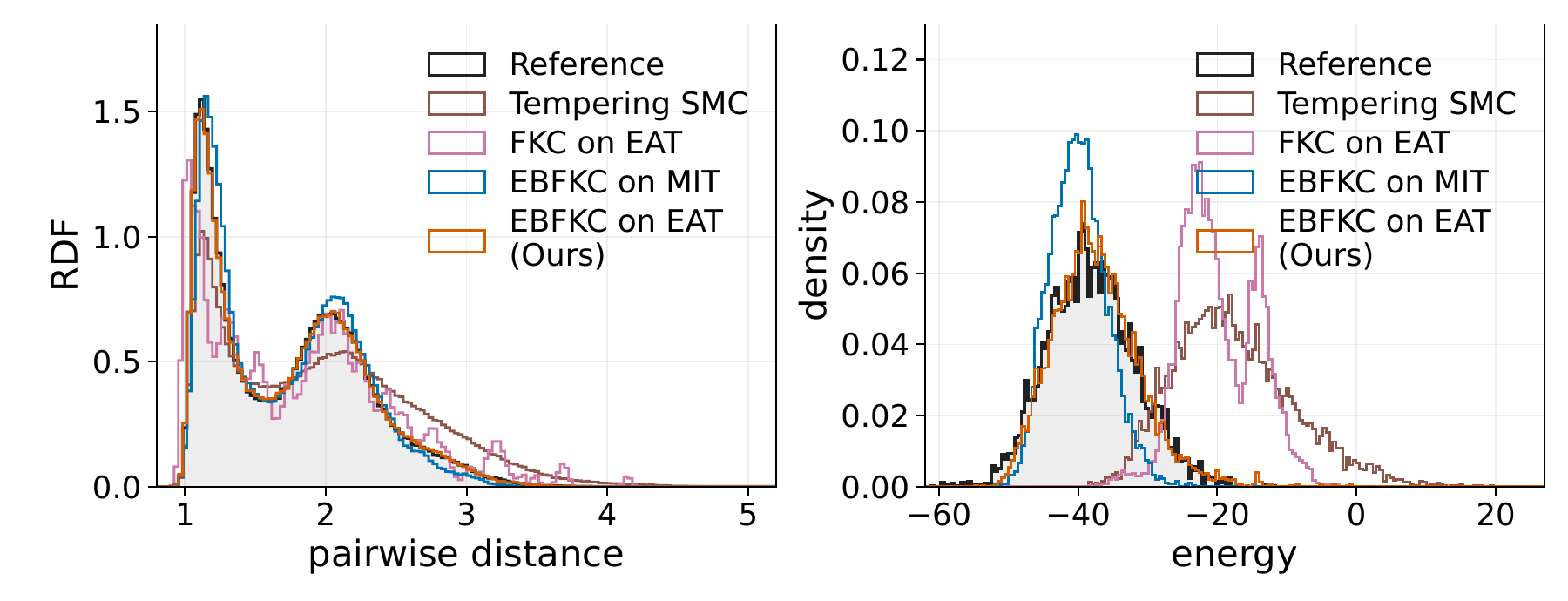}
    \begin{minipage}[t]{0.48\textwidth}
        \centering
        \textbf{(a) Radial Distribution Function}
    \end{minipage}
    \hfill
    \begin{minipage}[t]{0.48\textwidth}
        \centering
        \textbf{(b) Energy Distribution}
    \end{minipage}
    \caption{
    Comparison of the reference distribution, Tempering SMC, FKC on EAT, EBFKC on MIT and EBFKC on EAT for the Lennard-Jones system under annealing.
    }
    \label{fig:lj_annealing}
\end{figure}

\paragraph{Annealing results.} 
Fig.~\ref{fig:lj_annealing} compares the radial distribution functions and energy distributions of tempering SMC, FKC on EAT, EBFKC on MIT and EBFKC on EAT with the reference under LJ-13 annealing.
EBFKC on EAT closely reproduces the reference RDF peaks and energy distribution, whereas EBFKC on MIT deviates from the reference RDF and shifts the energy distribution severely. 
Tempering SMC shifts the energy distribution toward higher energies, which demonstrates tempering SMC alone is insufficient to accurately sample from the Lennard--Jones system.
Table~\ref{tab:particle_results} summarizes their performance across five metrics on DW-4 and LJ-13 under annealing and reward tilting.

\begin{table}[!htbp]
\centering
\caption{Performance comparison on particle systems (DW-4 and LJ-13).
Results are reported as mean $\pm$ std over 5 runs.}
\label{tab:particle_results}
\scriptsize
\setlength{\tabcolsep}{2.5pt}
\resizebox{\textwidth}{!}{%
\begin{tabular}{lllrrrrr}
\toprule
System & Task & Method & $|\Delta\mathrm{NLL}|$ & MMD & SWD
& $W_1^{\mathrm{RDF}}$ & $W_1^E$ \\
\midrule
DW
& Annealing
& FKC on EAT
& $23.9565 \pm 0.9045$ & $0.2659 \pm 0.0059$ & $0.6263 \pm 0.0511$
& $1.6010 \pm 0.0090$ & $23.9565 \pm 0.9045$ \\
&
& EBFKC w/o VCG on MIT
& $18.1923 \pm 1.0910$ & $0.4315 \pm 0.0645$ & $0.9894 \pm 0.1340$
& $0.7005 \pm 0.0307$ & $18.1923 \pm 1.0910$ \\
&
& EBFKC on MIT
& $19.7381 \pm 0.2039$ & $0.0996 \pm 0.0029$ & $0.2341 \pm 0.0217$
& $0.7582 \pm 0.0069$ & $19.7381 \pm 0.2039$ \\
&
& EBFKC on EAT (Ours)
& $\boldsymbol{0.3389 \pm 0.2107}$ & $\boldsymbol{0.0816 \pm 0.0193}$
& $\boldsymbol{0.2032 \pm 0.0441}$ & $\boldsymbol{0.3943 \pm 0.1437}$
& $\boldsymbol{0.3754 \pm 0.1556}$ \\
&
Reward tilting
& FKC on EAT
& $7.2620 \pm 0.3554$ & $0.1506 \pm 0.0037$ & $0.3644 \pm 0.0313$
& $0.9265 \pm 0.0152$ & $7.2620 \pm 0.3554$ \\
&
& EBFKC w/o VCG on MIT
& $8.1395 \pm 1.0000$ & $0.5666 \pm 0.2091$ & $1.1923 \pm 0.4556$
& $0.7240 \pm 0.1778$ & $8.1395 \pm 1.0000$ \\
&
& EBFKC on MIT
& $9.3752 \pm 0.0563$ & $0.1500 \pm 0.0029$ & $0.3403 \pm 0.0113$
& $1.0112 \pm 0.0023$ & $9.3752 \pm 0.0563$ \\
&
& EBFKC on EAT (Ours)
& $\boldsymbol{0.3769 \pm 0.0568}$ & $\boldsymbol{0.0178 \pm 0.0028}$
& $\boldsymbol{0.0461 \pm 0.0052}$ & $\boldsymbol{0.0690 \pm 0.0065}$
& $\boldsymbol{0.3769 \pm 0.0568}$ \\
\midrule
LJ
& Annealing
& Tempering SMC
& $22.1007 \pm 3.3704$ & $0.3914 \pm 0.1333$ & $0.4064 \pm 0.1320$
& $0.2928 \pm 0.0501$ & $22.1007 \pm 3.3704$ \\
&
& FKC on EAT
& $30.7764 \pm 3.4780$ & $0.3153 \pm 0.1421$ & $0.3384 \pm 0.1562$
& $0.2553 \pm 0.0397$ & $30.7764 \pm 3.4780$ \\
&
& EBFKC w/o VCG on MIT
& $3.9507 \pm 1.1581$ & $0.2068 \pm 0.0420$ & $0.2222 \pm 0.0311$
& $0.0467 \pm 0.0099$ & $4.0748 \pm 1.0444$ \\
&
& EBFKC on MIT
& $3.4750 \pm 0.8181$ & $0.1891 \pm 0.0747$ & $0.2176 \pm 0.0831$
& $0.0411 \pm 0.0108$ & $3.5841 \pm 0.7376$ \\
&
& EBFKC on EAT (Ours)
& $\boldsymbol{0.9211 \pm 0.3994}$ & $\boldsymbol{0.0600 \pm 0.0059}$
& $\boldsymbol{0.0647 \pm 0.0098}$ & $\boldsymbol{0.0121 \pm 0.0075}$
& $\boldsymbol{0.9411 \pm 0.3991}$ \\
&
Reward tilting
& Tempering SMC
& $14.4944 \pm 3.6800$ & $0.3141 \pm 0.1519$ & $0.3439 \pm 0.1917$
& $0.4414 \pm 0.2101$ & $14.4944 \pm 3.6800$ \\
&
& FKC on EAT
& $17.5700 \pm 0.4250$ & $0.1246 \pm 0.0159$ & $0.1536 \pm 0.0328$
& $0.3275 \pm 0.0051$ & $17.5700 \pm 0.4250$ \\
&
& EBFKC w/o VCG on MIT
& $7.3426 \pm 1.3823$ & $0.2026 \pm 0.0798$ & $0.2316 \pm 0.0901$
& $0.1528 \pm 0.0262$ & $7.3493 \pm 1.3764$ \\
&
& EBFKC on MIT
& $6.4575 \pm 0.5650$ & $0.1441 \pm 0.0356$ & $0.1658 \pm 0.0526$
& $0.1358 \pm 0.0137$ & $6.4618 \pm 0.5605$ \\
&
& EBFKC on EAT (Ours)
& $\boldsymbol{0.1366 \pm 0.0832}$ & $\boldsymbol{0.0345 \pm 0.0041}$
& $\boldsymbol{0.0425 \pm 0.0023}$ & $\boldsymbol{0.0140 \pm 0.0024}$
& $\boldsymbol{0.2454 \pm 0.0837}$ \\
\bottomrule
\end{tabular}
}
\end{table}

\paragraph{Runtime.}

Table~\ref{tab:lj_runtime} reports the runtime of each method on the Lennard--Jones benchmark.
We use the look-ahead map on LJ-13 and compute the required second-order derivatives using finite differences, whose evaluations can be parallelized across particles. 
These additional evaluations make the relative runtime of EBFKC on EAT longer than EBFKC on MIT with/without VCG.

\begin{table}[!htbp]
\centering
\caption{
Runtime comparison on the LJ-13 tasks. Results are
mean $\pm$ std over 5 runs.
}
\label{tab:lj_runtime}
\small
\setlength{\tabcolsep}{5pt}
\begin{tabular}{lcc}
\toprule
Method & Annealing ($\gamma=2.0$) & Reward tilting ($\lambda'=0.5$) \\
\midrule
FKC on EAT & $1484.691 \pm 1.749$ (s) & $2074.467 \pm 2.337$ (s) \\
EBFKC w/o VCG on MIT & $670.970 \pm 0.693$ (s)  & $1199.705 \pm 1.273$ (s)\\
EBFKC on MIT  & $671.882 \pm 0.804$ (s)  & $1201.382 \pm 1.070$ (s) \\
EBFKC on EAT (Ours) & $1339.710 \pm 0.454$ (s) & $1820.829 \pm 1.568$ (s) \\
\bottomrule
\end{tabular}
\end{table}

\subsection{ALDP annealing results}
\label{app:aldp_annealing_results}

\begin{figure}[!htbp]
    \centering
    \includegraphics[width=\textwidth]{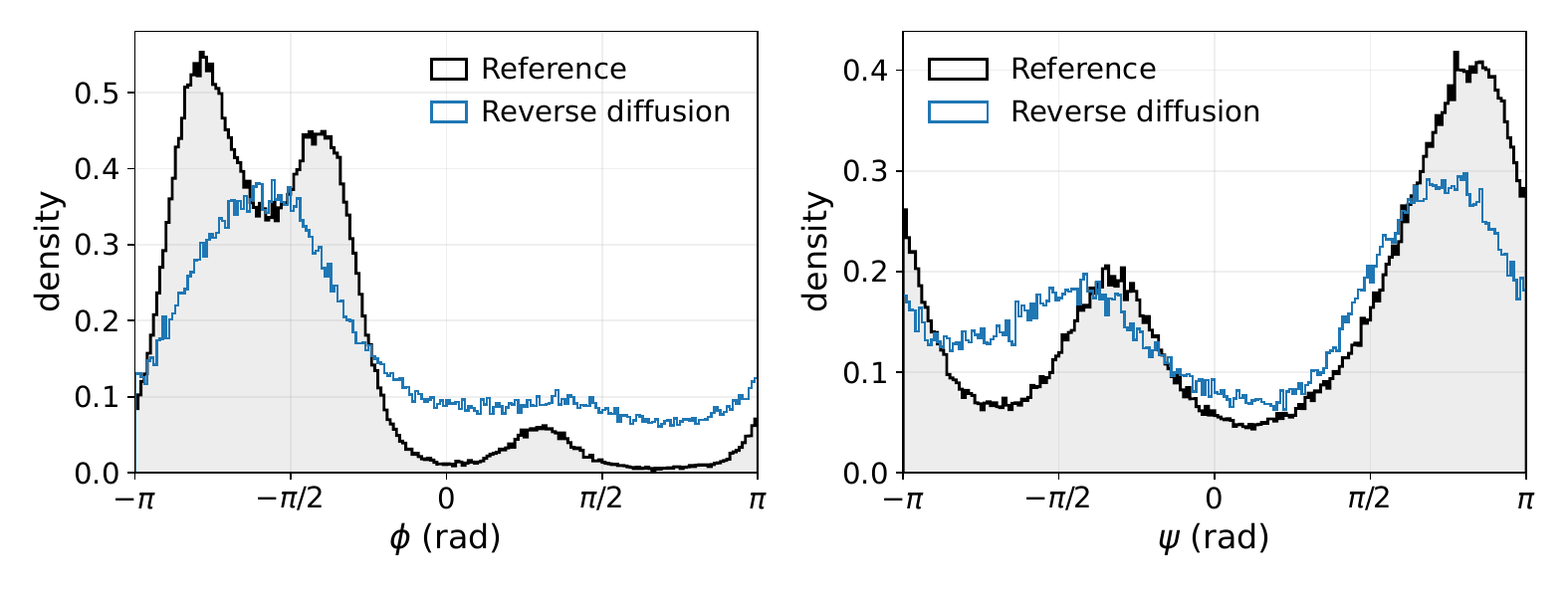}
    \caption{ALDP backbone dihedral marginals at $800\,\mathrm{K}$ from the reference distribution and pretrained reverse diffusion: $\phi$ (left) and $\psi$ (right). The reverse diffusion marginals are estimated from 100,000 generated samples.}
    \label{fig:aldp_800_ebm_phi_psi}
\end{figure}

\begin{figure}[!htbp]
    \centering
    \includegraphics[width=\textwidth]
    {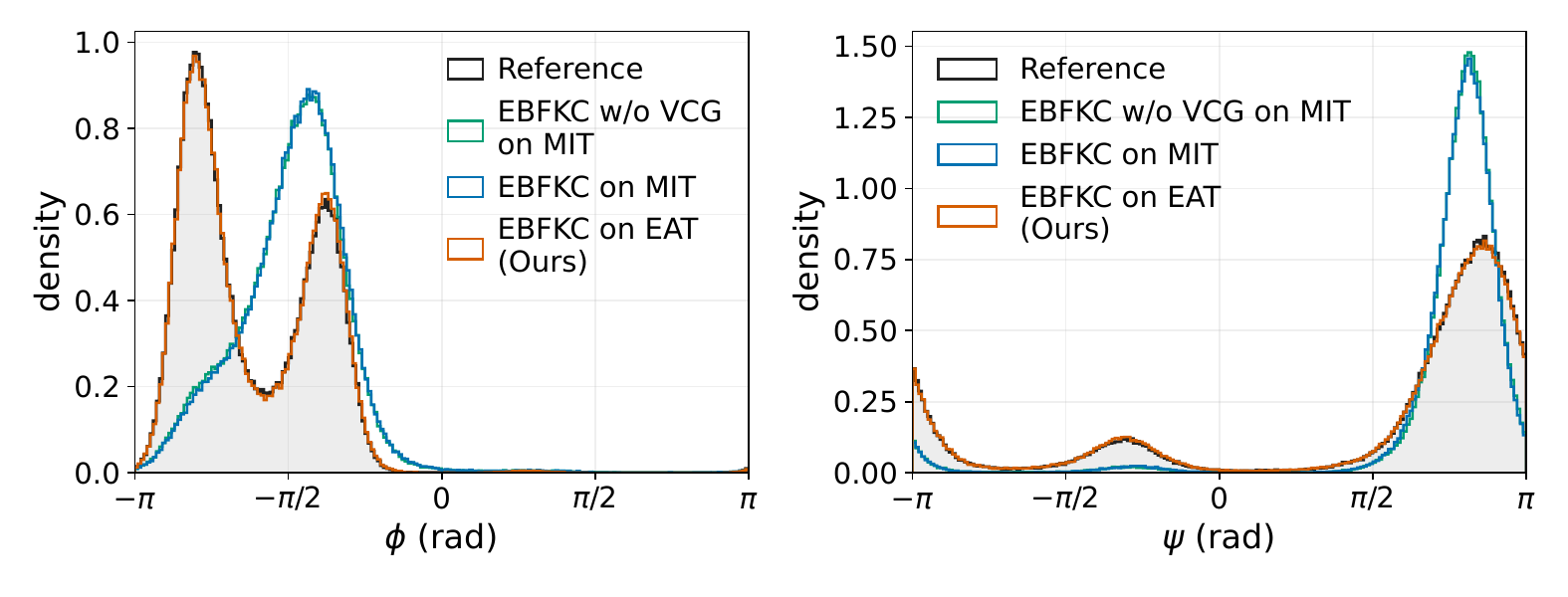}
    \caption{
    ALDP annealing marginals comparison for $\phi$ (left) and $\psi$ (right).
    EBFKC on EAT closely matches the $300\,\mathrm{K}$ reference,
    whereas both MIT variants deviate substantially. The marginals are
    estimated from all 250,000 samples generated by each method and
    all reference samples at $300\,\mathrm{K}$.
    }
    \label{fig:aldp_marginals}
\end{figure}

At $800\,\mathrm{K}$, the pretrained model captures the broad shapes of the dihedral marginals but retains visible discrepancies from the reference (Fig.~\ref{fig:aldp_800_ebm_phi_psi}).
After annealing to $300\,\mathrm{K}$, EBFKC on EAT closely reproduces the reference marginals of both backbone dihedrals (Fig.~\ref{fig:aldp_marginals}).
Both MIT variants underrepresent the leftmost major peak in $\phi$ and overconcentrate probability near the dominant positive-$\psi$ peak.

Table~\ref{tab:aldp_runtime} reports the runtime of each method for ALDP annealing using 1000 particles.
Unlike the particle system experiments, OpenMM provides only first-order derivatives of the target energy, so the second-order derivatives required by EBFKC on EAT are computed using finite-difference approximation.
Because these additional OpenMM evaluations cannot be parallelized across particles, EBFKC on EAT has a longer runtime than EBFKC on MIT with/without VCG.

\begin{table}[!htbp]
\centering
\caption{
Runtime comparison for ALDP annealing from $800\,\mathrm{K}$ to $300\,\mathrm{K}$. Results are reported as mean $\pm$ std over 5 runs.
}
\label{tab:aldp_runtime}
\small
\setlength{\tabcolsep}{8pt}
\begin{tabular}{lc}
\toprule
Method & Sampling time (s) \\
\midrule
EBFKC w/o VCG on MIT  & $142.202 \pm 0.636$ \\
EBFKC on MIT  & $142.932 \pm 0.119$ \\
EBFKC on EAT (Ours)  & $508.433 \pm 1.109$ \\
\bottomrule
\end{tabular}
\end{table}

\paragraph{Look-ahead map ablation study.}
We ablate the denoised look-ahead map in the ALDP annealing experiment to demonstrate its necessity in high-dimensional real-world applications.
The ablation reveals two concrete failure modes. \emph{First,} with alive filtering enabled, the sampler completes all 1,000 scheduled annealing transitions but is terminated by the endpoint alive check because none of the 1,000 terminal particles satisfies the joint validity criterion: energy below the cutoff and finite OpenMM energy and force.
\emph{Second,} without alive filtering, the 1,000-particle run becomes numerically unstable at the 816th transition when the target-energy correction activates,
and all 1,000 particle-wise VCG potentials become non-finite because the target energy is evaluated directly at noisy particle states.
Together, these failures show that the look-ahead map is necessary for producing physically meaningful terminal-space predictions before the target energy and its gradient are evaluated by OpenMM. Alive filtering can shield the VCG calculation from occasional invalid particles, but it cannot compensate for the systematic failure caused by removing the look-ahead map.

\subsection{Molecular free-energy estimation results}
\label{app:fkc_molecular_free_energy_results}

\begin{figure}[!htbp]
    \centering
    \includegraphics[width=\textwidth]{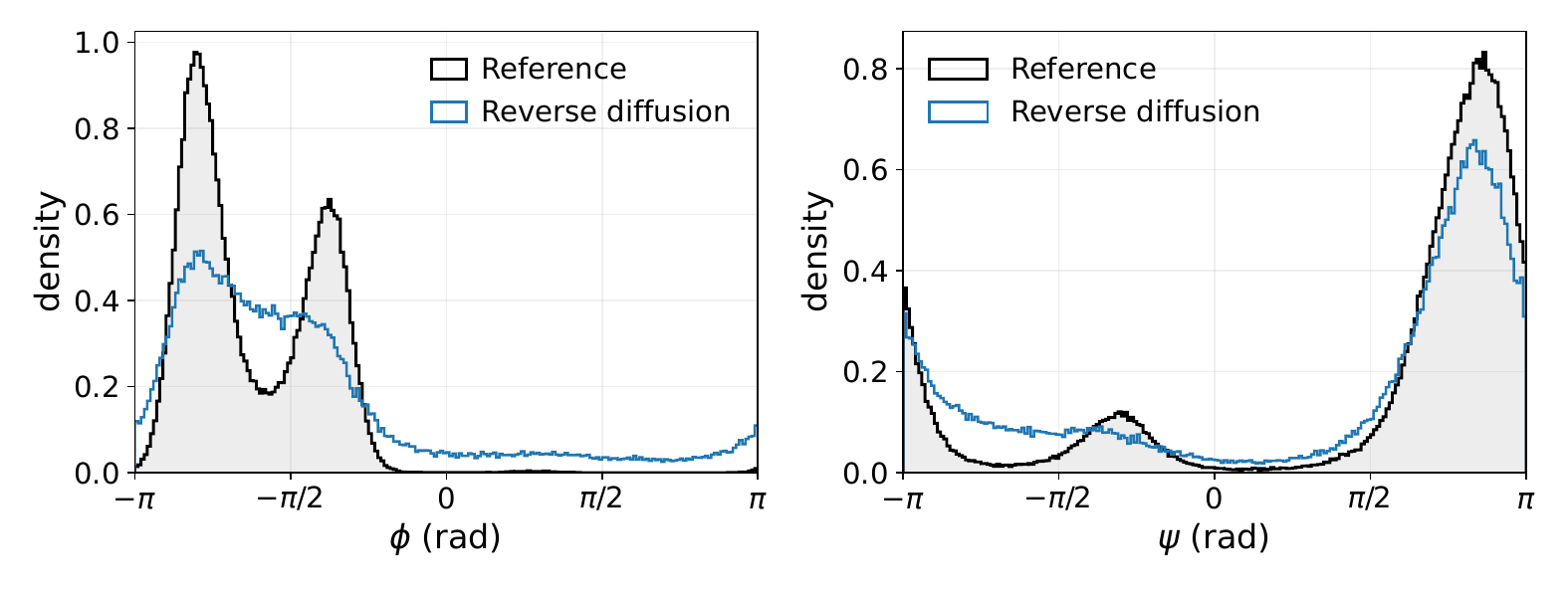}
    \caption{ALDP backbone dihedral marginals at $300\,\mathrm{K}$ from the reference distribution and pretrained reverse diffusion: $\phi$ (left) and $\psi$ (right). The reverse diffusion marginals are estimated from 100,000 generated samples.}
    \label{fig:aldp_300_ebm_phi_psi}
\end{figure}

\begin{figure}[!htbp]
    \centering
    \includegraphics[width=\textwidth]{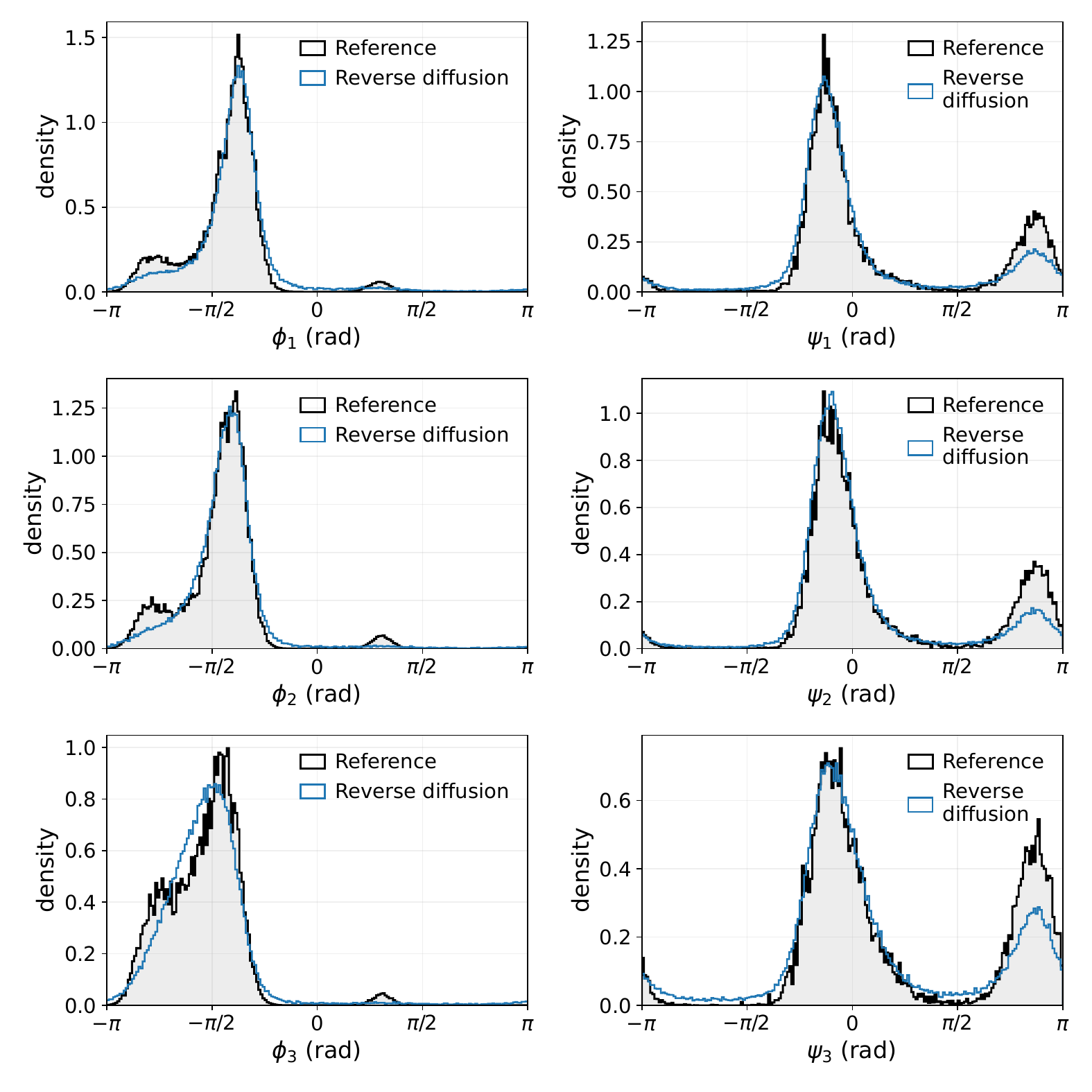}
    \caption{Ala-4 backbone dihedral marginals at $300\,\mathrm{K}$ from the reference distribution and pretrained reverse diffusion. The reverse diffusion marginals are estimated from 100,000 generated samples.}
    \label{fig:ala4_ebm_phi_psi}
\end{figure}

Figures~\ref{fig:aldp_300_ebm_phi_psi} and~\ref{fig:ala4_ebm_phi_psi} compare the dihedral marginals obtained by pretrained reverse diffusion with the reference distributions at $300\,\mathrm{K}$ for ALDP and Ala-4, respectively.
The pretrained ALDP model captures the broad shapes of the reference marginal distributions but fails to reproduce some details, while the pretrained Ala-4 model closely reproduces the marginal distributions with some remaining discrepancies. These results indicate that both pretrained models capture the main features of the reference distributions but remain imperfect.

\begin{figure}[!htbp]
    \centering
    \begin{minipage}[t]{0.48\textwidth}
        \centering
        \includegraphics[width=\linewidth]
        {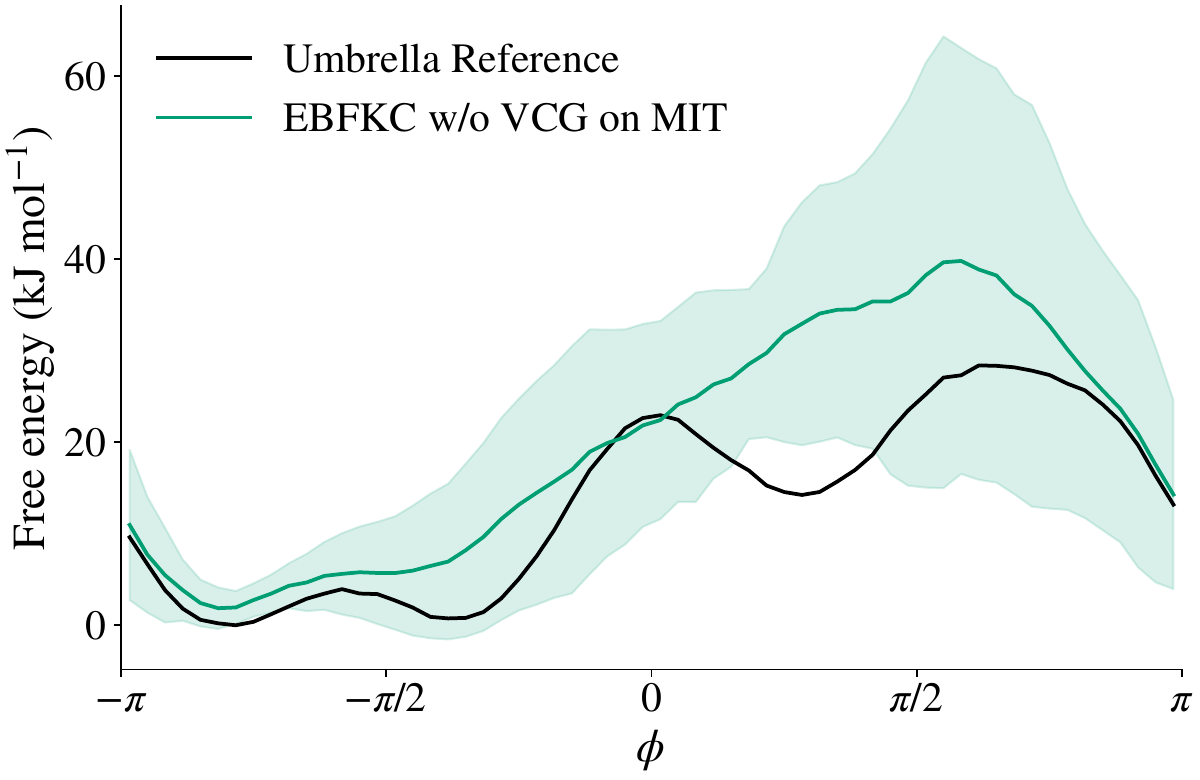}
        \par\small\textbf{(a) ALDP: compared with umbrella reference}
    \end{minipage}
    \hfill
    \begin{minipage}[t]{0.48\textwidth}
        \centering
        \includegraphics[width=\linewidth]
        {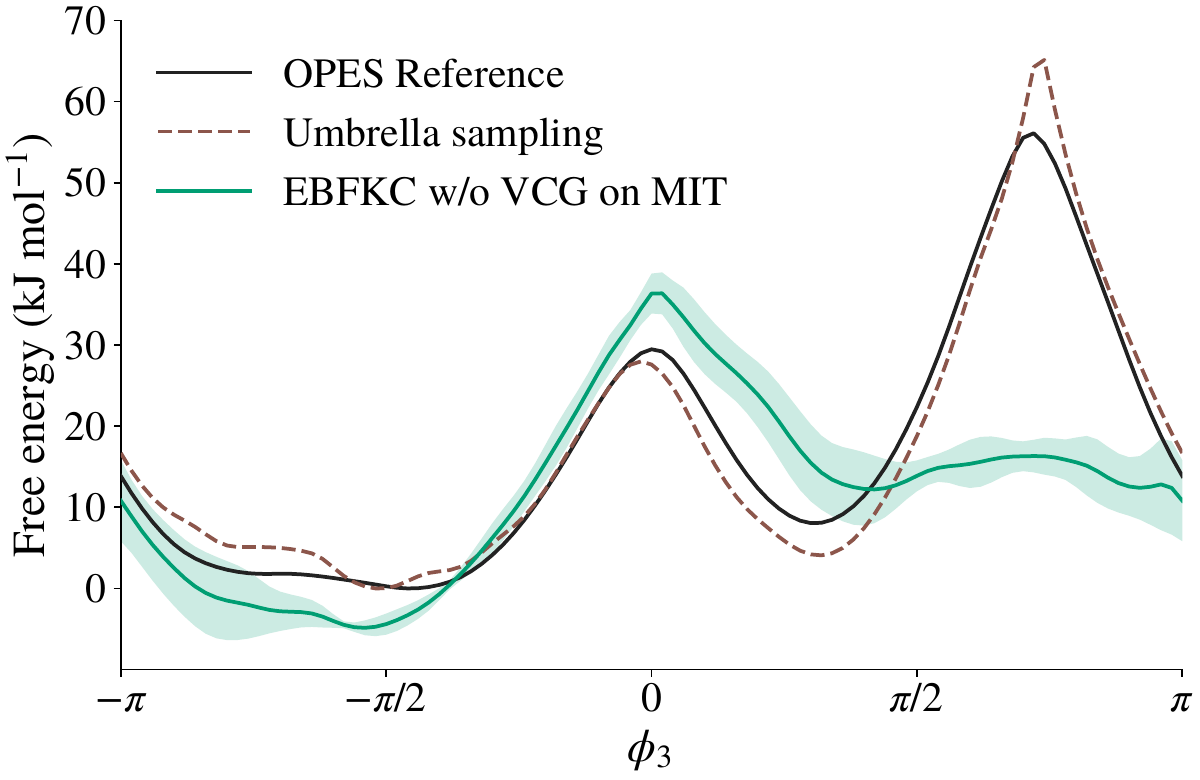}
        \par\small\textbf{(b) Ala-4: compared with OPES reference}
    \end{minipage}
    \caption{
    EBFKC w/o VCG on MIT free-energy estimates for the molecular systems. (a) ALDP along $\phi$, compared with traditional umbrella sampling reference.
    (b) Ala-4 along $\phi_3$, compared with the OPES reference and umbrella sampling results. Results are based on 5 independent runs.
    }
    \label{fig:fkc_molecular_free_energy}
\end{figure}

Fig.~\ref{fig:fkc_molecular_free_energy} presents the free-energy profiles estimated by EBFKC w/o VCG on MIT for ALDP and Ala-4.
For ALDP, the method exhibits large estimation errors and substantial variability across runs, overestimating the free energy in the positive-$\phi$ region and failing to recover the reference basin there.
Comparison with EBFKC on MIT in Fig.~\ref{fig:molecular_free_energy}(a)
highlights the benefit of VCG for ALDP free-energy estimation.
For Ala-4, the method substantially underestimates the high free-energy barrier in the positive-$\phi_3$ region.

Table~\ref{tab:aldp_free_energy_runtime} reports the runtime of each method for ALDP free-energy estimation in window 13 with 200 particles.
As in the ALDP annealing experiment, the finite-difference evaluation of the second-order target-energy derivatives introduces additional OpenMM evaluations which cannot be parallelized across particles, so EBFKC on EAT has a longer runtime than EBFKC on MIT with/without VCG. 
Also we note that the runtime of EBFKC on MIT is slightly shorter than EBFKC w/o VCG on MIT, which is reasonable because the VCG method can reduce the variance of particle weights and thus reduce the number of resampling steps, which can save some MALA runtime.

\begin{table}[!htbp]
\centering
\caption{
Runtime comparison for ALDP free-energy estimation in window 13 with
200 particles. Results are reported as mean $\pm$ std over 5 runs.
}
\label{tab:aldp_free_energy_runtime}
\small
\setlength{\tabcolsep}{8pt}
\begin{tabular}{lc}
\toprule
Method & Sampling time (s) \\
\midrule
EBFKC w/o VCG on MIT  & $119.30 \pm 10.43$ \\
EBFKC on MIT & $106.17 \pm 3.24$ \\
EBFKC on EAT (Ours) & $269.19 \pm 9.11$ \\
\bottomrule
\end{tabular}
\end{table}

Table~\ref{tab:ala4_free_energy_runtime} reports the corresponding runtime
for Ala-4 free-energy estimation in window 13 with 12 particles.
Because the larger pretrained model makes each baseline sampling step more
expensive, target-energy correction contributes a smaller relative runtime
overhead in Ala-4 than in ALDP.

\begin{table}[!htbp]
\centering
\caption{
Runtime comparison for Ala-4 free-energy estimation in window 13 with 12 particles. Results are reported as mean $\pm$ std over 5 runs.
}
\label{tab:ala4_free_energy_runtime}
\small
\setlength{\tabcolsep}{8pt}
\begin{tabular}{lc}
\toprule
Method & Sampling time (s) \\
\midrule
EBFKC w/o VCG on MIT & $389.05 \pm 14.20$ \\
EBFKC on MIT & $376.89 \pm 17.25$ \\
EBFKC on EAT (Ours) & $504.87 \pm 12.79$ \\
\bottomrule
\end{tabular}
\end{table}

\end{document}